\documentclass[11pt,letterpaper]{article}

\usepackage{hyperref} 
\usepackage{fullpage}
\usepackage[utf8]{inputenc}
\usepackage{amssymb,amsmath,amsthm}
\usepackage{float}
\usepackage{graphicx}
\usepackage{algorithmic}
\usepackage{algorithm}
\usepackage{tikz}

\usepackage[titletoc,title]{appendix}

\date{}
\newtheorem{theorem}{Theorem}[section]

\newtheorem{corollary}[theorem]{Corollary}
\newtheorem{proposition}[theorem]{Proposition}
\newtheorem{definition}[theorem]{Definition}
\newtheorem{problem}[theorem]{Problem}
\newtheorem{example}[theorem]{Example}

\title{Higher-Order Group Synchronization}

\author{
Adriana L. Duncan 
\thanks{
    Corresponding author. Department of Mathematics, The University of Texas at Austin, Texas, USA, \texttt{aduncan@math.utexas.edu}.
    Research supported in part by NSF DMS 1937215.
    }
\and Joe Kileel 
\thanks{
    Department of Mathematics and Oden Institute, The University of Texas at Austin, Texas, USA, \texttt{jkileel@math.utexas.edu}. Research supported in part by NSF DMS 2309782, NSF DMS 2436499, NSF CISE-IIS 2312746 and
DE SC0025312.
    } 
 }

\begin{document}

\maketitle

\begin{abstract}
Group synchronization is the problem of determining reliable global estimates from noisy local measurements on networks. 
The typical task for group synchronization is to assign elements of a group to the nodes of a graph in a way that respects  group elements given on the edges which encode information about local pairwise relationships between the nodes. 
In this paper, we introduce a novel \textit{higher-order group synchronization} problem which operates on a hypergraph and seeks to synchronize higher-order local measurements on the hyperedges to obtain global estimates on the nodes. 
Higher-order group synchronization is motivated by applications to computer vision and image processing, among other computational problems.
First, we define the problem of higher-order group synchronization and discuss its mathematical foundations. 
Specifically, we give necessary and sufficient synchronizability conditions which establish the importance of cycle consistency in higher-order group synchronization. 
Then, we propose the first computational framework for general higher-order group synchronization; it acts \textit{globally} and \textit{directly} on higher-order measurements using a message passing algorithm. 
We discuss theoretical guarantees for our framework, including convergence analyses under outliers and noise.   Finally, we show potential advantages of our method through numerical experiments. 
In particular, we show that in certain cases our higher-order method applied to rotational and angular synchronization outperforms standard pairwise synchronization methods and is more robust to outliers.
We also show that our method has comparable performance on simulated cryo-electron microscopy (cryo-EM) data compared to a standard cryo-EM reconstruction package.
\end{abstract}

\section{Introduction}

The classical problem of group synchronization is to recover a set of elements \(\{g_i\}_{i \in[m]}\) from a group \(G\) given a subset of their (noisy) pairwise ratios, \(g_ig_j^{-1} \in G\). 
This task has broad applications in computer vision \cite{arrigoni_robust_2015,fan_spectral_2023,huang_translation_2017,shkolnisky_viewing_2012,singer_angular_2011}, robotics \cite{curcuringu_sensor_2012,rosen_certifiably_2019}, community detection \cite{cucuringu_synchronization_2015}, among other applied domains. 
The goal is to de-noise local information by exploiting redundancies and global consistency constraints in order to obtain reliable global information. For example, \cite{lerman_robust_2022} proposes a general group synchronization framework on graphs that exploits a cycle consistency condition.

In this paper, we propose a novel extended group synchronization framework that seeks to recover group elements \(\{g_i\}_{i \in [m]}\) from triples or $n$-wise collections of relative local information. 
This formulation is motivated by natural higher-order structures in applications, such as the relative poses between triples or quadruples of cameras encoded by multi-focal tensors as in computer vision \cite{hartley_multiple_2004} and relative poses between triples of cryo-electron microscopy (cryo-EM) images as classically determined by common lines data \cite{muller_algebraic_2024,vanheel_angular_1987}. 
Recently, in \cite{miao_tensor_2024}, it was found in the computer vision context that synchronization directly on the data of trifocal tensors can improve camera location estimation as opposed to standard pairwise based synchronization methods. 
However, beyond \cite{miao_tensor_2024}, limited mathematical and computational work has explored the synchronization of higher-order structures directly. 
A motivation for developing a general higher-order synchronization framework is to take advantage of increased redundancies among $n$-wise local measurements to improve the accuracy of global estimates, and also to develop a  framework applicable when pairwise measurements are unavailable (e.g., texture-poor data sets in computer vision \cite{he_detector_2024}).

\subsection{Motivating Applications} \label{subsec:motivation}

Group synchronization appears in a variety of important computational pipelines. 
In structure from motion (SfM) \cite{agarwal_building_2009,arrigoni_robust_2015,arrigoni_spectral_2016,kasten_global_2019,sengupta_new_2017,shi_estimation_2018,shi_robust_2022,wang_exact_2013} or cryo-EM \cite{bandeira_nonunique_2020,shkolnisky_viewing_2012,singer_angular_2011,singer_viewing_2011}, group synchronization helps situate a set of local relative camera positions to a global frame of reference to complete a 2-D to 3-D reconstruction task. 
In simultaneous localization and mapping (SLAM) from robotics \cite{curcuringu_sensor_2012,rosen_certifiably_2019}, group synchronization coordinates local observations perceived by a mobile robot to establish a global model of the environment. 
In these applications, group synchronization is applied to groups such as \(SO(d)\) and \(SE(d)\), often for \(d = 2\) or \(3\). 
For certain community detection or max-cut problems \cite{cucuringu_synchronization_2015}, group synchronization on \(\mathbb{Z}/2\mathbb{Z}\) can be used to sort objects into communities from local data about relations between their members. 
Further, group synchronization over the permutation group \(S_n\) aids in establishing globally consistent feature point labels from relative feature point matches \cite{li_fast_2022,pachauri_solving_2013,shi_scalable_2021}. 
See \cite{lerman_robust_2022} for a more detailed overview of the many applications of group synchronization. 

In all of these applications the underlying conceptual problem is the same: 
to determine a global set of data from \textit{pairwise} local observations. 
However, there are other kinds of local information that naturally arise. 
In cryo-EM, a classical method for determining viewing angles is known as angular reconstitution \cite{vainshtein_determination_1986,vanheel_angular_1987} which establishes the relative angles between a \textit{triple} of images. 
The mathematical foundations of single particle reconstruction in cryo-EM are based on the Fourier projection-slice theorem which states that the 2-D Fourier transform of a projection image corresponds to a planar slice of the 3-D Fourier transform of the molecule where the slice is  perpendicular to the viewing direction of the image \cite{frank_electron_2006,hadani_representation_2011,singer_detecting_2010}. 
In particular, this theorem implies that the planer slices of any two projection images from distinct viewing angles will intersect along a line, called the \textit{common line}. 
The basis of angular reconstitution is that the common lines data from \textit{three} images is needed to determine their relative positions.
In standard cryo-EM reconstruction pipelines, this triple-wise data is then reduced to pairwise relations (for example, by voting \cite{singer_detecting_2010}) before synchronization techniques are applied. 

In structure from motion applications, trifocal tensors encode all of the projective geometric relations between a \textit{triple} of views (see \cite{hartley_multiple_2004}). 
Trifocal tensors are  the higher-order analog to fundamental and essential matrices between two views (e.g., \cite{fan_instability_2022,fan_condition_2023}). 
However, unlike fundamental matrices, trifocal tensors uniquely determine camera positions, even when cameras are collinear, and thus may be preferable to a triple of fundamental  matrices \cite{hartley_multiple_2004}. 
Moreover, trifocal tensors can be determined from combinations of point and line correspondences between images \cite{kileel_minimal_2017}. 
In certain structure from motion settings (for example, when the images are texture-poor) it is difficult to determine feature point correspondences between pairs of images. 
Determining fundamental matrices requires a sufficient number feature point matches and thus for texture-poor datasets the line correspondences that determine trifocal relations give another option for determining relative poses. 
There is also the potential for datasets to contain a mixture of trifocal tensors and fundamental matrices when only some of the images are texture-poor.  Furthermore, $4$-wise relative poses between quadruples of images can be encoded by \textit{quadrifocal} tensors \cite{hartley_multiple_2004}, and therefore also brought into the mix.

These higher-order relative poses could play an important role in applications. 
However, synchronization of these structures directly and robustly has not been sufficiently explored. 
This is a main motivation for our work. 
In addition to the usefulness of synchronizing higher-order poses in settings where they naturally occur, there are other motivations for considering higher-order synchronization. 
For one, as mentioned group synchronization can be thought of as a denoising problem, where global estimates are determined by exploiting the redundancy of noisy local estimates. 
For example, if the local data is sufficiently dense, there are \(O(m^2)\) pieces of pairwise data for determining the positions of \(O(m)\) objects. 
Higher-order local data improves this redundancy, giving \(O(m^n)\) \(n\)-wise local estimates for determining \(O(m)\) positions, and may potentially provide better global estimates. 
Moreover, in some applications of group synchronization, data sets may be too large to consider all at once. 
Distributed synchronization aims to resolve this problem by first synchronizing subsets of the data then synchronizing the collection of subsets \cite{li_efficient_2024}. 
Higher-order synchronization could provide a new perspective on the distributed synchronization problem where local subsets of \(n\)-wise synchronized data need to be globally aligned for large \(n\).

\subsection{Contribution of This Work}

These are the main contributions of our work:
\\
\textbf{Novel higher-order group synchronization setup:} We establish the higher-order synchronization problem and discuss its mathematical foundations. 
Theorem \ref{thm:1cycleissynch} establishes necessary and sufficient conditions for synchronization in a general setting and Theorem \ref{thm:generalsynch} gives a sufficient condition for synchronizing a dataset.
\\
\textbf{General message passing framework:} CHMP (see Algorithm \ref{alg:CHMP}) is a general algorithmic framework for higher-order synchronization that works for any compact group: the first general higher-order synchronization framework. 
In Section \ref{sec:chmp-analysis}, we discuss theoretical guarantees of CHMP and prove global linear convergence in the setting of outliers and noise (see Theorems \ref{thm:noiselessrecovery} and \ref{thm:noisyrecovery}).
\\
\textbf{Competitive numerical results:} Through numerical experiments in Section \ref{sec:numerical-results}, CHMP is shown to be competitive with existing pairwise synchronization methods for angular and rotational synchronization.

\section{Higher-Order Group Synchronization Problem Setup}

Section \ref{subsec:classical-gs} reviews the classical setup for pairwise group synchronization. 
Section \ref{subsec:HOGS-setup} introduces the setup for a general higher-order group synchronization problem. 
Pairwise group synchronization can be seen as a special case of the higher-order group synchronization problem. 

\subsection{Classical Group Synchronization} \label{subsec:classical-gs}

Fix a group \(G\). 
In the classical setting of group synchronization, it is assumed that there is an underlying set of group elements \(\{g_i\}_{i \in V} \subseteq G\) that are assigned to the vertices of a graph \(\mathcal{G}(V,E)\) with vertex set \(V\) and edge set \(E\). 
The graph \(\mathcal{G}(V,E)\) is a simple undirected graph so that the edges in \(E\) are subsets of \(V\) of size \(2\). 
For each edge \(ij \in E\), a group element \(\overline{\gamma}_{ij} \in G\) is observed that represents the group ratio between the connected vertices. 
If 
\begin{equation}
    \overline{\gamma}_{ij} = g_i g_j^{-1} \text{ for all } ij \in E,
\end{equation}
then the set of edge ratios \(\{\overline{\gamma}_{ij}\}_{ij \in E}\) is \textit{compatible} with the set of vertex elements \(\{g_i\}_{i \in V}\). 
The goal of group synchronization is to use the observed edge ratios to find a compatible assignment of group elements to the vertices. 
A compatible set of vertex elements is unique up to a global action of the group. 
Indeed if \(g \in G\) and \(\{g_i\}_{i \in V}\) is a compatible vertex set, then \(\{g_i\cdot g\}_{i \in V}\) is also a compatible vertex set since \(\overline{\gamma}_{ij} = g_ig_j^{-1} = (g_i\cdot g) (g_j\cdot g)^{-1}\). 
In practical applications, this global ambiguity is not a limitation as the solution is unique up to a chosen frame of reference. 

In the case that the group ratios are not noisy or corrupted, the synchronization step is trivial and can be accomplished by fixing the group element of one vertex, \(g_0\), and assigning group elements to the remaining vertices by 
\begin{equation}\label{eq:incrementalestimation}
    g_i = \overline{\gamma}_{ij}\cdot g_j.
\end{equation}
The introduction of noise and outliers in the observed group ratios makes synchronization a challenging task. 
In the noisy setting, the naive process of assigning vertex elements by \eqref{eq:incrementalestimation} does not guarantee even approximate recovery of the underlying vertex elements, as the noise of each edge ratio is accumulated in each multiplication step. 
Thus, the assignment of vertex elements is highly dependent on the path chosen on the graph from \(g_0\) to \(g_i\). 

This gives rise to so-called ``global" methods in group synchronization, which aim to solve the minimization problem
\begin{equation}\label{eq:minimizeredge}
    \min_{\{g_i\}_{i \in V}} \sum_{ij \in H} d_G(\overline{\gamma}_{ij},g_ig_j^{-1}),
\end{equation}
where \(d_G\) is some distance on \(G\), by considering the contribution of each edge ratio simultaneously. 
Unfortunately, this minimization problem is generally not convex for many groups of interest, and convex relaxation techniques such as SDP \cite{singer_angular_2011} can distort the problem enough so that exact recovery is difficult to achieve in high corruption domains.
Another global approach is to denoise the data by estimating the edge corruption levels so that severely corrupted measurements can be discarded or \eqref{eq:minimizeredge} can be refined by appropriate edge weights. 
This can be done, for example, by using a message passing framework \cite{lerman_robust_2022,perry_message_2018,shi_message_2020}.

\subsection{Higher-Order Group Synchronization} \label{subsec:HOGS-setup}

Assume a hypergraph \(\mathcal{H}(V,H)\) where \(V\) is a set of \(m\) vertices and \(H\) is a set of hyperedges which is a subset of \(\mathcal{P}(V)\), the power set of \(V\). 
That is, hyperedges may contain any number of vertices. 
Given a group \(G\), assume there exists an underlying ground truth assignment of elements of \(G\) to the vertices of the hypergraph. 
This assignment is called the \textit{vertex potential} and it can be given as a map \(\rho: V \to G\) which assigns \(i \mapsto g_i\) for \(i \in V\) and \(g_i \in G\). 

For each hyperedge, an \(n\)-wise group ratio is observed where \(n\) is the number of nodes in the hyperedge. 
A collection of measurements for the group ratio of a hyperedge can only be determined up to an action of the group so the \(n\)-wise group ratio of a hyperedge, \(h = \{i_1, \dots, i_n\}\), is represented as an element of the coset space \(G^n / \Delta\) where \(\Delta = \{(g, \dots, g) \in G^n : g \in G\}\) is the diagonal subgroup of \(G^n\):
\begin{equation}
    \gamma_{h} = ({g}_{i_1}, \dots,{g}_{i_n})\Delta \in G^n/\Delta.
\end{equation}
Then the map \(\phi: H \to \sqcup_{l \geq 2} G^{l}/\Delta\)  is called the \textit{hyperedge potential} where each hyperedge \(h\) is mapped to \(\gamma_h \in G^{|h|}/\Delta\) where \(|h|\) is the number of nodes in the hyperedge.

The vertex potential and hyperedge potential of a hypergraph are \textit{compatible} when 
\begin{equation}\label{eq:compatiblecoset}
    \phi(h) = (\rho({i_1}), \dots,\rho({i_n}))\Delta,
\end{equation}
for all \(h = \{i_1, \dots, i_n\} \in H\).

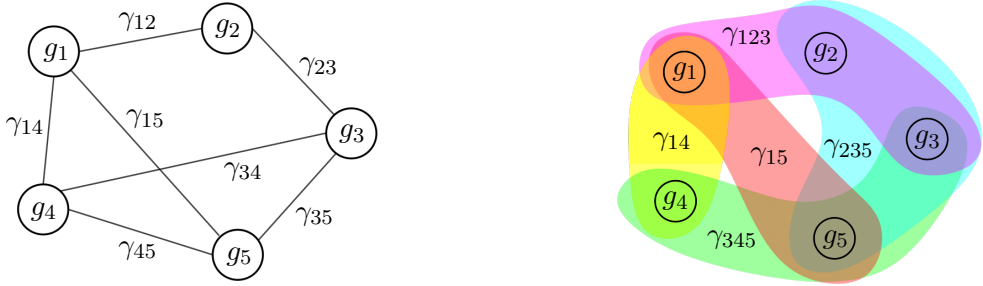
\begin{figure}[htb]
    \centering
    \begin{tikzpicture}
        \node[inner sep=0pt] (img1) at (-4,0) {\includegraphics[width=0.3\textwidth]{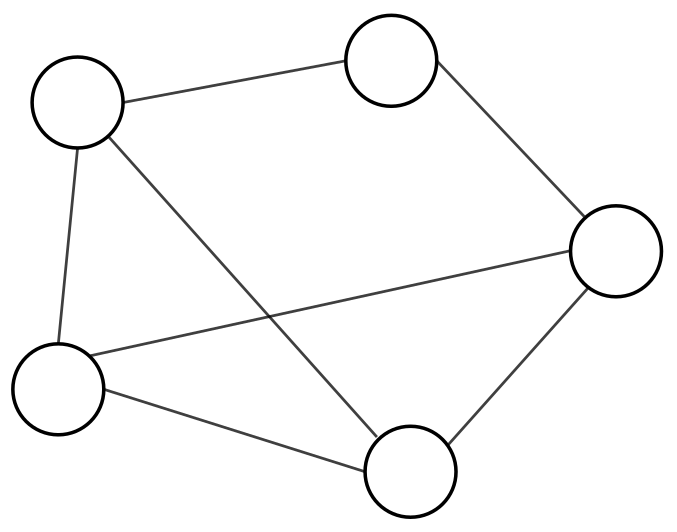}};
        \node at (-5.88,1.19) {\(g_1\)};
        \node at (-3.6,1.5) {\(g_2\)};
        \node at (-1.95,0.1) {\(g_3\)};
        \node at (-6.05,-0.93) {\(g_4\)};
        \node at (-3.45,-1.53) {\(g_5\)};
        \node at (-4.8,1.6) {\(\gamma_{12}\)};
        \node at (-2.4,1) {\(\gamma_{23}\)};
        \node at (-2.45,-1) {\(\gamma_{35}\)};
        \node at (-6.3,0.2) {\(\gamma_{14}\)};
        \node at (-4.7,0.3) {\(\gamma_{15}\)};
        \node at (-3.4,-0.4) {\(\gamma_{34}\)};
        \node at (-4.8,-1.45) {\(\gamma_{45}\)};

        \node[inner sep=0pt] (img2) at (4,0) {\includegraphics[width=0.3\textwidth]{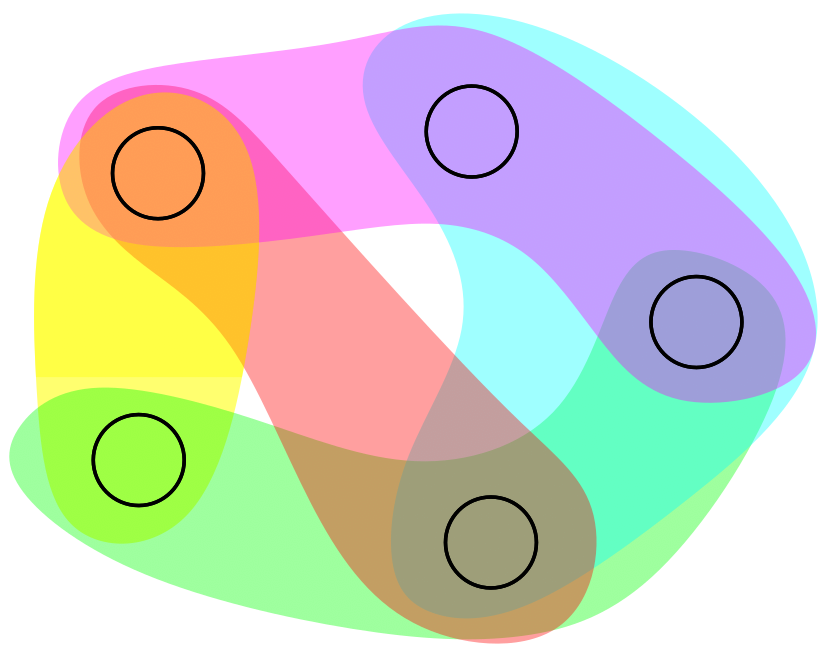}};
        \node at (2.48,0.93) {\(g_1\)};
        \node at (4.35,1.18) {\(g_2\)};
        \node at (5.7,0.02) {\(g_3\)};
        \node at (2.35,-0.79) {\(g_4\)};
        \node at (4.48,-1.3) {\(g_5\)};
        \node at (3.3,1.4) {\(\gamma_{123}\)};
        \node at (4.65,-0.1) {\(\gamma_{235}\)};
        \node at (3.1,-1.3) {\(\gamma_{345}\)};
        \node at (2.3,0) {\(\gamma_{14}\)};
        \node at (3.6,-0.2) {\(\gamma_{15}\)};
    \end{tikzpicture}
    \caption{\centering Illustration of the classical group synchronization setup (left) and the higher-order group synchronization setup (right).}
    \label{fig:HOS}
\end{figure}

If \(G\) is abelian then \(G^n/ \Delta\) is a group, however, for many practical applications, the group \(G\) (e.g. \(SO(2)\), \(SO(3)\), \(SE(3)\), etc) is not abelian, and so \(G^n/\Delta\) is a collection of right cosets \cite{herstein_abstract_1996}.
In this case, it is more convenient to consider a representation of the hyperedge measurements in the group \(G^{n-1}\). 
In fact, for any group \(G\), there is a non-canonical bijection (as sets), \(\tau: G^{n}/\Delta \to G^{n-1}\) given by
\begin{equation}\label{eq:bijection}
    \tau: (g_{i_1}, \dots g_{i_n})\Delta \longmapsto (g_{i_1}g_{i_2}^{-1}, \dots, g_{i_{n-1}}g_{i_n}^{-1}).
\end{equation}
Under the transformation \(\tau\), compatibility of vertex and hyperedge potentials is the condition that for all \(h = \{i_1, \dots, i_n\} \in H\),
\begin{equation}
    \tau \circ \phi(h) = (\rho(i_1)\rho(i_2)^{-1}, \dots, \rho(i_{n-1})\rho(i_n)^{-1}).
\end{equation}
Now we can formally define the problem of higher-order group synchronization.

\begin{problem}[Higher-Order Group Synchronization]
    Given a hyperedge potential \(\phi({\mathcal{H}}) = \{\gamma_h\}_{h \in H}\) on a hypergraph \(\mathcal{H}(V,H)\), find a vertex potential \(\rho({\mathcal{H}}) = \{g_i\}_{i \in V}\) such that \(\phi({\mathcal{H}})\) and \(\rho({\mathcal{H}})\) are compatible.
\end{problem}
As in the case of classical synchronization, compatible vertex potentials are unique up to an action of the group. 
Suppose \(\{g_i\}_{i \in V}\) is compatible with \(\{\gamma_h\}_{h \in H}\). 
Then for any \(g \in G\) and \(h = \{i_1, \dots, i_n\} \in H\), 
\begin{equation}
    \gamma_h =  (g_{i_1}, \dots, g_{i_n}) \Delta = (g_{i_1}\cdot g, \dots, g_{i_n}\cdot g) \Delta, 
\end{equation} 
so \(\{g_i\cdot g\}_{i \in V}\) is also a compatible vertex potential.

In an idealized setting with no noise or outliers, synchronization of the hyperedge potential is mathematically straightforward. Fix a vertex to be \(g_0\), then assign elements to the remaining vertices using the hyperedge measurements.
Suppose that \(h = \{i_1, \dots, i_n\} \in H\) so that 
\[
    \gamma_h = \left(g_{i_1}^{(h)}, \dots, g_{i_n}^{(h)}\right)\Delta.
\]
Then group elements can be assigned to the vertices by 
\begin{equation}\label{eq:iteratedassignment}
    g_{i_j} = g_{i_j}^{(h)} \left(g_{i_k}^{(h)}\right)^{-1} \cdot g_{i_k}.
\end{equation}

When noisy or corrupted hyperedge observations are introduced, we again pose a nonconvex minimization problem:
\begin{equation}\label{eq:higher-order-minimization}
    \min_{\{\rho(i)\}_{i \in V}} \sum_{h \in H} d_{G^{|h|-1}}(\tau\circ\phi(h),(\rho(i_1)\rho(i_2)^{-1}, \dots, \rho(i_{n-1})\rho(i_n)^{-1})),
\end{equation}
where \(d_{G^{|h| - 1}}\) is a chosen metric on \(G^{|h|-1}\). 
For example, when \(G\) is a matrix group, the Frobenius norm on \(G\) can be extended to a \(p\)-product metric on \(G^{|h|-1}\).
In this paper, we develop a global method to solve \eqref{eq:higher-order-minimization} that doesn't rely on iterated vertex assignments as in \eqref{eq:iteratedassignment} so that higher-order group synchronization can be useful in low signal-to-noise domains and robust to outliers.

\section{Higher-Order Synchronizability} \label{sec:synchronizability}
 
The first step is to understand how hyperedge and vertex potentials interact when \eqref{eq:higher-order-minimization} has a \(0\)-cost solution.
A hyperedge potential is \textit{synchronizable} if there exists a vertex potential satisfying the compatibility condition in \eqref{eq:compatiblecoset} and such a vertex potential is unique up to a global action. 
A natural question is to determine when a hyperedge potential is synchronizable. 
In classical pairwise group synchronization on a graph, the existence of a compatible vertex potential for a given edge potential is identified with a cycle consistency condition \cite{gao_geometry_2021,lerman_robust_2022}. 
\begin{theorem}[Lerman and Shi \cite{lerman_robust_2022}, Proposition 3]
    An edge potential is synchronizable if and only if all cycles are consistent.    
\end{theorem}

In Section \ref{subsec:hypergraph-cycles}, we introduce important hypergraph concepts that will allow us to state and analyze the compatibility conditions for higher-order synchronization. 
In Section \ref{subsec:synch-conditions} we state and prove a complete characterization of higher-order synchronizability.

\subsection{Hypergraph Cycles} \label{subsec:hypergraph-cycles}

For a hypergraph \(\mathcal{H}(V,H)\) there are many ways to define paths and cycles among the vertices and hyperedges \cite{berge_graphs_1973,gyarfas_odd_2006,jegou_notion_2009}. 
For higher-order synchronization, we adopt a definition that generalizes the concept of a Berge cycle on a hypergraph \cite{dorbec_monochromatic_2008}. 
The formal definitions are stated below.
\begin{definition}[\(k\)-Path]
    A \textbf{path} of order \(k\), or a \(k\)-path, on a set of \(l\) distinct vertices \(v_1, \dots, v_l \in V\) is a sequence of \(l-k\) hyperedges \(h_1, \dots, h_{l-k} \in H\) such that \(v_i,v_{i+1},\dots, v_{i+k} \in h_i\). 
    Such a \(k\)-path is denoted by \((v_1, \dots, v_l; h_1, \dots, h_{l-k})_k\).
\end{definition}

\begin{definition}[\(k\)-Cycle]
    A \textbf{cycle} of order \(k\), or a \(k\)-cycle, on a set of \(l\) vertices \(v_1, \dots, v_l \in V\) is a sequence of \(l-k\) hyperedges \(h_1, \dots, h_{l-k} \in H\) such that \(v_{1}, \dots, v_{l-k}\) are distinct, \(v_{l-k+j} = v_j\) for \(j = 1, \dots, k\), and \(v_i,v_{i+1},\dots, v_{i+k} \in h_i\) for all \(i = 1, \dots, l\) where addition on the indices is taken modulo \(l\). 
    A \(k\)-cycle is denoted by \([v_1, \dots, v_l; h_1, \dots, h_{l-k}]_k\).
\end{definition}

In the definitions above, \(h_i\) is allowed to have any number of vertices, not just those designated by the path or cycle, however \(|h|\) is required to be at least \(k\) for every hyperedge in a \(k\)-cycle or \(k\)-path. 
The set of vertices \(v_1, \dots, v_l\) in a cycle \([v_1, \dots, v_l; h_1, \dots, h_{l-k}]_k\) are called the \textit{base points} of the cycle \cite{gyarfas_odd_2006}. 
While the base points must be distinct (except for the first and last \(k\) base points) hyperedges are allowed to be repeated. 

\begin{figure}[htb]
    \centering    
    \includegraphics[width=0.3\textwidth]{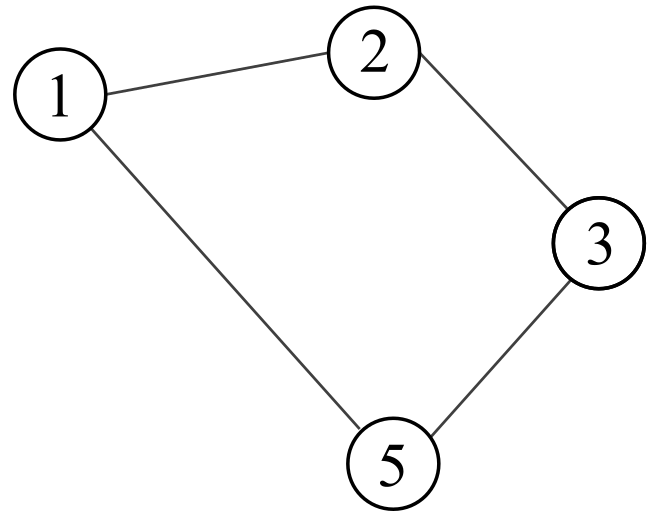}
    \includegraphics[width=0.3\textwidth]{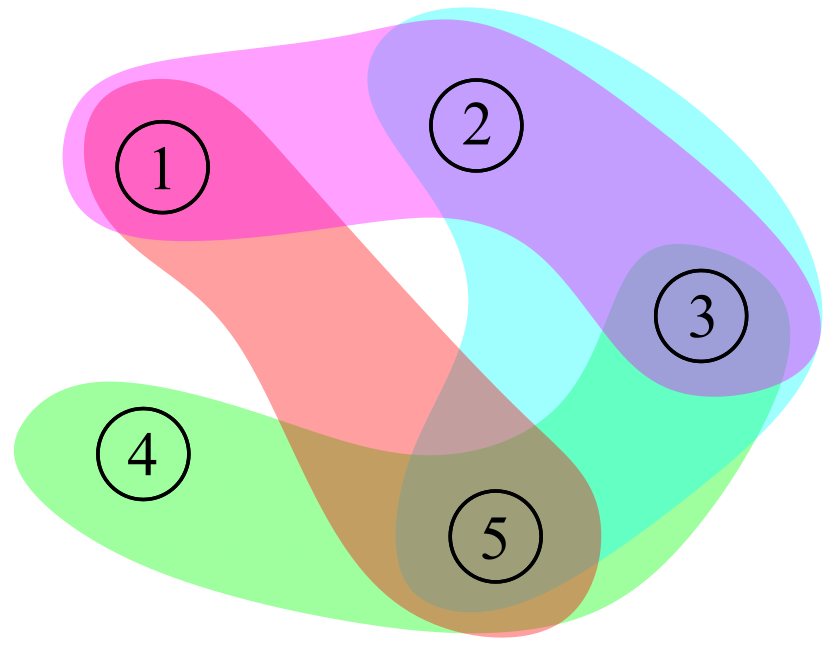}
    \includegraphics[width=0.3\textwidth]{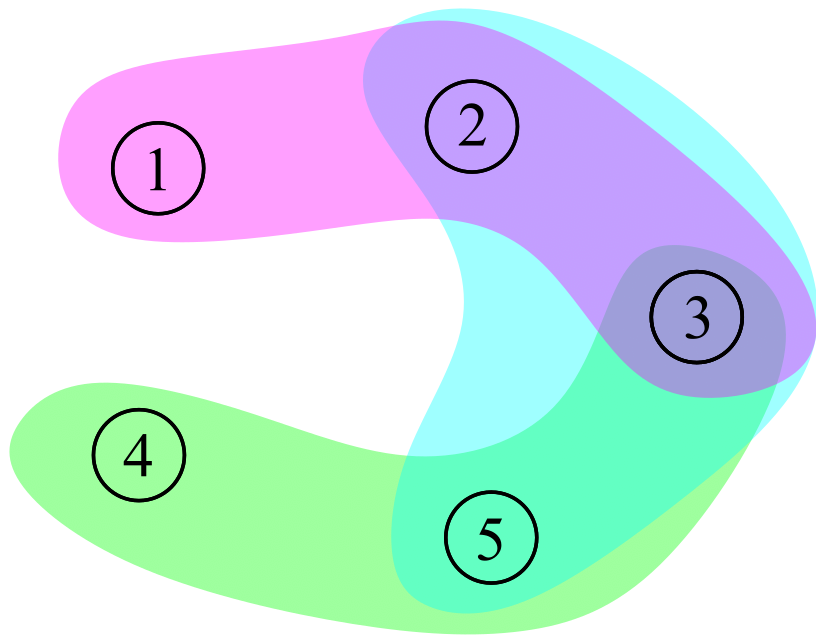}\\
    \caption{\centering Illustration of a \(1\)-cycle: \([1,2,3,5,1;12,23,35,15]_1\) (left), a \(1\)-cycle: \([1,2,3,5,1;123,235,345,15]_1\) (middle), and a \(2\)-path: \((1,2,3,5,4;123,235,345)_1\) (right).}
    \label{fig:HOS_cycles}
\end{figure}

A hypergraph is \textit{\(k\)-connected} if for every pair of vertices \(v_i\) and \(v_j\) there exists a \(k\)-path that starts at \(v_i\) and ends at \(v_j\). 
If a hypergraph is \(1\)-connected we simply call it a connected hypergraph. 

It is easy to show that if a hypergraph is \(k\)-connected, then there exists an \(i\)-path between any two vertices for all \(i \leq k\), and thus the hypergraph is also \(i\)-connected. 
Indeed, if 
\[
    (v_1, \dots, v_{l-1}, v_l; h_1, \dots, h_{l-k})_k
\]
is a \(k\)-path from \(v_1\) to \(v_l\), then 
\[
    (v_1, \dots, v_{l+i-k-2}, v_{l+i-k-1}, v_l; h_1, \dots, h_{l-k})_i
\]
is an \(i\)-path between the same vertices. 
For the remainder of this paper it is assumed that all hypergraphs are at least \(1\)-connected because any disconnected vertices can be discarded from the observation set to obtain a connected hypergraph.

We introduce restriction maps on the hyperedge measurements. 
Assume that the ordering of the vertices is fixed and let \(\bar{h}\) be a sub-hyperedge of the hyperedge \(h\), whose vertices are a subset of the vertices contained in \(h\). 
Then the image of the hyperedge potential at \(h\) can be restricted to assign a value to \(\phi(\bar{h})\) using the following canonical \textit{restriction map} from \(h\) to \(\bar{h}\):
\begin{equation}
    \operatorname{Res}_{h \to \bar{h}}: G^{|h|}/\Delta \to G^{|\bar{h}|}/\Delta.
\end{equation}
In particular, if two hyperedges \(h_i\) and \( h_j\) overlap on at least one vertex then there are restriction maps
\begin{equation}
    \operatorname{Res}_{h_i \to h_i \cap h_j}: G^{|h_i|}/\Delta \to G^{|h_i \cap h_j|}/\Delta
\end{equation}
and
\begin{equation}
    \operatorname{Res}_{h_j \to h_i \cap h_j}: G^{|h_j|}/\Delta \to G^{|h_i \cap h_j|}/\Delta,
\end{equation}
that compare their images under the hyperedge potential on just the overlapping vertices.

\subsection{Synchronization Condition} \label{subsec:synch-conditions}

Cycle consistency, as in classical synchronization, is a crucial ingredient for the existence of a compatible vertex potential. 
\begin{definition}[Consistent \(k\)-Cycle]
    A \(k\)-cycle, given by 
    \(C = [v_1, \dots, v_l; h_1, \dots, h_{l-k}]_k,\)
    is \textbf{consistent} if 
    \begin{equation}\label{eq:cycleconsistency}
        \Phi(C) := \prod_{i=1}^l \tau\circ \operatorname{Res}_{h_i \to \{v_i, v_{i+1},\dots, v_{i+k}\}}(\phi({h_i})) = 1,
    \end{equation}
    where \(\tau\) is the bijection from \(G^{k}/\Delta \to G^{k-1}\) defined in \eqref{eq:bijection} and \(1\) is implied to be the identity of \(G^{k-1}\).
\end{definition}
Now we state our main synchronizability result.
\begin{theorem} \label{thm:1cycleissynch}
    A hyperedge potential is synchronizable if and only if all \(1\)-cycles are consistent.
\end{theorem}

\begin{proof}
    Assume a connected hypergraph \(\mathcal{H}(V,H)\) and hyperedge potential \(\phi\) such that all \(1\)-cycles in \(\mathcal{H}\) are consistent. 
    To construct a compatible vertex potential \(\rho\), fix \(\rho(v_0) = 1 \) for a chosen vertex \(v_0\). 
    Since the hypergraph is connected, there exists a \(1\)-path from \(v_0\) to \(v_l\) for all \(v_l \in V\). 
    Denote this path by \(P = (v_0,v_1,\dots, v_l; h_0,h_1,\dots, h_{l-1})_1\) and assign the image of \(v_l\) under the vertex potential to be
    \begin{equation}\label{eq:vertex-potential-assignment}
        \rho(v_l):= \left(\prod_{i=0}^{l-1} \tau\circ \operatorname{Res}_{e_i \to \{v_i, v_{i+1}\}}(\phi({h_i})) \right)^{-1} \cdot \rho(v_0).
    \end{equation}
    To show that this assignment is well-defined, it suffices to show that it is path-independent. 
    Let \(P' =(v_0,v_1',\dots,v_{l-1}' v_l; h_0',h_1',\dots, h_{l-1}')_1\) be another path from \(v_0\) to \(v_l\). 
    If \(v_i \not= v_i'\) for all \(i = 1,\dots, l-1\) then the paths \(P\) and \(P'\) form a \(1\)-cycle: 
    \[
        C = [v_0, v_{1}, v_{2}, \dots, v_{l}, v_{l-1}',\dots, v_{2}', v_{1}', v_0; h_0, h_{1}, \dots, h_{l-1}, h_{l-1}', h_{l-2}',\dots, h_{1}', h_{0}']_1.
    \]
    Since all \(1\)-cycles are consistent
    \begin{equation}\label{eq:1-cycle-constistent}
        1 = \Phi(C) =\prod_{i=0}^{l-1} \tau\circ \operatorname{Res}_{h_i \to \{v_i, v_{i+1}\}}(\phi({h_i})) \prod_{i=l-1}^{0} \tau\circ \operatorname{Res}_{h_i' \to \{v_{i+1}',v_i'\}}(\phi({h_i'})).
    \end{equation}
    Equation \eqref{eq:1-cycle-constistent} implies
    \begin{align*}
        \prod_{i=0}^{l-1} \tau\circ \operatorname{Res}_{h_i \to \{v_i, v_{i+1}\}}(\phi({h_i})) &= \prod_{i=0}^{l-1} \left(\tau\circ \operatorname{Res}_{h_i' \to \{v_{i+1}',v_i'\}}(\phi({h_i'}))\right)^{-1} 
        \\
        &= \prod_{i=0}^{l-1} \tau\circ \operatorname{Res}_{h_i' \to \{v_{i}',v_{i+1}'\}}(\phi({h_i'})).
    \end{align*}
    Therefore \(\rho(v_l)\) is defined independent of the path chosen. 
    If \(v_i = v_i'\) for some \(i\) then there is a \(1\)-cycle that can be defined from \(v_0\) to \(v_i\) by 
    \[
        [v_0, v_{1}, v_{2}, \dots, v_{i}, v_{i-1}',\dots, v_{2}', v_{1}',v_0; h_0, h_{1}, \dots, h_{i}, h_{i-1}', h_{l-2}',\dots, h_{1}', h_{0}']_1,
    \]
    and a second \(1\)-cycle from \(v_i\) to \(v_l\) by 
    \[
        [v_i, v_{i+1},\dots, v_{l}, v_{l-1}',\dots, v_{i+1}', v_{i}',v_i; h_i, h_{i+1}, \dots, h_{l}, h_{l-1}', h_{l-2}',\dots, h_{i+1}', h_{i}']_1.
    \]
    By the same argument above, \(\rho(v_i)\) is defined independent of the path from \(v_0\) to \(v_i\) and \(\rho(v_l)\) is defined independent of the path from \(v_i\) to \(v_l\) so together there is a path independent assignment for \(\rho(v_l)\).

    Now to show that \(\rho\) and \(\phi\) are compatible, let \(h = \{v_{i_1}, \dots, v_{i_l}\} \in H\) and suppose \(\phi(h) = (g_{i_1}, \dots, g_{i_l})\Delta\). 
    Since \(\rho(v_0)\) is known, \(\rho(v_{i_1})\) can be determined. 
    Using path independence, for each \(j = 2, \dots, l\), an assignment for \(\rho(v_{i_j})\) can be determined from the path 
    \[
        (v_{i_1}, v_{i_2}, \dots, v_{i_j}; h, h, \dots, h)_1
    \]
    so that 
    \begin{align*}
        \rho(v_{i_j}) &= \left(\prod_{k=1}^{j-1} \tau\circ \operatorname{Res}_{h \to \{v_{i_k}, v_{i_{k+1}}\}}(\phi({h})) \right)^{-1} \cdot \rho(v_{i_1}) 
        \\
        &= \left(\prod_{k=1}^{j-1} (g_{i_k}g_{i_{k+1}}^{-1}) \right)^{-1} \cdot \rho(v_{i_1}) 
        = \left( g_{i_1}g_{i_j}^{-1} \right)^{-1} \cdot \rho(v_{i_1})
        = g_{i_j}g_{i_1}^{-1} \cdot \rho(v_{i_1}).
    \end{align*}
    Then indeed \(\rho\) and \(\phi\) are compatible:
    \[
       (\rho(v_{i_1}), \dots, \rho(v_{i_l}))\Delta  = (\rho(v_{i_1}), g_{i_2}g_{i_1}^{-1}\rho(v_{i_1}), \dots, g_{i_l}g_{i_1}^{-1}\rho(v_{i_1}))\Delta = (g_{i_1}, g_{i_2}, \dots, g_{i_l})\Delta.
    \]
    
    Finally, we show that this vertex potential is unique up to a global action. 
    Suppose towards a contradiction that there exists another vertex potential \(\rho'\) that is compatible with \(\phi\) and that there is no \(g \in G\) such that \(\rho(v) = \rho'(v) \cdot g\) for all \(g \in G\). 
    In other words there exists \(g_1 \not= g_2 \in G\) and \(v_1,v_2 \in V\) such that \(\rho(v_1) =  \rho'(v_1) \cdot g_1\) and \(\rho(v_2) = \rho'(v_2) \cdot g_2\). 

    First consider the case when \(v_1\) and \(v_2\) both belong to the same hyperedge \(h \in H\). 
    Suppose with out loss of generality \(h = \{v_1,v_2, v_{i_1}, \dots, v_{i_l}\}\). 
    Then since \(\rho\) and \(\rho'\) are both compatible,
    \begin{equation}\label{eq:compatiblepotentials}
        (\rho(v_{1}),\rho(v_{2}),\rho(v_{i_1}), \dots, \rho(v_{i_l}))\Delta = \phi(h) =  (\rho'(v_{1}),\rho'(v_{2}),\rho'(v_{i_1}), \dots, \rho'(v_{i_l}))\Delta .
    \end{equation}
    Equation \eqref{eq:compatiblepotentials} implies that there exists \(g \in G\) such that \(\rho(v_1) =  \rho'(v_1) \cdot g\) and \(\rho(v_2) = \rho'(v_2) \cdot g\) which contradicts the assumption. 

    Now consider the case when there is no \(h \in H\) containing both \(v_1\) and \(v_2\).
    Since \(\mathcal{H}\) is connected, there exists a \(1\)-path between \(v_1\) and \(v_2\), say \(P = (v_1, v_{i_1}, \dots, v_{i_l},v_2; h_{j_1},h_{j_2}, \dots, h_{j_{l+1}})_1\).
    Using the same argument as in equation \ref{eq:compatiblepotentials}, the compatibility of \(\rho\) and \(\rho'\) implies that there exists \(g \in G\) such that \(\rho(v_1) = \rho'(v_1) \cdot g\) and \(\rho(v_{i_1}) = \rho'(v_{i_1}) \cdot g\). 
    Similarly, there is an element \(g' \in G\) such that \(\rho(v_{i_1}) = \rho'(v_{i_1}) \cdot g'\) and \(\rho(v_{i_2}) = \rho'(v_{i_2}) \cdot g'\). 
    But since \(\rho(v_{i_1}) = \rho'(v_{i_1}) \cdot g\) and \(\rho(v_{i_1}) = \rho'(v_{i_1}) \cdot g'\), \(g = g'\). 
    Applying this argument along the entire path \(P\) implies that \(\rho(v_{2}) = \rho'(v_{2}) \cdot g\), again giving a contradiction to the assumption. 
    Thus any for any two vertex potentials compatible with \(\phi\) differ only by an action of \(G\). 
 
    Conversely, assume a hypergraph \(\mathcal{H}(V,H)\) with an observed hyperedge potential \(\phi\) and a compatible vertex potential \(\rho\). 
    Since \(\rho\) and \(\phi\) are compatible, for any \(h \in H\), where \(h = \{v_1,\dots, v_{|h|}\}\),  
    \[
        \phi(h) = (\rho(v_1), \dots, \rho(v_{|h|}) )\Delta.
    \]
    Then any \(1\)-cycle, \(C = [v_1,\dots, v_l; h_1,\dots, h_{l-1}]_1\), can be shown to be consistent:
    \[
        \Phi(C) = \prod_{i=1}^{l-1} \tau ((\rho(v_i),\rho(v_{i+1}))\Delta) = \prod_{i=1}^{l-1} (\rho(v_i)\rho(v_{i+1})^{-1}, \dots, \rho(v_{i})\rho(v_{i+1})^{-1}) = 1. \qedhere
    \]
\end{proof}

Theorem \ref{thm:1cycleissynch} provides a complete answer to the existence and construction of a compatible vertex potential for a higher-order group synchronization problem on a general hypergraph. 
It is not hard to see that consistency of \(1\)-cycles implies the consistency of higher-order cycles. 
One might consider a different synchronizability condition that depends on the consistency of higher-order cycles. 
We leave this for future work. 
However, in the next section, Theorem \ref{thm:1cycleissynch} is used to build a message passing framework which depends on the consistency of higher-order cycles. 
In that specific setting it is shown that with assumptions on the structure of the hypergraph and hyperedge potential, synchronizability can be achieved by considering a special set of higher-order cycles.

\section{Cycle-Hyperedge Message Passing} \label{sec:CHMP}

Now that the role of cycle consistency in higher-order group synchronization has been established, the focus shifts to adopting the cycle consistency based message passing method of \cite{lerman_robust_2022} to the higher-order setup. 
In Section \ref{subsec:preliminaries}, the theory from Section \ref{sec:synchronizability} is used to motivate a framework for synchronizing a hyperedge potential by estimating the hyperedge corruption levels.  
Then in Section \ref{subsec:CHMPalgorithm} the specific details and steps of our proposed algorithm are stated. 
For simplicity, only \(n\)-uniform hypergraphs (\(n\)-hypergraphs) are considered but the framework can be extended to consider other types of hypergraphs which is discussed in Section \ref{subsec:general}. 

\subsection{Using Cycle Consistency to Estimate Hyperedge Corruption} \label{subsec:preliminaries}

Recall the setup for the higher-order synchronization problem outlined in Section \ref{subsec:HOGS-setup}. 
Fix an \(n \geq 2\) and for each hyperedge so that our hypergraph is \(n\)-uniform and measure \(\overline{\gamma}_h \in G^{n-1}\) where \(G\) is some compact group. 
Here the observation \(\overline{\gamma}_h\) comes from applying \(\tau : G^{n}/\Delta \to G^{n-1}\) in \eqref{eq:bijection} to the hyperedge potential, i.e. \(\overline{\gamma}_h = \tau(\gamma_h)\). 
Let \(\{g_i^*\}_{i \in V}\) be the underlying ground truth vertex potential. 
Since the measurements \(\{\overline{\gamma}_h\}_{h \in H}\) may be noisy or corrupted, the ground truth hyperedge measurement for \(h = \{i_1,\dots, i_n\}\) is denoted by
\begin{equation}
    \overline{\gamma}_h^* = (g_{i_1}^*(g_{i_2}^*)^{-1},\dots, g_{i_{n-1}}^*(g_{i_n}^*)^{-1}).
\end{equation}
The set of ground truth hyperedge measurements \(\{\overline{\gamma}_h^*\}_{h \in H}\) are compatible with the vertex potential \(\{g_i^*\}_{i \in V}\). 
The goal of the Cycle-Hyperedge Message Passing method will be to estimate how much the observed measurements \(\overline{\gamma}_h\) differ from this ground truth.

Assume a metric on \(G\) given by \(d_G(\cdot,\cdot)\). 
Further, assume that this metric is bi-invariant:
\begin{equation}
    d_G(g_1,g_2) = d_G(g_1\cdot g, g_2 \cdot g) = d_G(g \cdot g_1, g \cdot g_2) \text{ for } g,g_1,g_2 \in G.
\end{equation}
The metric \(d_G\) is extended to a metric on \(G^{n-1}\) by some suitable product metric which preserves bi-invariance. 
Due to the compactness of \(G\), \(d_G\) and \(d_{G^{n-1}}\) are chosen so that \(d_{G^{n-1}} \leq 1\). 
For notational convenience, \(d_{n-1}(\cdot,\cdot)\) will refer to the metric \(d_{G^{n-1}}\) unless otherwise noted. 
In Section \ref{sec:numerical-results}, some examples of metrics for different groups are given.

To motivate the algorithm design, the adversarial noise model of \cite{lerman_robust_2022} is adopted, though our framework is applicable irrespective of the noise model. 
The motivating noise model is as follows: first, the hyperedges of \(\mathcal{H}\) are partitioned into two sets. 
The set \(H_g\), or the ``good", hyperedges are considered uncorrupted, but possibly noisy. 
For hyperedges in \(H_g\), \(\overline{\gamma}_h^*\) is replaced by \(\overline{\gamma}_h^*\cdot \overline{\gamma}_h^{\epsilon}\) where \(\overline{\gamma}_h^{\epsilon}\) is an \((n\!-\!1)\)-tuple of i.i.d. \(G\)-valued random variables such that \(d_{{n-1}}(\overline{\gamma}_h^{\epsilon}, 1)\) is sub-gaussian with variance proxy \(\sigma^2\). 
The noiseless case is when \(\overline{\gamma}_h^{\epsilon} = 1\) for all \(h \in H_g\); otherwise, we say the model is noisy. 
Corrupted, or ``bad", hyperedges form the set \(H_b\), corresponding to outliers. For \(\overline{\gamma}_h\) where \(h \in H_b\), \(\overline{\gamma}_h^*\) is replaced by \(\widetilde{{\gamma}}_h \in G^{n-1}\) where each entry of \(\widetilde{\gamma}_h\) is drawn independently from a distribution on \(G\). 
Together the sets \(H_g\) and \(H_b\) form the noise model:
\begin{equation}\label{eq:noisemodel}
    \overline{\gamma}_h = 
    \begin{cases}
        \overline{\gamma}_h^*\overline{\gamma}_h^{\epsilon}, & h \in H_g
        \\
        \widetilde{\gamma}_h, & h \in H_b.
    \end{cases}
\end{equation}

This noise model is natural to the higher-order setup and is distinct from the noise model generated by constructing higher-order measurements from lower order structures such as using triangles in a graph to form a \(3\)-hypergraph. 
As in classical synchronization, exact recovery (up to an action of \(G\)) is only possible in the noiseless case, otherwise the best we can hope for is approximate recovery of the vertex potential.

The goal of the algorithm is to estimate the hyperedge corruption levels.
\begin{definition}[Hyperedge Corruption Level]
    For a hyperedge \(h\), the \textbf{corruption level} is given by
    \begin{equation}
        s_h^* := d_{n-1}(\overline{\gamma}_h,\overline{\gamma}_h^*).
    \end{equation}
\end{definition}
While the ultimate goal of higher-order synchronization is to recover the vertex potential, Proposition \ref{prop:exact-recovery} below shows that solving for the hyperedge corruption levels is sufficient. 
In Section \ref{sec:recovery} methods for recovering or estimating the vertex potential using the hyperedge corruption levels are discussed.
\begin{proposition} \label{prop:exact-recovery}
    Suppose \(\{\overline{\gamma}_h\}_{h \in H}\) is generated by the noiseless corruption model given by \eqref{eq:noisemodel} on a hypergraph \(\mathcal{H}(V,H)\) such that \(\mathcal{H}(V,H_g)\) is connected.  
    Then, exact estimation of \(\{s^*_{h}\}_{{h} \in H}\) is equivalent to exact recovery of \(\{g_i^*\}_{i \in V}\) (up to a global action).
\end{proposition}

\begin{proof}
    Assume \(\{\overline{\gamma}_{h}\}_{{h} \in H}\) is observed and \(\{\overline{\gamma}_{h}^*\}_{{h} \in H}\) has been exactly recovered, then 
    \[
        s^*_{h} = d_{n-1}(\overline{\gamma}_{h},\overline{\gamma}_{h}^*)
    \]
    can be calculated for all \({h} \in H\).

    Conversely, assume that \(s^*_{h}\) is known for all \({h} \in H\), then \(H_g \subseteq H_g' = \{{h} \in H : s^*_{h} = 0\}\). 
    Since \(H_g\) is connected, \(H_g'\) is connected For every hyperedge in \(H_g'\), \(\overline{\gamma}_h = \overline{\gamma}_h^*\) which implies that all \(1\)-cycles in \(\mathcal{H}(V,H_g')\) are consistent. 
    Thus by Theorem \ref{thm:1cycleissynch}, \(\{g_i^*\}_{i \in V}\) can be recovered up to a global action.
\end{proof}

Now recall the definition of cycle consistency from \eqref{eq:cycleconsistency}. 
Clearly, in the noisy case, cycle consistency is unlikely to hold for any cycle, but the failure of cycle consistency for a cycle \(C\) can be measured which motivates the following definition.
\begin{definition}[Cycle Consistency Measure]
    For a \(k\)-cycle, \(C\), the \textbf{cycle consistency measure} is
    \begin{equation}\label{eq:cycleinconsistency}
        d_C := d_{k}(\Phi(C),1).
    \end{equation}
\end{definition}
Our framework will only consider the consistency of \((n\!-\!1)\)-cycles on \(n\)-hypergraphs.
This choice of cycle type is justified in Section \ref{subsec:general}. 
Denote the set of \((n\!-\!1)\)-cycles on a hypergraph by \(\mathcal{C}^{n-1}\). 
If \(\overline{\gamma}_h \in G^{n-1}\) for all \(h \in H\) and \(C \in \mathcal{C}^{n-1}\) then \(\Phi(C)\) can be simplified to:
\begin{equation}
    \Phi(C) = \overline{\gamma}_{h_1}\overline{\gamma}_{h_2} \cdots \overline{\gamma}_{h_l}.
\end{equation}
where \(h_1, \dots, h_l\) are the hyperedges in \(C\).

The motivation for using cycle consistency to recover the hyperedge corruption levels comes from the idea of a ``good cycle". 
Let \(N_C:= \{h \in H : h \in C\}\) and assume a noiseless setting. 
If \(C \in \mathcal{C}^{n-1}\), then for any \(h\) in \(N_C\), if every hyperedge in \(N_C \setminus \{h\}\) is a good hyperedge, then 
\begin{equation}\label{eq:goodcycle}
    s_h^* = d_C.
\end{equation} 
In other words, if there is at most one bad hyperedge in a cycle, the cycle consistency measure of the cycle gives the corruption level for that hyperedge.
This can be seen by using the bi-invariance of the metric:
\begin{equation}\label{eq:good-cycle-proof}
    d_C = d_{n-1}(\Phi(C),1) = d_{n-1}(\overline{\gamma}_h\overline{\gamma}_{h_1}^*\cdots \overline{\gamma}_{h_l}^*,1) = d_{n-1}(\overline{\gamma}_h\overline{\gamma}_{h_1}^*\cdots \overline{\gamma}_{h_l}^*\overline{\gamma}_h^*,\overline{\gamma}_h^*) = d_{n-1}(\overline{\gamma}_h,\overline{\gamma}_h^*) = s_h^*
\end{equation}
where the second to last equality comes from the fact that the ground truth hyperedge measurements are compatible with the underlying vertex potential.
Cycles where \eqref{eq:goodcycle} holds for \(h \in H\) are considered \textit{good cycles} with respect to \(h\), otherwise they are \textit{bad cycles} with respect to \(h\). The following definition is an extension from \cite{lerman_robust_2022}.
\begin{definition}[The Good Cycle Condition]
    A \(n\)-hypergraph, hyperedge potential, and set of cycles \(\mathcal{C} \subseteq \mathcal{C}^{n-1}\) satisfy the \textbf{Good Cycle Condition} if for every \(h \in H\) there exists a cycle \(C \in \mathcal{C}\) such that 
    \begin{equation}\label{eq:good-cycle-condition}
        N_C \setminus \{h\} \subseteq H_g,
    \end{equation}
\end{definition}

The following proposition formalizes the relationship between good cycles and exact recovery of the hyperedge corruption levels.
\begin{proposition}\label{prop:corruptionisinconsitency} 
    Suppose \(\{\overline{\gamma}_h\}_{h \in H}\) is generated by the noiseless corruption model in \eqref{eq:noisemodel} on a \(n\)-hypergraph \(\mathcal{H}(V,H)\). 
    Fix a set of cycles, \(\mathcal{C} \subseteq \mathcal{C}^{n-1}\). 
    Then \(s_h^* = d_C\) for every pair \(h \in H\) and \(C \in \mathcal{C}\) that satisfy \eqref{eq:good-cycle-condition}. 
\end{proposition}

Proposition \ref{prop:corruptionisinconsitency} is proven by a direct application of \eqref{eq:good-cycle-proof} to every hyperedge of \(\mathcal{H}\). 
Together with Proposition \ref{prop:exact-recovery}, Proposition \ref{prop:corruptionisinconsitency} implies the following corollary. 

\begin{corollary} \label{cor:GCCissynch}
    Suppose \(\{\overline{\gamma}_h\}_{h \in H}\) is generated by the noiseless corruption model in \eqref{eq:noisemodel} on a \(n\)-hypergraph \(\mathcal{H}(V,H)\). 
    If a set of cycles \(\mathcal{C} \subseteq \mathcal{C}^{n-1}\) on the hypergraph satisfy the Good Cycle Condition, then the data is synchronizable.
\end{corollary}

Corollary \ref{cor:GCCissynch} gives s sufficient condition for the synchronizability of a noiseless hyperedge potential (that may potentially be corrupted). 
However, in practice, we rarely observe a noiseless hyperedge potential. 
The following proposition shows that even for bad cycles, cycle consistency is sill a useful way to estimate hyperedge corruption. 

\begin{proposition} \label{prop:corruptionbound}
    Suppose \(\{\overline{\gamma}_h\}_{h \in H}\) is generated by the noiseless corruption model in \eqref{eq:noisemodel} on a \(n\)-hypergraph \(\mathcal{H}(V,H)\).
    For all \({h} \in H\) and any cycle \(C \in \mathcal{C}^{n-1}\) containing \({h}\),
    \begin{equation}
        |d_C - s^*_{h}| \leq \sum_{{h'} \in N_C \setminus {h}}s^*_{h'}.
    \end{equation}  
\end{proposition}

\begin{proof}
Suppose without loss of generality \(C =  \{h_0,h_1,\dots, h_l\}\). Using the triangle inequality and bi-invariance, it follows that:
\begin{align} \label{eq:expanddifference}
    |d_C - s^*_{h_0}| &= |d_{n-1}(\overline{\gamma}_{h_0} \overline{\gamma}_{h_1} \cdots \overline{\gamma}_{h_l},1) - d_{n-1}(\overline{\gamma}_{h_0}\overline{\gamma}_{h_0}^{*-1},1)| \nonumber
    \\ 
    &\leq d_{n-1}(\overline{\gamma}_{h_0} \overline{\gamma}_{h_1} \cdots \overline{\gamma}_{h_l},\overline{\gamma}_{h_0}\overline{\gamma}_{h_0}^{*-1}) = d_{n-1}(\overline{\gamma}_{h_0}^* \overline{\gamma}_{h_1} \cdots \overline{\gamma}_{h_l},1).
\end{align}
Further, for any \(i = 0,\dots, l\), \(s_{h_i}^*\) can be expanded so that:
\begin{equation}\label{eq:expand-s^*}
    s^*_{h_i} = d_{n-1}(\overline{\gamma}_{h_i},\overline{\gamma}^*_{h_i}) = d_{n-1}(\overline{\gamma}^*_{h_0} \cdots \overline{\gamma}^*_{h_{i-1}} \overline{\gamma}_{h_i} \cdots \overline{\gamma}_{h_l},\overline{\gamma}^*_{h_0} \cdots \overline{\gamma}^*_{h_i} \overline{\gamma}_{h_{i+1}} \cdots \overline{\gamma}_{h_l}).
\end{equation}

Finally, combining \eqref{eq:expanddifference} and \eqref{eq:expand-s^*}, it can be shown that
\begin{align*}
    d_{n-1}(\overline{\gamma}_{h_0}^* \overline{\gamma}_{h_1} \cdots \overline{\gamma}_{h_l},1) =& d_{n-1}(\overline{\gamma}_{h_0}^* \overline{\gamma}_{h_1} \cdots \overline{\gamma}_{h_l},g^*_{h_0} \overline{\gamma}_{h_1}^* \cdots \overline{\gamma}_{h_l}^*) 
    \\
    \leq& d_{n-1}(\overline{\gamma}_{h_0}^* \overline{\gamma}_{h_1} \cdots \overline{\gamma}_{h_l},g^*_{h_0} \overline{\gamma}_{h_1}^* \overline{\gamma}_{h_2} \cdots \overline{\gamma}_{h_l}) 
    \\
    &+ d_{n-1}(g^*_{h_0} \overline{\gamma}_{h_1}^* \overline{\gamma}_{h_2} \cdots \overline{\gamma}_{h_l},g^*_{h_0} \overline{\gamma}_{h_1}^* \cdots \overline{\gamma}_{h_l}^*) 
    \\
    \leq& \sum_{k=1}^l d_{n-1}(\overline{\gamma}_{h_0}^* \cdots \overline{\gamma}_{h_{i-1}}^* \overline{\gamma}_{h_i} \cdots \overline{\gamma}_{h_l}, g^*_{h_0} \cdots \overline{\gamma}_{h_i}^* \overline{\gamma}_{h_{i+1}} \cdots \overline{\gamma}_{h_l}) = \sum_{n=1}^l s^*_{h_i}. \qedhere
\end{align*}
\end{proof}

\subsection{The Cycle-Hyperedge Message Passing Algorithm} \label{subsec:CHMPalgorithm}

Before the main algorithm is stated, there are a few more preliminaries to discuss. 
The main inputs for Algorithm \ref{alg:CHMP} are a \(n\)-hypergraph \(\mathcal{H}(V,H)\), hyperedge potential \(\{\overline{\gamma}_h\}_{h \in H}\), and cycle set \(\mathcal{C} \subseteq \mathcal{C}^{n-1}\). 
The algorithm is initialized by first constructing a bi-partite cycle-hyperedge graph, \textbf{CHG}. 
On one side of the graph are nodes labeled by the cycles \(C \in \mathcal{C}\) and on the other side are nodes labeled by the hyperedges \(h \in H\). 
Edges in CHG are placed wherever \(h \in N_C\). 
For notational convenience, define the following sets for each hyperedge \(h \in H\) that are the set of neighboring cycles, good cycles, and bad cycles respectively:
\begin{align}
    N_h &:= \{C \in \mathcal{C} : h \in N_C\}
    \\
    G_h &:= \{C \in N_h : N_C \setminus \{h\} \subseteq H_g\}
    \\
    B_h &:= N_h \setminus G_h.
\end{align}
Each of these sets induce a set on the nodes of CHG. 
Figure \ref{fig:CHMPschematic} below illustrates CHG.

The algorithm will make use of a reweighting function \(f\), with increasing parameters \(\beta_t\). 
The function \(f\) is chosen to be \(f(x;\beta_t)= \exp({-\beta_t x})\), though any nonincreasing reweighting function could be substituted. 
The increasing parameters are chosen to be \(\beta_t = \beta_0r^t\) for some rate \(r > 1\) and initial parameter \(\beta_0\), though again, any increasing parameters could be substituted. 
Further, the algorithm can be generally stated for any selected subset of cycles \(\mathcal{C} \subseteq \mathcal{C}^{n-1}\) but we choose \(\mathcal{C}\) to be the set of \((n\!-\!1)\)-cycles in \(\mathcal{H}\) that are of length \(n + 1\), and this choice is denoted by \(\mathcal{C}_{n+1}^{n-1}\). 
Finally, at each iteration \(t\), let \(s_h(t)\) denote the current estimate for \(s_h^*\). 
Our method is stated in Algorithm \ref{alg:CHMP}.

\begin{algorithm}[H]
    \caption{Cycle-Hyperedge Message Passing (CHMP)}\label{alg:CHMP}
    \begin{algorithmic}
        \REQUIRE \(n\)-hypergraph \(\mathcal{H}(V,H)\), observed measurements \(\{\overline{\gamma}_{h}\}_{{h} \in H}\), cycle set \(\mathcal{C}\), number of iterations \(T\), increasing parameters \(\{\beta_t\}_{t=0}^T\).
        \\
        \STATE \textbf{Steps:}
        \STATE Generate CHG from \(\mathcal{H}(V,H)\) and \(\mathcal{C}\)
        \FORALL {\(h \in H\) and \(C\in N_{h}\)}
            \STATE 
                \begin{equation}
                    d_C = d_{n-1}(\Phi(C), 1) 
                \end{equation}
            \ENDFOR
        \FORALL {\(h \in H\)}
            \STATE  
                \begin{equation}
                    s_{h}(0) = \frac{1}{|N_{h}|} \sum_{C\in N_{h}} d_C
                \end{equation}
            \ENDFOR
        \FOR {\(t=0, \dots, T-1\)}
            \FORALL {\(h \in H\) and \(C\in N_{h}\)}
                \STATE
                    \begin{equation}\label{eq:weightupdate}
                        w_{h,C}(t) = \frac{1}{Z_{h}(t)}\prod_{h'\in N_C\setminus \{h\}}f(s_{h'}(t);\beta_t)
                    \end{equation}
                    where
                    \begin{equation}\label{eq:normalizationfactor}
                    Z_{h}(t)=\sum_{C\in N_{h}} \prod_{h'\in N_C\setminus \{h\}}f(s_{h'}(t);\beta_t)
                    \end{equation}
                \ENDFOR
            \FORALL {\(h \in H\)}
                \STATE
                    \begin{equation}\label{eq:corruptionupdate}
                        s_{h}(t+1)=\sum_{C\in N_{h}} w_{h,C}(t)d_C
                    \end{equation}
            \ENDFOR
        \ENDFOR
        \ENSURE \(\{s_{h}(T)\}_{h \in H}\)
    \end{algorithmic}
\end{algorithm}

\begin{figure}[htb]
    \centering
    \begin{tikzpicture}
        \node[inner sep=0pt] (img) at (0,0) {\includegraphics[width=0.6\textwidth]{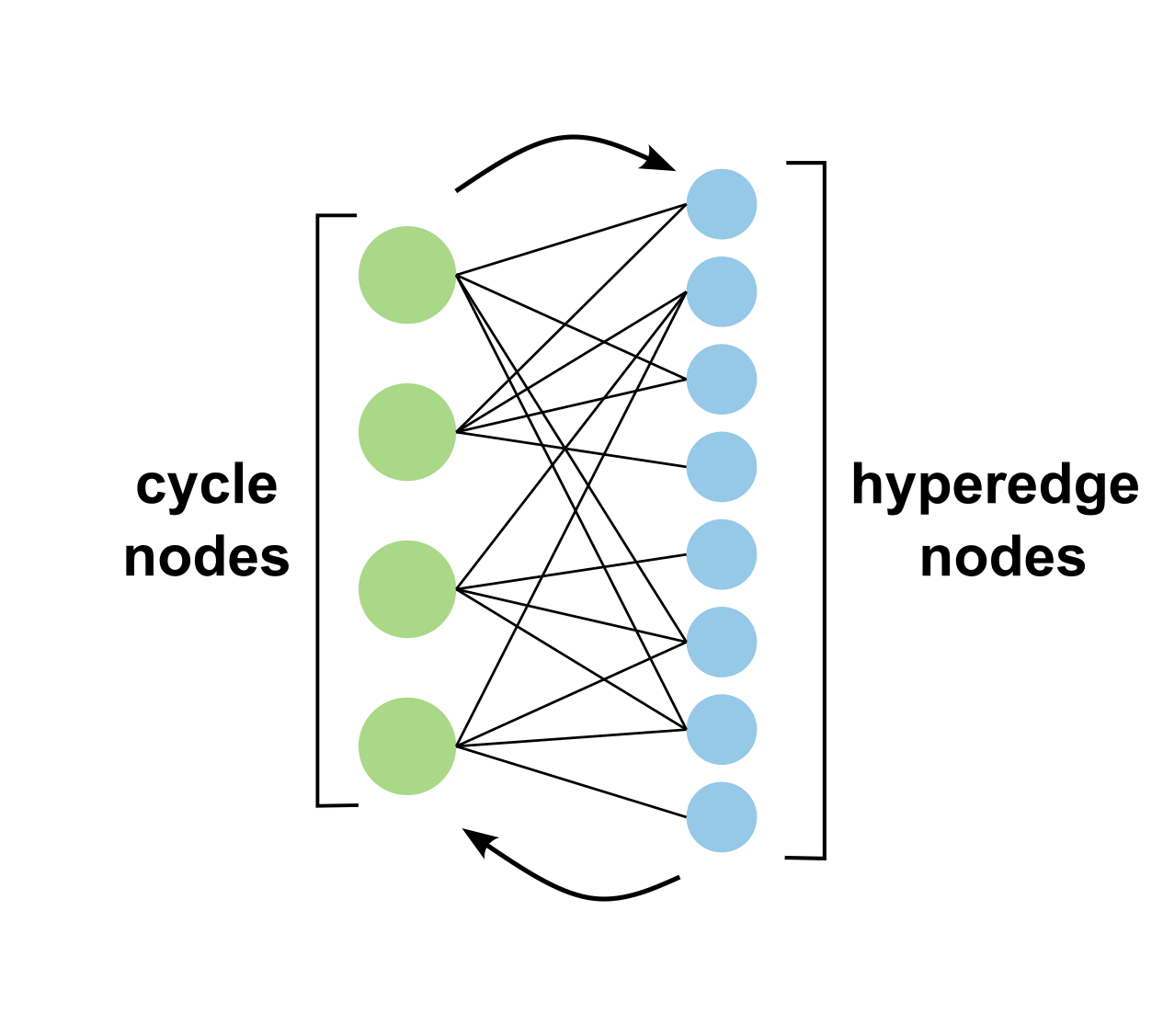}};
        \node at (0,3.6) {\eqref{eq:corruptionupdate}};
        \node at (0,-3.6) {\eqref{eq:weightupdate} and \eqref{eq:normalizationfactor}};
        \node at (1.2,-1.79) {\tiny \(s_h(t)\)};
        \node at (0.2,-2.02) {\tiny \(w_{h,C}(t)\)};
        \node at (-1.45,-1.95) {\(d_C\)};
    \end{tikzpicture}
    \caption{\centering Illustration of CHG. 
        The cycle nodes are on the left (green) and the hyperedge nodes are on the right (blue).
        In CHMP (Algorithm \ref{alg:CHMP}), messages are passed from cycles to hyperedges by \eqref{eq:corruptionupdate} and messages are passed from hyperedges to cycles by \eqref{eq:weightupdate} and \eqref{eq:normalizationfactor}}
    \label{fig:CHMPschematic}
\end{figure}
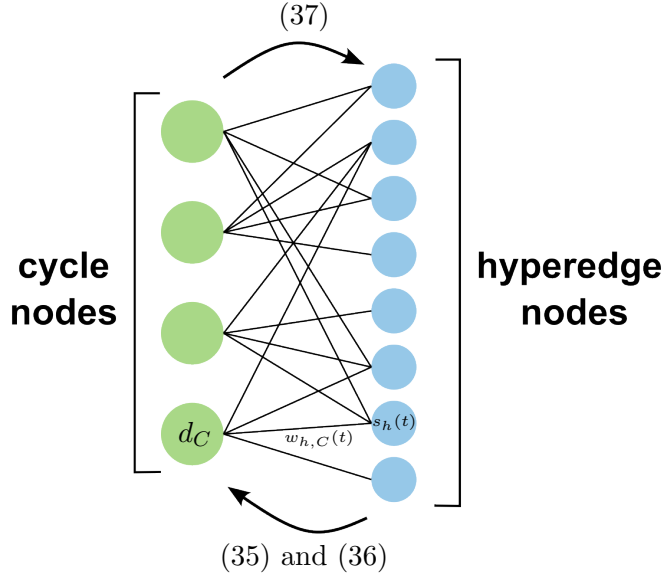

Figure \ref{fig:CHMPschematic} illustrates CHMP as a message passing procedure on CHG. 
The message passing interpretation can be described by interpreting the weights \(\{w_{h,C}(t)\}_{h \in H, C \in N_h}\) given in \eqref{eq:weightupdate} and \eqref{eq:normalizationfactor} as estimates for how likely \(C\) is a good cycle with respect to \(h\). 
For example, a small value for \(w_{h,C}(t)\) indicates that \(C\) is not likely a good cycle with respect to \(h\) and that there is likely at least one other bad hyperedge in \(C\). 
Conversely, a large value for \(w_{h,C}(t)\) indicates that \(C\) is likely a good cycle with respect to \(h\). 
This information about the cycles is passed to the hyperedges through \eqref{eq:corruptionupdate} and is used to estimate the corruption levels of the hyperedges. 
In \eqref{eq:corruptionupdate}, the cycles that are more likely to be good are given more influence in the estimation of \(s_h(t)\). 

The corruption estimates \(\{s_h(t)\}_{h \in H}\) on the hyperedges pass information to the cycles through \eqref{eq:weightupdate} and \eqref{eq:normalizationfactor}. 
In these equations, the reweighting function \(f(s_h(t);\beta_t)\) gives larger values when the hyperedge corruption of \(h\) is estimated to be low and smaller values when the corruption is estimated to be high. 
Then if, for example, all the hyperedges in \(N_C\setminus \{h\}\) are estimated to have low corruption, \(w_{h,C}(t)\) will be large and if one or more of the hyperedges in \(N_C\setminus \{h\}\) are estimated to have high corruption, \(w_{h,C}(t)\) will be small. 
For a more precise probabilistic interpretation of CHMP as a message passing algorithm see Section \ref{subsec:fixedpoint}.

The CEMP method from \cite{lerman_robust_2022} can be viewed as a special case of Algorithm \ref{alg:CHMP} (see Section \ref{subsec:comparison} for further discussion). 
Assuming that the complexity of the group operation is fixed based on the choice of \(G\), the complexity of CHMP comes from computing the cycle consistency measure \(d_C\) for each cycle which is dependent on the choice of cycle set \(\mathcal{C}\). 
In general the complexity per iteration is \(O(|\mathcal{C}|)\). 
For \(\mathcal{C} = \mathcal{C}_{n+1}^{n-1}\) and a sufficiently dense hypergraph the complexity per iteration is of order \(O(n^2m^{n+1})\) since there are \(O(m^{n+1})\) cycles in \(\mathcal{C}_{n+1}^{n-1}\) and computing \(d_C\) for a cycle \(C \in \mathcal{C}_{n+1}^{n-1}\) takes \(O(n^2)\) operations. 
Smaller subsets of \(\mathcal{C}_{n+1}^{n-1}\) and/or sparser hypergraphs can be chosen to reduce the computation time. 
Numerical comparisons for the runtime of CHMP are in Section \ref{subsec:synthetic}. 

The main storage costs for CHMP come from storing CHG, \(\{s_h(t)\}_{h \in H}\), \(\{w_{h,C}(t)\}_{h \in H, C \in N_h}\), and \(\{d_C\}_{C \in \mathcal{C}}\). 
The original hypergraph can be discarded once CHG is computed. 
Let \(E := \{\{h,C\} : h \in N_C \}_{h \in H, C \in \mathcal{C}}\).
In general the memory required to store CHG is \(O(|\mathcal{C}|+|H|+ |E|)\). 
Storage for \(\{s_h(t)\}_{h \in H}\), \(\{w_{h,C}(t)\}_{h \in H, C \in N_h}\), and \(\{d_C\}_{C \in \mathcal{C}}\) is \(O(|H|)\), \(O(|E|)\), and \(O(|\mathcal{C}|\), respectively.
Thus the general storage cost for CHMP is \(O(|\mathcal{C}|+|H|+ |E|)\). 
If \(\mathcal{C} = \mathcal{C}_{n+1}^{n-1}\) and the hypergraph is dense, then the memory required is \(O(n\cdot m^{n+1})\), which again can be lessened for sparser hypergraphs or subsets of \(\mathcal{C}_{n+1}^{n-1}\).

\subsection{CHMP for General Hypergraphs and Cycles} \label{subsec:general}

The CHMP framework above assumes the setting of an \(n\)-hypergraph and the cycle set \(\mathcal{C}\) is chosen from a subset of the \((n\!-\!1)\)-cycles on the hypergraph, \(\mathcal{C}^{n-1}\). 
However, as mentioned in Section \ref{subsec:motivation}, there are settings where more flexibility in the hypergraph and cycle structure is desired. 
For example, a data set may contain both pairwise and triple-wise measurements and thus the underlying hypergraph is nonuniform, or we may wish to synchronize 
local measurements of various larger cardinalities in a distributed synchronization problem.
In order to extend CHMP to more general hypergraph settings, the main idea is to determine a way that the cycle inconsistency information \(\{d_C\}_{h \in N_C}\) can be used to construct an estimate for \(\{s_{h}^*\}_{h \in H}\). 
The following theorem extends the ideas of Proposition \ref{prop:corruptionisinconsitency} to a general setting and gives a sufficient condition that a set of cycles must satisfy to guarantee the synchronizability of a noiseless hyperedge potential. 

\begin{theorem} \label{thm:generalsynch}
    Suppose \(\{\overline{\gamma}_h\}_{h \in H}\) is generated by the noiseless corruption model in \eqref{eq:noisemodel} on a hypergraph \(\mathcal{H}(V,H)\). 
    Fix a set of cycles \(\mathcal{C}\). 
    If for every hyperedge \(h = \{v_1, \dots, v_n\} \in H\) and every pair of vertices \((v_i,v_{i+1})\) in the set \(V_h := \{(v_i,v_{i+1})\}_{i = 1,\dots, n-1}\) there is a cycle \(C \in \mathcal{C}\) such that \(v_i\) and \(v_{i+1}\) are adjacent base points in \(C\), \(h\) is a hyperedge containing \(v_i\) and \(v_{i+1}\) in \(C\), and \(N_C \setminus \{h\} \subseteq H_g\), then \(s_h^*\) can be recovered exactly for every \(h \in H\) from a certain set of cycle consistency measures. 
\end{theorem}

\begin{proof}
    Let \(h = \{v_1, \dots, v_n\} \in H\) such that \(|h| = n\). The exact corruption level of \(h\) is given by
    \begin{equation}\label{eq:corruptionlevel}
        s^*_h = d_{n-1}(\overline{\gamma}_h,\overline{\gamma}_h^*),
    \end{equation}
    where \(\overline{\gamma}_h = (g_{v_1}g_{v_2}^{-1},g_{v_2}g_{v_3}^{-1}, \dots, g_{v_{n-1}}g_{v_n}^{-1}) \in G^{n-1}\) is the observed value of the hyperedge potential on \(h\) and \(\overline{\gamma}_h^* = (g_{v_1}^*(g_{v_2}^*)^{-1},g_{v_2}^*(g_{v_3}^*)^{-1}, \dots, g_{v_{n-1}}^*(g_{v_n}^*)^{-1}) \in G^{n-1}\) is the ground truth hyperedge value for \(h\). 
    Since \(d_{n-1}\) is a product metric on an \((n\!-\!1)\)-tuple induced by \(d_{G}\), we can consider it as a function \(d_{n-1}=f: \mathbb{R}^{n-1} \to \mathbb{R}\). 
    Then \eqref{eq:corruptionlevel} can be written as
    \begin{equation}\label{eq:expandedcorruption}
        s^*_h = d_{n-1}(\gamma_h,\gamma_h^*) = f\left(d_G(g_{v_1}g_{v_2}^{-1},g_{v_1}^*(g_{v_2}^*)^{-1}), \dots, d_G(g_{v_{n-1}}g_n^{-1},g_{v_{n-1}}^*(g_{v_n}^*)^{-1})\right).
    \end{equation}
    Thus according to \eqref{eq:expandedcorruption}, \(s^*_h\) can be determined if \(d_G(g_{v_j}g_{v_{j+1}},g_{v_j}^*(g_{v_{j+1}}^*)^{-1})\) is known for all \(j = 1, \dots, n-1\).
    
    According to the assumptions, for every pair \((v_i,v_{i+1}) \in V_h\) there exists a \(C\) such that \(v_i,v_{i+1}\) are in \(C\), \(h \in N_C\) so that \(v_i,v_{i+1}\) are contained in \(h\) in the cycle, and \(N_C \setminus \{h\} \subseteq H_g\). 
    Suppose this \(C\) is a \(k\)-cycle given by 
    \[
        C = [v_{j_1},v_{j_2}, \dots, v_{j_l}, v_{j_1}, \dots, v_{j_k}; h_{j_1}, \dots, h_{j_l}]_k.
    \]
    Without loss of generality, assume further that \(v_{j_1} = v_i\), \(v_{j_2} = v_{i+1}\), and \(h_{j_1} = h\). 
    The cycle \(C\) induces a \(1\)-cycle on the same base points given by 
    \[
        C' = [v_{j_1},v_{j_2}, \dots, v_{j_l}, v_{j_1}; h_{j_1}, \dots, h_{j_l}]_1.
    \]
    In this \(1\)-cycle, \(C'\), \(v_{j_1} , v_{j_2} \in h_{j_1}\) in the cycle and \(N_{C'} \setminus \{h_{j_1}\} \subseteq H_g\) still.
    
    Now, for a hyperedge \(h_{j_k}\) in \(C'\), let 
    \[
        \widehat{\gamma}_{h_{j_k}} := \tau \circ \operatorname{Res}_{h_{j_k} \to \{v_{j_k},v_{j_{k+1}}\}}(\gamma_{h_{j_k}}),
    \]
    where \(\gamma_{h_{j_k}} = \tau^{-1}(\overline{\gamma}_{h_{j_k}})\) is the value of the hyperedge potential on \(h\) in \(G^{|h|}/\Delta\). 
    If \(h_{j_k} \in H_g\), then \(\widehat{\gamma}_{h_{j_k}} = g_{v_{j_k}}^* (g_{v_{j_{k+1}}}^*)^{-1}\) and for \(h\), \(\widehat{\gamma}_{h} = g_{v_i}g_{v_{i+1}}\).
    Then using the bi-invariance of \(d_1\),
    \begin{align*}
        d_{C'} = d_G(\Phi(C'),1) = d_G(\widehat{\gamma}_{h_{j_1}}\widehat{\gamma}_{h_{j_2}}\cdots\widehat{\gamma}_{h_{j_l}},1) &= d_G(g_{v_i}g_{v_{i+1}} g_{v_{i+1}}^* (g_{v_{j_{3}}}^*)^{-1}\cdots g_{v_{j_l}}^* (g_{v_{i}}^*)^{-1},1) 
        \\
        &= d_G(g_{v_i}g_{v_{i+1}} ,g_{v_{i}}^*(g_{v_{i+1}}^*)^{-1}).
    \end{align*}
    Thus we have determined every element of \ref{eq:expandedcorruption} needed to compute \(s_h^*\).
\end{proof}

Theorem \ref{thm:generalsynch} makes no assumption on the hypergraph, allowing for hyperedges of varying size, or the order of the cycles. This suggests that Algorithm \ref{alg:CHMP} could be extended to a general hypergraph setting since the set \(\{d_C\}_{h \in N_C}\), for \(C\) belonging to a set of cycles satisfying the assumptions of the proposition, contains enough information to determine \(\{s_{h}^*\}_{h \in H}\). But we leave the details of this extension for future work.

Proposition \ref{prop:corruptionisinconsitency} is a special case of Theorem \ref{thm:generalsynch}. 
Indeed for any hyperedge \(h\) on a \(n\)-hypergraph, a good \((n\!-\!1)\)-cycle \(C\) with respect to \(h\) satisfies the conditions of Theorem \ref{thm:generalsynch} for all pairs in \(V_h\). 
However, even though any order cycles may be considered, our motivation for considering \(n-1\)-cycles of \(n\)-hypergraphs comes from the fact that in that setting, the Good Cycle Condition implies that \(s_h^* = d_C\) directly for any good cycle \(C\). This one cycle \(C\) covers the assumptions of the proposition for all pairs in \(V_h\) simultaneously which significantly simplifies the computations.

\section{Recovery of the Vertex Potential} \label{sec:recovery}

Proposition \ref{prop:exact-recovery} shows that exact recovery of the hyperedge corruption levels is equivalent to recovering the underlying vertex potential.
In this section, the methods for recovering a vertex potential from pairwise edge corruption estimates in \cite{lerman_robust_2022} are reviewed. 
Then the extension of these methods to the higher-order synchronization setup is considered.

In classical pairwise group synchronization, one method for estimating the group elements \(\{g_i\}_{i \in V}\) uses the minimal spanning tree (MST) of the graph weighted by the estimated corruption levels from CEMP \cite{lerman_robust_2022}. 
This procedure is stated in Algorithm \ref{alg:MST}. 
In this procedure, edges in the graph \(\mathcal{G}(V,E)\) are given weights \(\{s_{ij}(T)\}_{ij \in E}\) so that a minimally weighted spanning tree (MST) of this weighted graph can be constructed. 
Then an arbitrary vertex is labeled \(1\) and \(g_1\) is set to be the identity element \(1\). The remaining vertices are assigned values in \(G\) through the equation \(g_i = {\overline{\gamma}}_{ij}\cdot g_j\) where \(ij\) is an edge in MST. 
The complexity of Algorithm \ref{alg:MST} depends mainly on the complexity of method for finding MST as the vertex assignment step is \(O(|\widetilde{E}|)\) where \(\widetilde{E}\) is the set of edges in MST.

\begin{algorithm}[htb]
    \caption{Vertex Potential Recovery via Minimal Spanning Tree (MST) \cite{lerman_robust_2022}}\label{alg:MST}
    \begin{algorithmic}
        \REQUIRE Graph \(\mathcal{G}(V,E)\), observed measurements \(\{\overline{\gamma}_{ij}\}_{{ij} \in E}\), 
        edge corruption estimates\\ \(\{s_{ij}(T)\}_{ij \in E}\)  (output of CEMP \cite{lerman_robust_2022}).
        \\
        \STATE \textbf{Steps:}
        \STATE Generate MST\((V,\widetilde{E})\) from \(\mathcal{G}(V,E)\) and \(\{s_{ij}(T)\}_{ij \in E}\)
        \STATE \(g_1 = 1\)
        \FORALL {\(ij \in \widetilde{E}\)}
            \STATE \(g_i = \overline{\gamma}_{ij}\cdot g_j\)
        \ENDFOR
        \ENSURE \(\{g_{i}\}_{i \in V}\)
    \end{algorithmic}
\end{algorithm}

If the data is noiseless (but possibly corrupted) and the set of good edges form a connected graph on all the vertices, then Algorithm \ref{alg:MST} achieves exact recovery \cite{lerman_robust_2022}. 
However Algorithm \ref{alg:MST} is not recommended for data that is sufficiently noisy since the errors accumulate in each multiplication step. 
In the noisy domain, the edge corruption estimates can be used to form weights to refine an energy minimization process that solves \eqref{eq:minimizeredge}. 
For example, when \(G\) is a subgroup of \(O(d)\), \cite{lerman_robust_2022} proposes a weighted spectral method using the estimated weights \(\{s_{ij}(T)\}_{ij \in E}\). 
In particular they estimate the probability that an edge is a good edge by computing
\begin{equation}
    \widetilde{p}_{ij} := \frac{f(s_{ij}, \beta_T)}{\sum_{j \in V : ij \in E} f(s_{ij}, \beta_T)},
\end{equation}
and use these probabilities \(\{\widetilde{p}_{ij}\}_{ij \in E}\) as weights in a spectral method as described in Algorithm \ref{alg:GCW}, which is modeled after graph connection weight matrices from vector diffusion maps \cite{singer_vector_2012}. 
Algorithm \ref{alg:GCW} can also be thought of as a convex relaxation of a weighted version of \eqref{eq:minimizeredge}:
\begin{equation}\label{eq:weightedminimizeredge}
    \min_{\{g_i\}_{i \in V}} \sum_{ij \in H} \widetilde{p}_{ij}\cdot d_G(\overline{\gamma}_{ij},g_ig_j^{-1}),
\end{equation}
In \cite{lerman_robust_2022}, this method is said to be close to a direct least squares solver for \eqref{eq:weightedminimizeredge} on the set of good edges when the corruption estimates \(\{s_{ij}(T)\}_{ij \in E}\) are sufficiently close to \(\{s_{ij}^*(T)\}_{ij \in E}\). 
For a more detailed description of Algorithm \ref{alg:GCW} see \cite{arrigoni_spectral_2016,lerman_robust_2022}.

The complexity of Algorithm \ref{alg:GCW} depends on the cost of computing eigenvectors of GCW. 

\begin{algorithm}[htb]
    \caption{Vertex Potential Recovery via Graph Connection Weight Matrix (GCW) \cite{lerman_robust_2022}}\label{alg:GCW}
    \begin{algorithmic}
        \REQUIRE Observed measurements \(\{\overline{\gamma}_{ij}\}_{{ij} \in E}\), edge probability estimates \(\{\widetilde{p}_{ij}\}_{ij \in E}\).
        \\
        \STATE \textbf{Steps:}
        \FORALL{\(ij \in E\)}
            \STATE \(\mathrm{GCW}_{ij} = \widetilde{p}_{ij}\overline{\gamma}_{ij}\)
        \ENDFOR
        \FORALL{\(ij \not\in E\)}
            \STATE \(\mathrm{GCW}_{ij} = 0\)
        \ENDFOR
        \STATE Compute the block vector \(\widehat{X}\) from the top \(d\) eigenvectors of \(\mathrm{GCW}\)
        \FORALL{\(i \in V\)}
            \STATE Compute \(\mathbf{x}_{i}\) by projecting the \(i\)-th block of \(\widehat{X}\) onto \(G\)
            \STATE \(g_i = \mathbf{x}_i\)
        \ENDFOR
        \ENSURE \(\{g_{i}\}_{i \in V}\)
    \end{algorithmic}
\end{algorithm}

For higher-order group synchronization, consider noiseless data (with possible corruption) which can be used to achieve exact recovery of the vertex potential if the set of good hyperedges forms a connected hypergraph on the full set of vertices, as stated in Proposition \ref{prop:exact-recovery}. 
To construct a naive algorithm, analogous to Algorithm \ref{alg:MST}, which uses the hyperedge corruption estimates to find an ideal subset of hyperedges which can define the vertex potential, it is necessary to consider what such an ideal subset would be. 
For one, the set of hyperedges would need to cover all of the vertices and the hyperedges would need to overlap on at most one vertex. 
Moreover, the hypergraph should contain no \(1\)-cycles, or be \(1\)-acyclic. 
Restricting to just one vertex overlaps and requiring acyclicity ensures that the vertex potential is well defined since there would be only one \(1\)-path between any two vertices. 

Such a subhypergraph as described above is called a \textit{spanning tree}, however, even for ``nice" hypergraphs, a spanning tree is not always guaranteed to exist \cite{budden_minimally_2022,greenhill_spanning_2022}. 
For this reason, the direct hypergraph spanning tree construction is avoided and instead the hyperedge corruption estimates \(\{\overline{\gamma}_h\}_{h \in H}\) are used to form a refined set of pairwise data (graph, edge potential, and edge corruption estimates). 
The construction of this refinement is detailed in Algorithm \ref{alg:refine}. 
For a hypergraph \(\mathcal{H}(V,H)\), hyperedge potential \(\{\overline{\gamma}_h\}_{h \in H}\), and hyperedge corruption estimates \(\{s_h(T)\}_{h \in H}\) the output graph, edge potential, and edge corruption estimates of Algorithm \ref{alg:refine} are referred to as \(\mathcal{G}_{\min}(V,E)\), \(\{\overline{\gamma}_{ij}^{\min}\}_{ij \in E}\), and \(\{s_{ij}^{\min}(T)\}_{ij \in E}\), respectively. 

For each pair of vertices \(\{i,j\} \subseteq V\), Algorithm \ref{alg:refine} first determines the hyperedges which contain both \(i\) and \(j\). 
Then the hyperedge or hyperedges with the smallest estimated corruption level (from the output of CHMP) among hyperedges containing \(i\) and \(j\) are selected. 
From this set of hyperedges with minimal corruption estimation, one hyperedge, \(\widehat{h}\), is chosen arbitrarily. 
The corruption estimate for the edge \(ij\) is assigned to be the corruption estimate for the chosen hyperedge \(\widehat{h}\) and the group measurement for edge \(ij\) is induced from the group measurement of \(\widehat{h}\) restricted to \(i\) and \(j\). 
If there is no hyperedge containing \(i\) and \(j\), then edge \(ij\) is not included in the resulting graph.

The complexity of Algorithm \ref{alg:refine} depends on the number of pairs of vertices which is \(O(m^2)\) and the number of hyperedges containing each pair of vertices. 
For a sufficiently dense hypergraph the complexity is \(O(m^3)\).

\begin{algorithm}[htb]
    \caption{Hypergraph to Graph Refinement}\label{alg:refine}
    \begin{algorithmic}
        \REQUIRE Hypergraph \(\mathcal{H}(V,H)\), hyperedge potential \(\{\overline{\gamma}_h\}_{h \in H}\), hyperedge corruption estimates \(\{s_h(T)\}_{h \in H}\) (output of CHMP).
        \\
        \STATE \textbf{Steps:}
        \FORALL{\(\{i,j\} \subseteq V\)}
            \STATE \(h_{ij} = \arg\min_{h \in H : \{i,j\} \subseteq h} s_{h}(T) \subseteq H\)
            \IF{\(h_{ij} \not= \emptyset\)}
                \STATE Choose \(\widehat{h} \in h_{ij}\) (deterministically or probabilistically).
                \STATE \(E = E \cup \{ij\}\)
                \STATE \(s_{ij}^{\min}(T) = s_{\widehat{h}}(T) \in \mathbb{R}\) 
                \STATE \(\overline{\gamma}_{ij}^{\min} = \tau \circ \operatorname{Res}_{\widehat{h} \to \{i,j\}}(\tau^{-1}(\overline{\gamma}_{\widehat{h}})) \in G\)
            \ENDIF
        \ENDFOR
        \ENSURE  \(\mathcal{G}_{\min}(V,E)\), \(\{\overline{\gamma}_{ij}^{\min}\}_{ij \in E}\), and \(\{s_{ij}^{\min}(T)\}_{ij \in E}\).
    \end{algorithmic}
\end{algorithm}

Algorithm \ref{alg:refine} builds a graph and hyperedge potential from the higher-order data by selecting the hyperedge with the lowest estimated corruption that contains a pair of vertices \(i\) and \(j\) and using the data from that hyperedge to assign data to the edge \(ij\). 
Even in the case that a minimal spanning tree of a hypergraph exists, the output of Algorithm \ref{alg:refine} might give a different vertex potential than the one induced by the minimal spanning tree. 
This is to our advantage, since Algorithm \ref{alg:refine} induces data per pair of vertices rather than having one hyperedge induce estimates for each of its vertices. 
To demonstrate why this is an advantage, consider the following example.
\begin{example} \label{ex:refinement}
    Define the hypergraph: 
    \[
        \mathcal{H}(\{1,2,3,4,5\},\{h_{123},h_{234},h_{345},h_{451},h_{512}\})
    \]
    such that \(s^*_{h_{123}} = 0.1\), \(s^*_{h_{234}} = 0.05\), \(s^*_{h_{345}} = 0.1\), \(s^*_{h_{451}} = 0.2\), and \(s^*_{h_{541}} = 0.2\). 
    Clearly the minimal spanning tree for \(\mathcal{H}\) is \(\{h_{123},h_{345}\}\). 
    However, this minimal spanning tree discards \(h_{234}\) which has the lowest corruption of all the hyperedges. 
    In the Algorithm \ref{alg:refine} construction, \(h_{234}\) is chosen to induce data for \(23\) and \(34\), while data for \(12\) and \(45\) are induced by \(h_{123}\) and \(h_{345}\), respectively. 
    While this construction doesn't necessarily guarantee the best possible vertex potential to fit the observations, it is certainly a more natural way to denoise the hyperedge potential from the information estimated. 
    What this example demonstrates is one of the core principles of higher-order synchronization that Algorithm \ref{alg:refine} exploits: redundancy of higher-order information leads to better denoising of the data.
\end{example}

The full pipeline proposed for recovering the vertex potential from higher-order data is to first compute the hyperedge corruption estimates using Algorithm \ref{alg:CHMP}, refining the data using Algorithm \ref{alg:refine}, and finally apply Algorithm \ref{alg:MST} or Algorithm \ref{alg:GCW} to the output of Algorithm \ref{alg:refine} as dictated by the data. 
When using Algorithm \ref{alg:MST} or Algorithm \ref{alg:GCW} the pipeline is referred to as \textbf{CHMP + MST} or \textbf{CHMP + GCW}, respectively. 
In Section \ref{sec:numerical-results} the two proposed pipelines are applied to the groups \(SO(2)\) and \(SO(3)\). 
It is also possible to apply other pairwise synchronization methods to the output of Algorithm \ref{alg:refine} by, for example, using \(\{s_{ij}^{\min}(T)\}_{ij \in E}\) as weights directly in the method or indirectly to clean the data by removing edges with high corruption values. 

\section{Analysis of CHMP} \label{sec:chmp-analysis}

This section provides a theoretical basis for the performance of the main algorithm, CHMP. 
First, in Section \ref{subsec:fixedpoint} it is shown that the exact hyperedge corruption levels form a fixed point of CHMP. 
In Section \ref{subsec:exactrecovery} exact recovery and convergence guarantees are established using the adversarial corruption model defined in \eqref{eq:noisemodel}. 
Section \ref{subsec:samplecomplexity} discusses the sample complexity of CHMP. 
Much of the analysis and techniques come from extending the work in \cite{lerman_robust_2022} to the higher-order domain. 
Section \ref{subsec:comparison} concludes with a discussion of the theoretical guarantees of CHMP compared to those of CEMP, which can be considered a special case of CHMP.

\subsection{Fixed Point of CHMP} \label{subsec:fixedpoint}

The ground truth hyperedge corruption levels form a fixed point of the CHMP algorithm. 
To see this, define a set of \textit{ideal weights} for a hyperedge \(h \in H\) and a cycle \(C \in N_h\):
\begin{equation}\label{eq:ideal-weights}
    w^*_{{h},C} := \frac{1}{|G_{{h}}|} \mathbf{1}_{\{C \in G_{h}\}}.
\end{equation}

\begin{proposition}
    Suppose \(\{\overline{\gamma}_h\}_{h \in H}\) is generated by the noiseless corruption model in \eqref{eq:noisemodel} on a \(n\)-hypergraph \(\mathcal{H}(V,H)\).
    The ground truth hyperedge corruption estimates and the ideal weights,
    \begin{equation}\label{eq:fixedpoint}
        \left(\{s^*_{h}\}_{{h} \in H}, \{w^*_{{h},C}\}_{{h} \in H, C \in N_{h}}\right),
    \end{equation}
    are a fixed point of Algorithm \ref{alg:CHMP}. 
\end{proposition}

\begin{proof}
    Proposition \ref{prop:corruptionisinconsitency} can be used to show that the ideal weights defined in \eqref{eq:ideal-weights} give an exact estimate of the hyperedge corruption levels:
    \[
        \sum_{C \in N_{{h}}} w^*_{{h},C} d_C = \frac{1}{|G_{h}|} \sum_{C \in N_{h}} \mathbf{1}_{\{C \in G_{h}\}} d_C = \frac{1}{|G_{h}|} \sum_{C \in G_{h}} d_C = \frac{1}{|G_{h}|} \sum_{C \in G_{h}} s^*_{h} = s^*_{h}.
    \]
    To see that the exact corruption measurements give the ideal weights, the reweighting function of CHMP can be interpreted as a probability using a statistical model for the exact and estimated corruption levels. 
    Suppose \(\{s^*_{h}\}_{{h} \in H}\) and \(\{s_{h}(t)\}_{{h} \in H}\) are i.i.d. random variables for all \(t\) and that for any \({h} \in H\), \(s^*_{h}\) is independent of \(s_{h'}\) for \({h} \not= {h'} \in H\). 
    Then reweighting function \(f\) gives the probability
    \begin{equation}\label{eq:reweightasprob}
        f(x;\beta_t) = \mathbb{P}(s^*_{h} = 0\ |\ s_{h}(t) = x).
    \end{equation}
    In light of \eqref{eq:reweightasprob}, equation \eqref{eq:weightupdate} becomes 
    \begin{align*}
        w_{{h},C}(t) &= \frac{1}{Z_{h}(t)} \prod_{{h'} \in N_C \setminus \{{h}\}} \mathbb{P}\left(s^*_{h'} = 0\ |\ s_{h'}(t)\right) 
        \\
        &= \frac{1}{Z_{h}(t)} \mathbb{P}\left(\{s^*_{h'}\}_{{h'} \in N_C \setminus \{{h}\}} = \bar{0}\ |\ \{s_{h'}(t)\}_{{h'} \in N_C \setminus \{{h}\}}\right) 
        \\
        &= \frac{1}{Z_{h}(t)} \mathbb{P}\left(C \in G_{h}\ |\ \{s_{h'}(t)\}_{{h'} \in N_C \setminus \{{h}\}}\right),
    \end{align*}
    where the normalization factor in \eqref{eq:normalizationfactor} is
    \begin{align*}
        Z_{{h}}(t) &= \sum_{C \in N_{h}} \prod_{{h'} \in N_C \setminus \{{h}\}} f(s_{h'}(t), \beta_t) 
        \\ 
        &= \sum_{C \in N_{h}} \prod_{{h'} \in N_C \setminus \{{h}\}} \mathbb{P}\left(s^*_{h'} = 0\ |\ s_{h'}(t)\right) 
        \\
        &= \sum_{C \in N_{h}} \mathbb{P}\left(C \in G_{h}\ |\ (s_{h'}(t))_{{h'} \in N_C \setminus \{{h}\}}\right).
    \end{align*}
    Then if \(\{s_{h}(t)\}_{h \in H} = \{s^*_{h}\}_{{h} \in H}\), the weights are given by: 
    \begin{equation}\label{eq:weightatexact}
        w^*_{{h},C}(t) = \frac{1}{Z^*_{h}(t)} \mathbb{P}(C \in G_{h}\ |\ (s^*_{h'}(t))_{{h'} \in N_C \setminus \{{h}\}})
    \end{equation}
    and
    \begin{equation}\label{eq:normalizationatexact}
        Z^*_{{h}}(t) = \sum_{C \in N_{h}} \mathbb{P}(C \in G_{h}\ |\ (s^*_{h'}(t))_{{h'} \in N_C \setminus \{{h}\}}).
    \end{equation}
    Since Proposition \ref{prop:corruptionisinconsitency} says that \(C \in G_{h}\) is equivalent to \(s^*_{h'} = 0\) for all \({h'} \in N_C \setminus \{{h}\}\), \eqref{eq:weightatexact} and \eqref{eq:normalizationatexact} become
    \[
        w^*_{{h},C} = \frac{1}{Z^*_{h}(t)} \mathbf{1}_{\{C \in G_{h}\}} \text{ and } Z^*_{{h}}(t) = \sum_{C \in N_{h}} \mathbf{1}_{\{C \in G_{h}\}}  = |G_{{h}}|,
    \]
    which together are equivalent to \eqref{eq:ideal-weights}.
\end{proof}

\subsection{Exact Recovery and Global Linear Convergence Under Adversarial Corruption} \label{subsec:exactrecovery}

In this section, the global linear convergence of CHMP assuming adversarial convergence is established both in a noisy and noiseless setting. 
In particular, deterministic exact recovery of the hyperedge corruption levels is given according to a bound on the ratio of bad cycles per hyperedge. 
Clearly, the Good Cycle Condition in Proposition \ref{prop:corruptionisinconsitency} is a necessary assumption for exact recovery. 
A connected hypergraph and \(G_h \not=\emptyset\) for all \(h \in H\) are requirements for the Good Cycle Condition to hold. 
This observation implies the need for a non-zero lower bound on the value of 
\(|G_h|/|N_h|\) for all \(h \in H\) which in turn implies an upper bound on
\begin{equation}
    \lambda_{h} := \frac{|B_{h}|}{|N_{h}|} \text{ and } \lambda := \max_{h \in H} \lambda_{h}.
\end{equation}
Following the argument in \cite{lerman_robust_2022}, the only assumption made on the properties of the graph and hyperedge potential is an upper bound on \(\lambda\). 

\begin{theorem} \label{thm:noiselessrecovery}
    Assume a connected \(n\)-hypergraph and a hypergraph potential generated by the noiseless adversarial corruption model \eqref{eq:noisemodel} with \(\lambda < 1/(2n+1)\). 
    If CHMP has parameters \(\mathcal{C} = \mathcal{C}_{n+1}^{n-1}\) and \(\{\beta_t\}_{t\geq 0}\) such that \(\beta_0 \leq 1/2n\lambda\) and  \(\beta_{t+1} = r\beta_t\) for all \(t \geq 1\) and some \(1 < r < (1-\lambda)/2n\lambda\), then the estimates \(\{s_{h}(t)\}_{{h} \in H}\) for \(t\geq 0\) computed by CHMP satisfy
    \begin{equation}
        \max_{{h} \in H} |s_{h}(t) - s_{h}^*| \leq \frac{1}{2n\beta_0 r^t} \text{ for all } t \geq 0.
    \end{equation}
\end{theorem}

\begin{proof}
    Suppose \(\mathcal{C} = \mathcal{C}_{n+1}^{n-1}\) and let \(h \in H\). 
    If \(C \in \mathcal{C}_{n+1}^{n-1}\) such that \(h \in N_C\), then Proposition \ref{prop:corruptionbound} implies that
    \begin{equation} \label{eq:restatedlemma}
        |d_C - s^*_{h}| \leq \sum_{h' \in N_C \setminus h} s^*_{h'}.
    \end{equation}
    Further, \eqref{eq:weightupdate} and \eqref{eq:normalizationfactor} can be combined into a single update step:
    \begin{equation}\label{eq:restatedcombinedsteps}
            s_{h}(t+1) = \frac{\sum_{C\in N_{{h}}} \exp\left({-\beta_t\left(\sum_{h' \in N_C \setminus h} s_{h'}(t)\right)}\right) \cdot d_C}{\sum_{C\in N_{{h}}} \exp\left({-\beta_t\left(\sum_{h' \in N_C \setminus h} s_{h'}(t)\right)}\right)}.
    \end{equation}
    Define
    \begin{equation}
       \epsilon_{h}(t) := |s_{h}(t) - s^*_{h}| \text{ and } \epsilon(t) := \max_{{h} \in H}|s_{h}(t) - s_{h}^*|.
    \end{equation}
    Then combining \eqref{eq:restatedlemma} and \eqref{eq:restatedcombinedsteps} gives
    \begin{align}
        \epsilon_{h}(t+1) &\leq \frac{\sum_{C\in N_{{h}}} \exp\left({-\beta_t\left(\sum_{h' \in N_C \setminus h} s_{h'}(t)\right)}\right) \cdot |d_C - s_{h}^*|}{\sum_{C\in N_{{h}}} \exp\left({-\beta_t\left(\sum_{h' \in N_C \setminus h} s_{h'}(t)\right)}\right)} \nonumber
        \\
        & \leq \frac{\sum_{C\in N_{{h}}} \exp\left({-\beta_t\left(\sum_{h' \in N_C \setminus h} s_{h'}(t)\right)}\right) \cdot \left(\sum_{h' \in N_C \setminus h} s^*_{h'}\right)}{\sum_{C\in N_{{h}}} \exp\left({-\beta_t\left(\sum_{h' \in N_C \setminus h} s_{h'}(t)\right)}\right)}. \label{eq:inequalityforerror}
    \end{align}
    Further, since 
    \[
        \sum_{h' \in N_C \setminus h} s^*_{h'} = 0
    \]
    for \(C \in G_{h}\), \eqref{eq:inequalityforerror} becomes
    \begin{align}
        \epsilon_{h}(t+1) & \leq \frac{\sum_{C\in B_{{h}}} \exp\left({-\beta_t\left(\sum_{h' \in N_C \setminus h} s_{h'}(t)\right)}\right) \cdot \left(\sum_{h' \in N_C \setminus h} s^*_{h'}\right)}{\sum_{C\in G_{{h}}} \exp\left({-\beta_t\left(\sum_{h' \in N_C \setminus h} s_{h'}(t)\right)}\right)} \nonumber
        \\
        & \leq \frac{\sum_{C\in B_{{h}}} \exp\left({\beta_t\left(\sum_{h' \in N_C \setminus h} \epsilon_{h'}(t)\right)}\right) \cdot \exp\left({-\beta_t\left(\sum_{h' \in N_C \setminus h} s_{h'}^* \right)}\right)  \cdot\left(\sum_{h' \in N_C \setminus h} s^*_{h'}\right)}{\sum_{C\in G_{{h}}} \exp\left({-\beta_t\left(\sum_{h' \in N_C \setminus h} \epsilon_{h'}(t)\right)}\right)}. \label{eq:errorbound}
    \end{align}
    Assume the induction hypothesis \(\epsilon(t) \leq 1/2n\beta_t\) for some \(t\).  Let \(t=0\), then 
    \[
        \epsilon_{h}(0) = |s_{h}(0)-s_{h}^*| \leq \frac{\sum_{C\in N_{{h}}} |d_C  - s_{h}^*|}{|N_{{h}}|} = \frac{\sum_{C\in B_{{h}}} |d_C  - s_{h}^*|}{|N_{{h}}|}
        \leq \frac{|B_{{h}}|}{|N_{{h}}|} \leq \lambda \leq \frac{1}{2n\beta_0}.
    \]
    Now to show that \(\epsilon(t+1) \leq 1/2n\beta_{t+1}\). Remark that for \(x \geq 0\) and \(a > 0\), the following inequality holds: \(xe^{-ax} \leq 1/ea\). 
    Let \(x = \sum_{h' \in N_C \setminus h} s^*_{h'}\) and \(a = \beta_t\). 
    Then applying the induction hypothesis to \eqref{eq:errorbound} gives
    \begin{align} \label{eq:finalinequality}
        \epsilon_{h}(t+1) & \leq \frac{ (e\beta_t)^{-1} \cdot\left(\sum_{C\in B_{{h}}} \exp\left({\beta_t(\frac{n}{2n\beta_t})}\right)\right) }{\sum_{C\in G_{{h}}} \exp\left({-\beta_t\left(\sum_{h' \in N_C \setminus h} \epsilon_{h'}(t)\right)}\right)} \nonumber
        \\
        &\leq \frac{ |B_{{h}}|\cdot (e^{\frac{1}{2}}\beta_t)^{-1} }{ \sum_{C\in G_{{h}}} \exp\left({-\beta_t\left(\sum_{h' \in N_C \setminus h} \epsilon_{h'}(t)\right)}\right)} 
        \leq  \frac{ |B_{{h}}|\cdot (e^{\frac{1}{2}}\beta_t)^{-1} }{\sum_{C\in G_{{h}}} \exp\left({-\beta_t(\frac{n}{2n\beta_t})}\right)} 
        = \beta_t^{-1} \cdot \frac{|B_{{h}}|}{|G_{{h}}|}.
    \end{align}
    Maximizing over \({h}\) on both sides of \eqref{eq:finalinequality} and applying the assumptions on \(\lambda\) shows that indeed
    \[
        \epsilon(t+1) \leq \frac{\lambda}{1-\lambda}\cdot \frac{1}{\beta_t} < \frac{1}{2n\beta_{t+1}}. \qedhere
    \] 
\end{proof}

Now assume the same setting as in Theorem \ref{thm:noiselessrecovery} but allow the uncorrupted hyperedge measurements in the adversarial corruption model to be perturbed by noise. 
The following theorem shows that approximate recovery of the corruption levels is guaranteed with error on the order of the noise level.

\begin{theorem} \label{thm:noisyrecovery}
    Assume a connected \(n\)-hypergraph and a hyperedge potential generated by the noisy adversarial corruption model \eqref{eq:noisemodel} with \(\lambda < 1/(2n+1)\). 
    Let \(\delta >0\) such that \(s_{h}^* = d_{n-1}(\overline{\gamma}_{h}^{\epsilon},1) \leq \delta\) for all \(h \in H_g\). If CHMP has parameters \(\mathcal{C} = \mathcal{C}_{n+1}^{n-1}\) and \(\{\beta_t\}_{t\geq 0}\) such that 
    \begin{equation}
        \frac{1}{2n\beta_0} > \max\left\{\frac{(2n+1)(1-\lambda)\delta}{2(1-(2n+1)\lambda)},\lambda + \frac{(2n+1)\delta}{2}\right\}
    \end{equation}
    and
    \begin{equation}
        (2n^2+n)\delta + \frac{2n\lambda}{(1-\lambda)\beta_t} \leq \frac{1}{\beta_{t+1}} < \frac{1}{\beta_t}.
    \end{equation}
    Then 
    \begin{equation}\label{eq:max_error}
        \max_{h \in H} |s_{h}(t) - s_{h}^*| \leq \frac{1}{2n\beta_t} - \frac{1}{2}\delta \text{ for all } t \geq 0.
    \end{equation}
    Additionally, the following asymptotic bound holds:
    \begin{equation}
        \limsup_{t \to \infty} \max_{h \in H} |s_{h}(t) - s_{h}^*| \leq \left(\frac{2n+1}{2\varepsilon} \cdot \frac{(1-\lambda)}{(1-(2n+1)\lambda)} - \frac{1}{2}\right)\delta
    \end{equation}
    where \(0 < \varepsilon \leq 1\) is defined as
    \begin{equation} \label{eq:varepsilon}
        \varepsilon:= \lim_{t \to \infty} \beta_t\frac{(2n^2+n)(1-\lambda)\delta}{(1-(2n+1)\lambda)}. 
    \end{equation}
\end{theorem}

The proof of Theorem \ref{thm:noisyrecovery} follows the proof of Theorem \ref{thm:noiselessrecovery}, so it is given in Appendix \ref{appendix:CHMP-analysis-proofs}.

\subsection{Sample Complexity} \label{subsec:samplecomplexity}

In this section, we state a simple result regarding the number of samples required to recover the corruption estimates from a \(n\)-hypergraph under a Uniform Corruption Model for Hypergraphs. 
While the guarantees of Section \ref{subsec:exactrecovery}  are quite strong, the assumptions on \(\lambda\) can be limiting, especially as \(n\) grows. 
The Uniform Corruption Model for Hypergraphs defined below is more generally applicable. 

One natural way to generate a random \(n\)-hypergraph is to extend the Erd\H{o}s--R\'enyi random graph model \cite{bick_what_2023}. 
Let \(\mathcal{H}(n,m,p)\) be an Erd\H{o}s--R\'enyi \(n\)-uniform random hypergraph with \(m\) nodes where \(p\) is the probability that a collection of \(n\) distinct vertices form an \(n\)-hyperedge. 
Then a Uniform Corruption Model on a Hypergraph (UCMH) can be defined using the same parameters that define a Uniform Corruption Model (UCM) on a graph. 
We extend the UCM models from \cite{lerman_robust_2022} and \cite{wang_exact_2013} to hypergraphs by defining \(\text{UCMH}(n,m,p,q)\) to be a Uniform Corruption Model on \(\mathcal{H}(n,m,p)\) where \(q\) is the corruption probability. 
For each hyperedge \(h\) in the UCMH, the image under the hyperedge potential is assigned using the model
\begin{equation}
    \overline{\gamma}_{h} = 
    \begin{cases} \overline{\gamma}_{h}^* & \text{ with probability } 1-q
    \\
    \widetilde{\gamma}_{h} & \text{ with probability } q,
    \end{cases}
\end{equation}
where each entry in \(\widetilde{\gamma}_h\) is independently drawn from the Haar measure on \(G\), denoted \(\text{Haar}(G)\). 
This model aligns with the noiseless model given in \eqref{eq:noisemodel} where the good and bad edges are partitioned probabilistically. 
For \(n = 2\), \(\text{UCMH}(n,m,p,q)\) is equivalent to \(\text{UCM}(m,p,q)\) from \cite{lerman_robust_2022}. 

The process of corrupting a hyperedge occurs in two steps. 
First, sample a set \(\widetilde{H}_b\) with probability \(q\). 
Then sample \(H_b\) from \(\widetilde{H}_b\) with probability \(1-(\widehat{p})^{n-1}\) where \(\widehat{p} := \mathbb{P}(g = 1)\) for arbitrary \(g \sim \text{Haar}(G)\). 
This second step accounts for the probability that \(\widetilde{\gamma}_{h} = \overline{\gamma}_{h}\). 
Thus the total probability that an edge is not corrupted is \(q_* = 1-q + q(\widehat{p})^{n-1}\).

For Lie groups such as \(SO(d)\), \(\widetilde{\gamma}_{h} \not= \overline{\gamma}_{h}^*\) with probability \(1\) since \(\widehat{p} = 0\) due to the invariance property of the Haar measure. 
Thus, \(\mathbb{P}(h \in H_b : h \in H) = q\) and \(\mathbb{P}(h \in H_g : h \in H) = 1-q = q_g\). 
For \(\mathbb{Z}_2\), \(\widehat{p} = 1/2\) so in general \(\widetilde{H}_b \not= H_b\). 
In particular, \(\mathbb{P}(h \in H_b : h \in H) = q(1-2^{1-n})\) and \(q_* = 1 - q(1-2^{1-n})\).

Now the UCMH model can be applied to analyze the sample complexity of CHMP. 
Define
\begin{equation}
    D_h := \{d_C : C \in N_C\}.
\end{equation}
Proposition \ref{prop:sample-complexity} below uses the simple estimates 
\begin{equation}
    \widehat{s}_{h} := \mathrm{mode}(D_{h}),
\end{equation}
and assumes that only cycles in \(\mathcal{C}_{n+1}^{n-1}\) are considered. 
Theoretical guarantees for this estimator assume an idealized scenario where there is no noise. 
As pointed out in \cite{lerman_robust_2022}, there are limitations for the mode statistic as it is highly sensitive to outliers and is highly unlikely to exist when the uncorrupted measurements are noisy. 
Nonetheless, this model can provide a baseline comparison with the sample complexity of \(\text{UCM}(m,p,q)\) in \cite{lerman_robust_2022}.

\begin{proposition} \label{prop:sample-complexity}
    Fix an order \(n\). 
    Let \(\mathcal{H}(n,m,p)\) be a hypergraph generated under the noiseless uniform corruption model \(\text{UCMH}(n,m,p,q)\) where the underlying group \(G\) satisfies \(\mathbb{P}(g = 1) = 0\) for any \(g \in G\). 
    If \(m / \log(m) \geq c/(p^n q_g^n)\) for some \(c \geq 75n / 16\), then for any \(0 < p \leq 1\) and \(0 \leq q < 1\), the estimates \(\widehat{s}_{h}\) give exact estimates of \(s^*_{h}\) with probability \(1 - m^{n^*}\), where \(n^* = n - 16c/75\).
\end{proposition}

The proof of Proposition \ref{prop:sample-complexity} is in Appendix \ref{appendix:CHMP-analysis-proofs}.

\subsection{Remarks on CEMP as a Special Case of CHMP} \label{subsec:comparison}

The results in \cite{lerman_robust_2022} for the exact recovery and sample complexity of CEMP can be viewed as a special case of the results above. 
It is natural to wonder how increasing the size of the hyperedges and the complexity of cycles affects the resulting estimates given by CHMP. 
For example, Theorem \ref{thm:noiselessrecovery} implies that the order of convergence is \(O(1/n)\), which is improved by considering larger values of \(n\). 

\begin{table}[htb] 
    \centering
    \begin{tabular}{|c|c|}
        \hline
        \(n\) & sample complexity     \\ \hline
        \(2\) & \(O(2p^{-2}q_g^{-2})\) \\ \hline
        \(3\) & \(O(3p^{-3}q_g^{-3})\) \\ \hline
        \(4\) & \(O(4p^{-4}q_g^{-4})\) \\ \hline
    \end{tabular}
    \caption{\centering Summary of sample complexity for various values of \(n\).}
    \label{tbl:samplecomplexity}
\end{table}

However, increasing the value of \(n\) harms the dependence on \(\lambda\), making it harder for higher-order setups to deal with large amounts of corruption. 
Along this trend, Proposition \ref{prop:sample-complexity} implies a sample complexity of order \(O(np^{-n}q_g^{-n})\). 
Table \ref{tbl:samplecomplexity} summarizes the sample complexity for \(n  = 2,3,\) and \(4\).
As \(n\) increases, so does the number of hyperedges required for a cycle. 
Thus to guarantee sufficient coverage of good cycles for each hyperedge the number of samples needs to increase. 
These comparisons are also confirmed by numerical experiments in the next section. 

\section{Numerical Results} \label{sec:numerical-results}

In this section, several numerical experiments to demonstrate the practical performance of CHMP are presented. 
While the main contribution of this paper is the theory and novel presentation of a higher-order synchronization method, we include these numerical experiments to demonstrate the potential for higher-order methods such as CHMP to be competitive with state-of-the-art pairwise group synchronization frameworks. 
All experiments will assume the setting of a \(3\)-hypergraph, unless otherwise stated, and \(\mathcal{C}\) is chosen to be \(\mathcal{C}_{n+1}^{n-1}\) as defined in Section \ref{subsec:CHMPalgorithm}. 
Further, the parameters of CHMP will be chosen such that \(\beta_0 = 1\) and \(\beta_t = 1.2^t\) for \(t = 1, \dots , 20\). 
This choice of parameter comes from \cite{lerman_robust_2022}, but the theory in Section \ref{subsec:exactrecovery} also aligns with this choice.

CHMP is tested on the groups \(SO(3)\) and \(SO(2)\). 
When \(G = SO(3)\), distance is measured by the geodesic distance on \(SO(3)\) given by:
\begin{equation}
    d_{SO(3)}(R_1,R_2) = \frac{1}{\pi}\arccos\left(\frac{\mathrm{trace}(R_1R_2^T) - 1}{2}\right),
\end{equation}
and for a metric on \(SO(3)^2\) the \(2\)-product metric is used:
\begin{equation}
    d_2((R_1,R_2),(\widetilde{R}_1,\widetilde{R}_2)) = \frac{1}{\sqrt{2}} \sqrt{d_{SO(3)}^2(R_1,R_2) + d_{SO(3)}^2(R_1,R_2)}.
\end{equation}
When \(G = SO(2)\), the group elements are represented as the set of angles modulo \(2\pi\) with the metric 
\begin{equation}
    d_{SO(2)}(\theta_1, \theta_2) = \frac{1}{\pi}|(\theta_1 - \theta_2)\! \mod (-\pi,\pi]|,
\end{equation}
and again use the \(2\)-product metric induced by \(d_{SO(2)}\) for \(SO(2)^2\).

In Section \ref{subsec:synthetic}, CHMP is demonstrated on synthetic data sets to show how the numerical performance aligns with our theory. 
Then CHMP is compared to CEMP and other standard synchronization methods for rotational (\(G = SO(3)\)) and angular (\(G = SO(2)\)) synchronization. 
In Section \ref{subsec:simulated}, an implementation of CHMP to recover the viewing angles from simulated cryo-EM images is presented and its performance is compared to the cryo-EM reconstruction package ASPIRE. 

The code for the algorithms and  experiments in this paper will be made available in the GitHub repository: 
\url{https://github.com/Addieduncan/CHMP}.

\subsection{Synthetic Experiments} \label{subsec:synthetic}

The experiments in this section demonstrate the convergence of CHMP and test its performance across the range of \(p\) and \(q\) values in the UCMH model defined in Section \ref{subsec:samplecomplexity}. 
All tests in this section use the group \(G = SO(3)\) where elements are represented as \(3 \times 3\) orthogonal matrices with determinant \(1\). 

\begin{figure}[htb]
    \centering
    \includegraphics[width=0.3\textwidth]{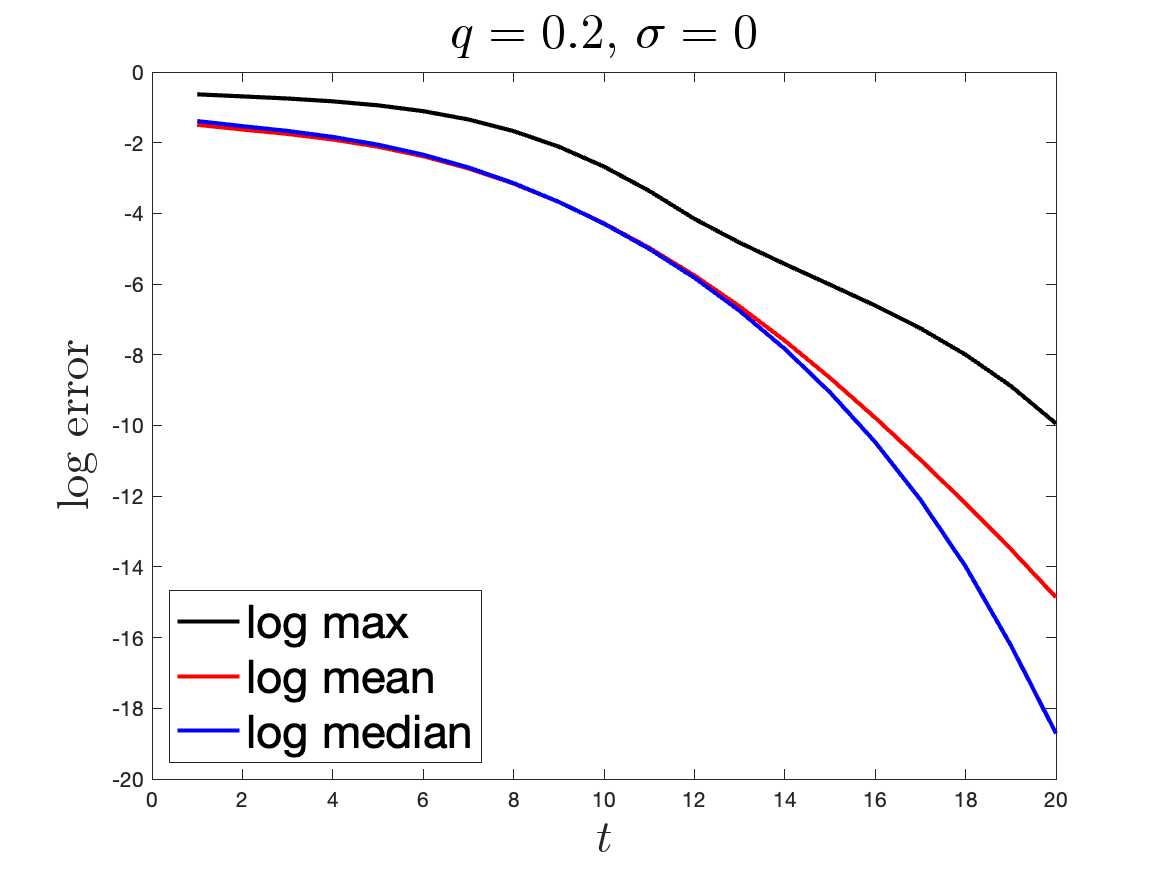} 
    \includegraphics[width=0.3\textwidth]{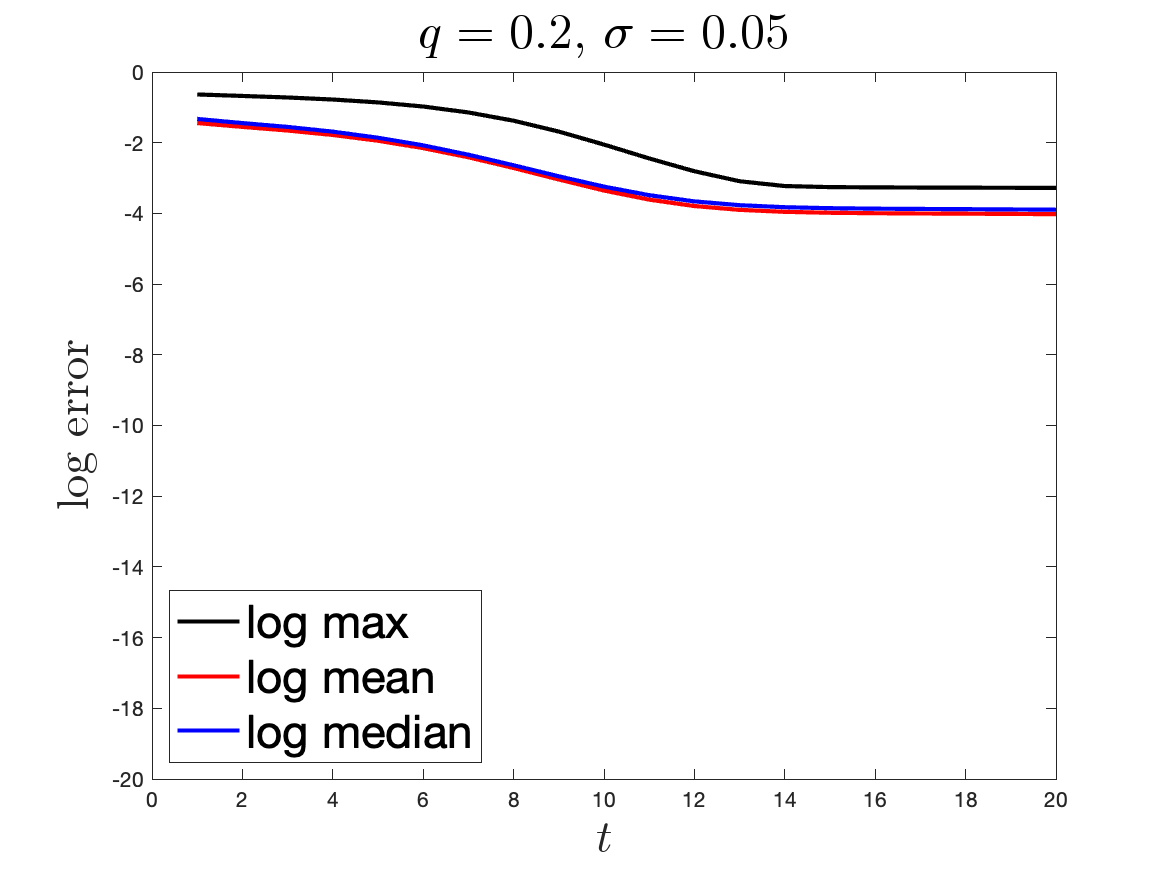}
    \includegraphics[width=0.3\textwidth]{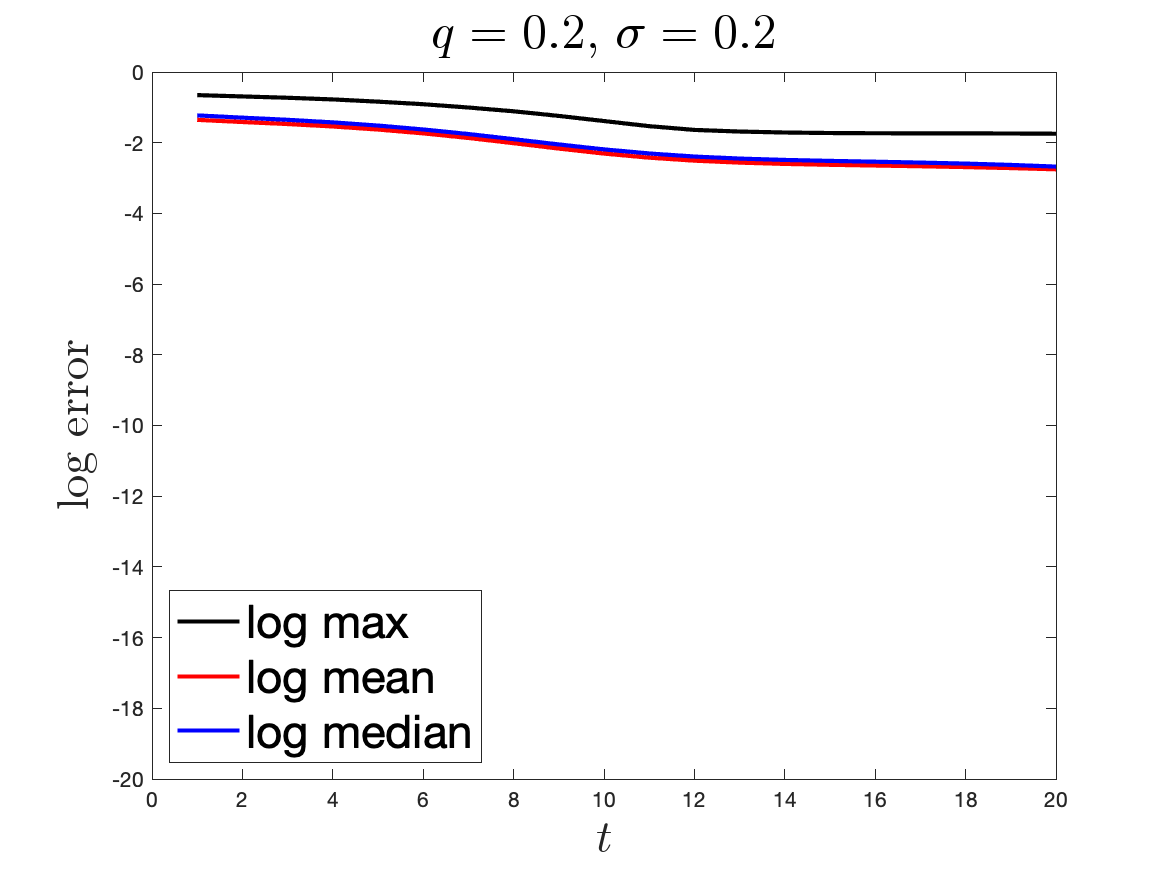} \\
    \caption{\centering Convergence results for CHMP under different levels of Gaussian noise.}
    \label{fig:convergence}
\end{figure}

Figure \ref{fig:convergence} shows the rate of convergence for CHMP. 
In each of these experiments, \(\mathcal{H}\) is generated using UCMH(\(3,50,1,0.2\)) with Gaussian noise on the uncorrupted measurements determined by a parameter \(\sigma\). 
In particular, noise is injected into \(\overline{\gamma}_{h}\), given as \((R_h^{(1)},R_h^{(2)}) \in SO(3)^2\), by first generating \(Q_h^{(1)}, Q_h^{(2)} \in \mathbb{R}^{3 \times 3}\) with i.i.d. Gaussian entries, then adding \(R_h^{(i)} + \sigma Q_h^{(i)}\), and finally projecting the resulting matrices back to \(SO(3)\) using SVD. 
Three noise levels are tested: \(\sigma = 0\) (noiseless) and \(\sigma = 0.05,0.2\) (noisy).
At each iteration the `log max', `log mean', and `log median' are recorded where 
\begin{align}
    &\log \mathrm{max}(t):= \ln\left(\max_{h \in H} |s_{h}(t) - s_h^*|\right)
    \\
    &\log \mathrm{mean}(t):= \ln\left(\frac{1}{|H|}\sum_{h \in H} |s_{h}(t) - s_h^*|\right)
    \\
    &\log \mathrm{median}(t):= \ln\left(\mathrm{median}\left(|s_{h}(t) - s_h^*| : h \in H\right)\right).
\end{align}
In the noiseless case Theorem \ref{thm:noiselessrecovery} predicts linear convergence, however in practice, the left most plot of Figure \ref{fig:convergence} demonstrates that the true convergence rate may be superlinear. 
The center and right plots of Figure \ref{fig:convergence} show the convergence for noisy data which gives a linear decay rate in the beginning that diminishes in the later iterations indicating convergence to a constant state. 
The experiments in these plots demonstrate the theory of Theorem \ref{thm:noisyrecovery}.

\begin{figure}[htb]
    \centering
    \includegraphics[width=0.3\textwidth]{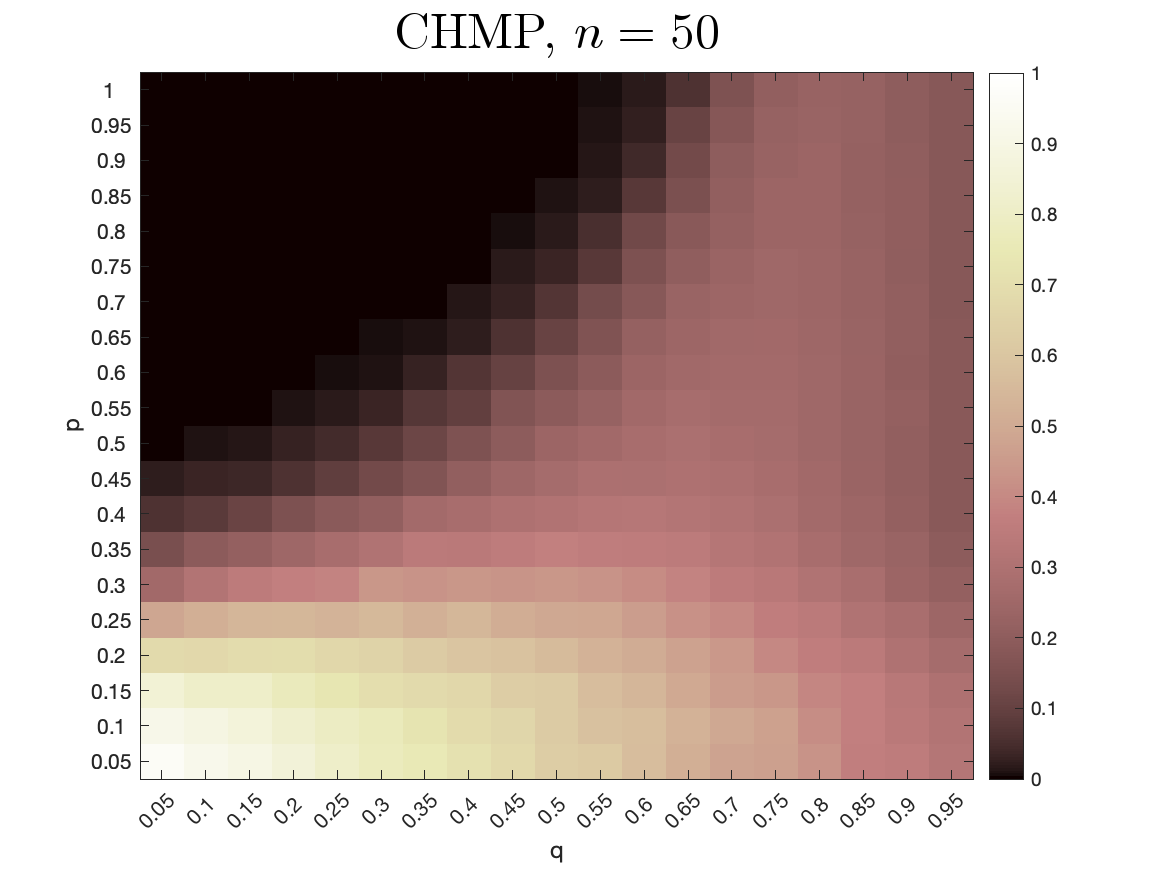}
    \includegraphics[width=0.3\textwidth]{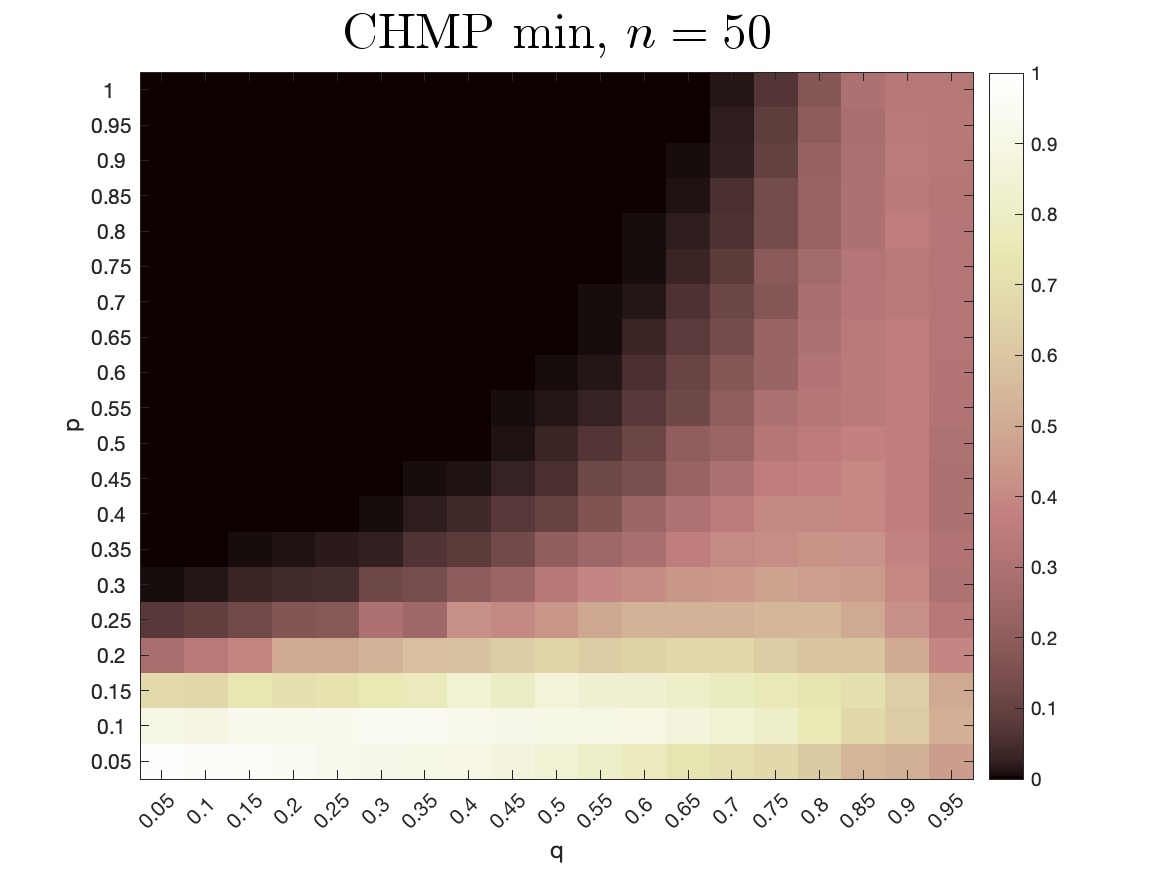} 
    \includegraphics[width=0.3\textwidth]{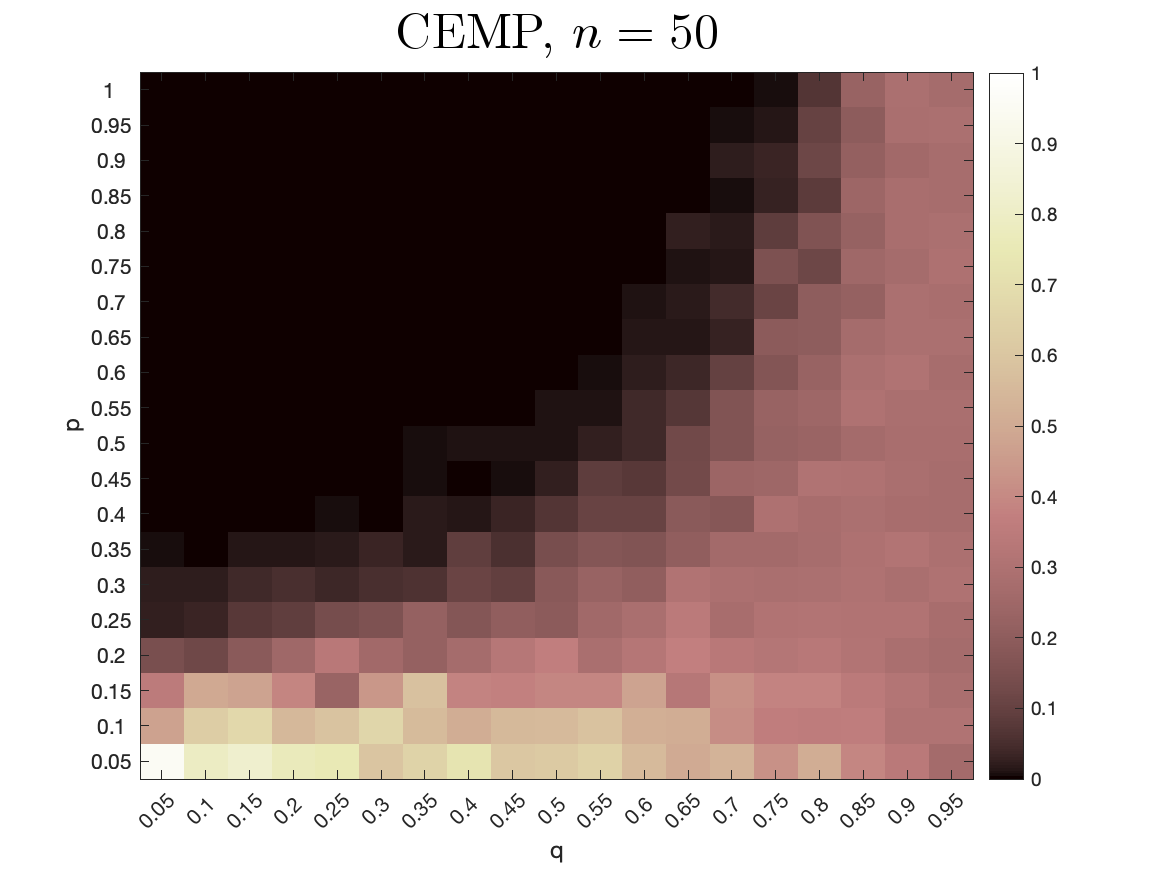} \\
    \caption{\centering Phase transitions plots for CHMP and CEMP.}
    \label{fig:transition}
\end{figure}

Figure \ref{fig:transition} presents phase transition plots for CHMP and CEMP under the {UCMH(\(3,50,p,q\))} and {UCMH(\(2,50,p,q\))} models, respectively. 
The error for CHMP (or CEMP as a special case of CHMP) is calculated by taking the average hyperedge corruption estimation error over all hyperedges:
\begin{equation}\label{eq:CHMPerror}
    \frac{1}{|H|} \sum_{h \in H} |s_h(T) - s_h^*|.
\end{equation}
For each value of \(0<p\leq 1\) and \(0\leq q<1\) the error is averaged from 10 runs in Figure \ref{fig:transition}. 
Darker values indicate lower error. 
The left most and right most plots give the results for CHMP and CEMP, respectively. 
In the center plot, a specialized error on the pairs of vertices \(i,j \in V\) such that \(i,j \in h\) for some \(h \in H\) is reported. 
Let \(V_h:= \{ \{i,j\} : i,j \in h\}\) and \(V_p : = \cup_{h \in H} V_h\). 
This special error is called the `CHMP min error' and is calculated by:
\begin{equation}\label{eq:minerror}
    \frac{1}{|V_p|} \sum_{\{i,j\} \in V_p} \left|\min_{h \ni i,j}(s_h(T)) - \min_{h \ni i,j}(s_h^*)\right|.
\end{equation}
Equation \eqref{eq:minerror} is meant to capture a notion of corruption estimation error that aligns with the output of Algorithm \ref{alg:refine}. 

Figure \ref{fig:transition} shows that given the same parameters \(m\), \(p\), and \(q\), CEMP achieves a lower error for estimating the (hyper)edge corruption levels than CHMP, especially in the more challenging domains where \(p < 0.5\) or  \(q_g < 0.5\). 
This behavior is predicted by Proposition \ref{prop:sample-complexity}. 
However, estimating the hyperedge corruption levels, as is the purpose of CHMP and CEMP, is only the first step in the process of solving the group synchronization problem. 
In the next section we will discuss the performance of CHMP+MST and CHMP+GCW.

\subsubsection{Rotational Synchronization} \label{subsubsec:rotation-synch}

In this section, CHMP + MST and CHMP + GCW, defined in Section \ref{sec:recovery}, are implemented to estimate rotations in \(SO(3)\). 
CHMP + MST will only be used when \(\sigma = 0\) since it is not recommended in noisy domains. 
Recall that a vertex potential can only be recovered up to a global action. 
To compare the recovered vertex potential with the ground truth, the two sets of rotations are globally aligned by solving the orthogonal Procrustes problem:
\begin{equation}\label{eq:procrustes}
    \min_{Q \in SO(3)} \frac{1}{m} \sum_{i = 1}^m\left\| R_i - R_i^* Q \right\|^2_F.
\end{equation}
The matrix \(Q\) can be found using SVD (See \cite{schonemann_generalized_1966}) and the rotation recovery error is reported as the Procrustes error from \eqref{eq:procrustes} which is averaged over each image.

\begin{figure}[htb]
    \centering
    \includegraphics[width=0.3\textwidth]{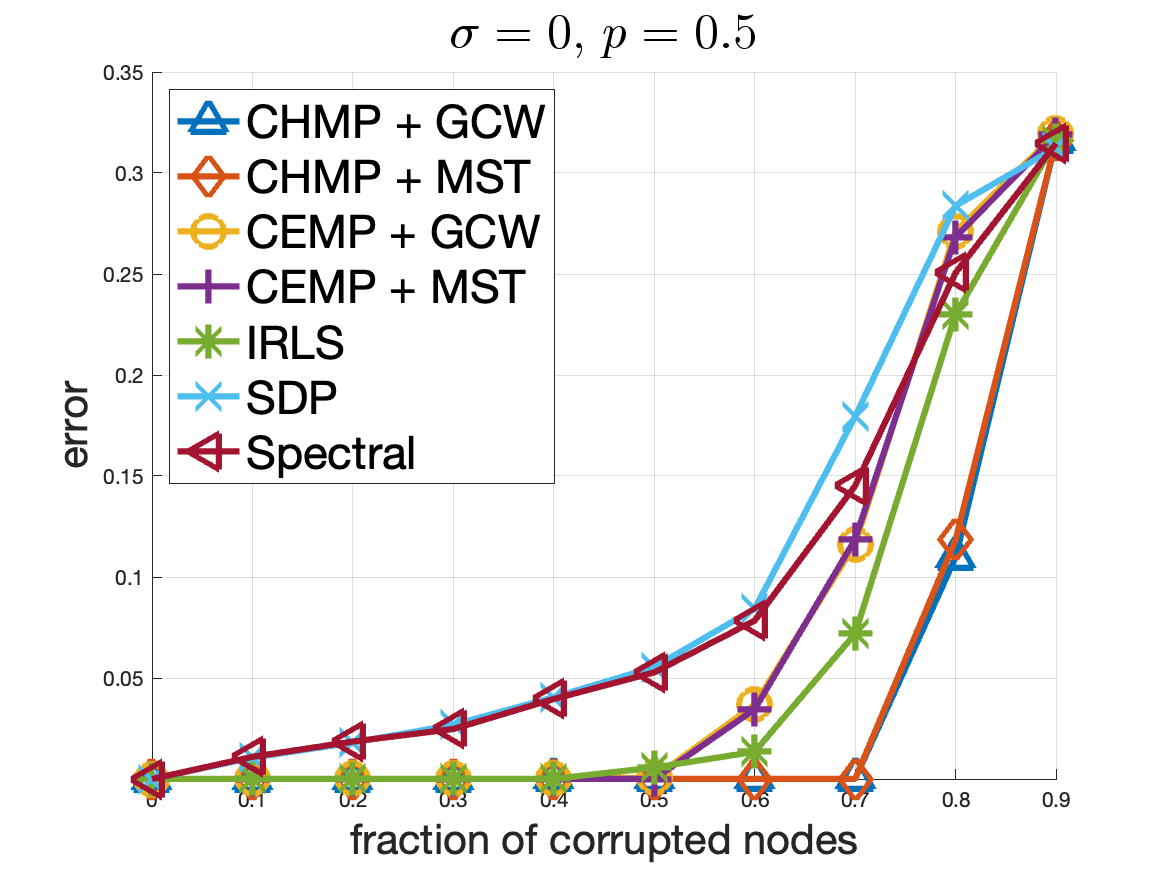}
    \includegraphics[width=0.3\textwidth]{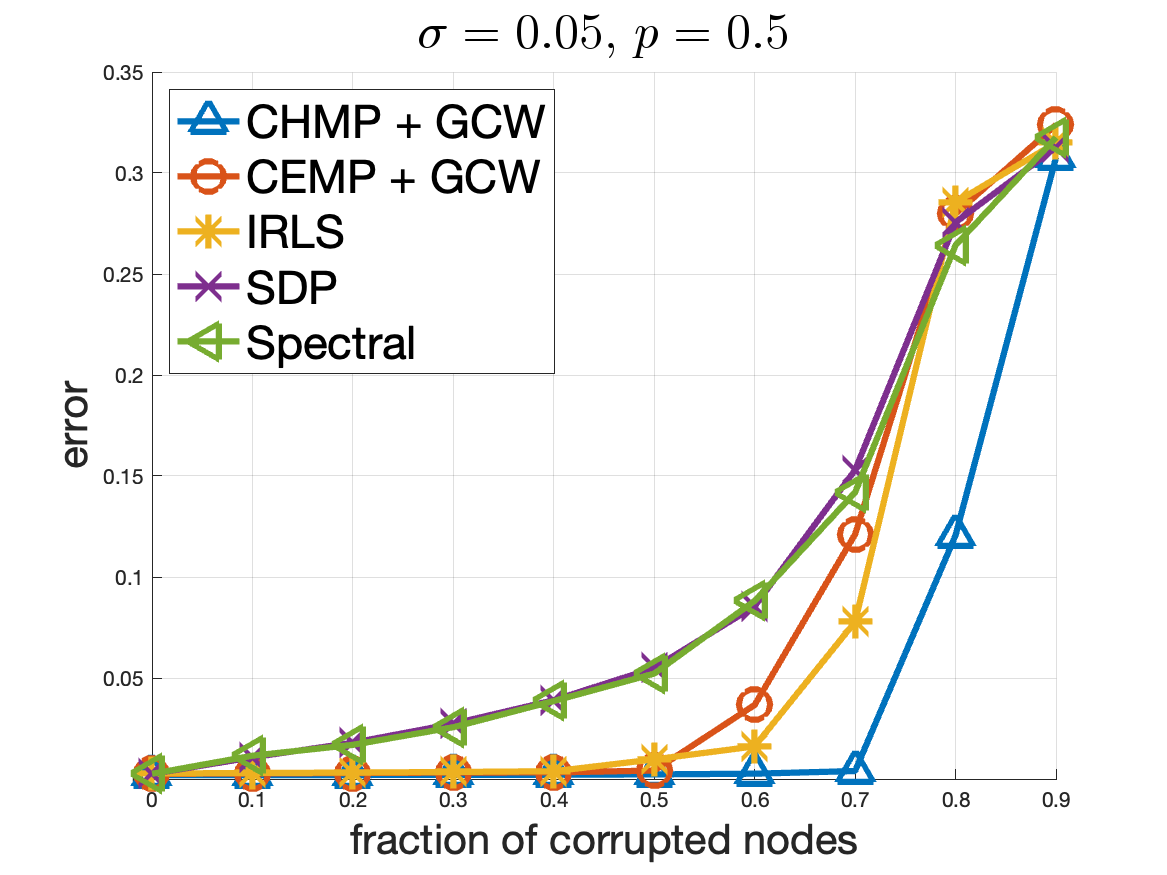} 
    \includegraphics[width=0.3\textwidth]{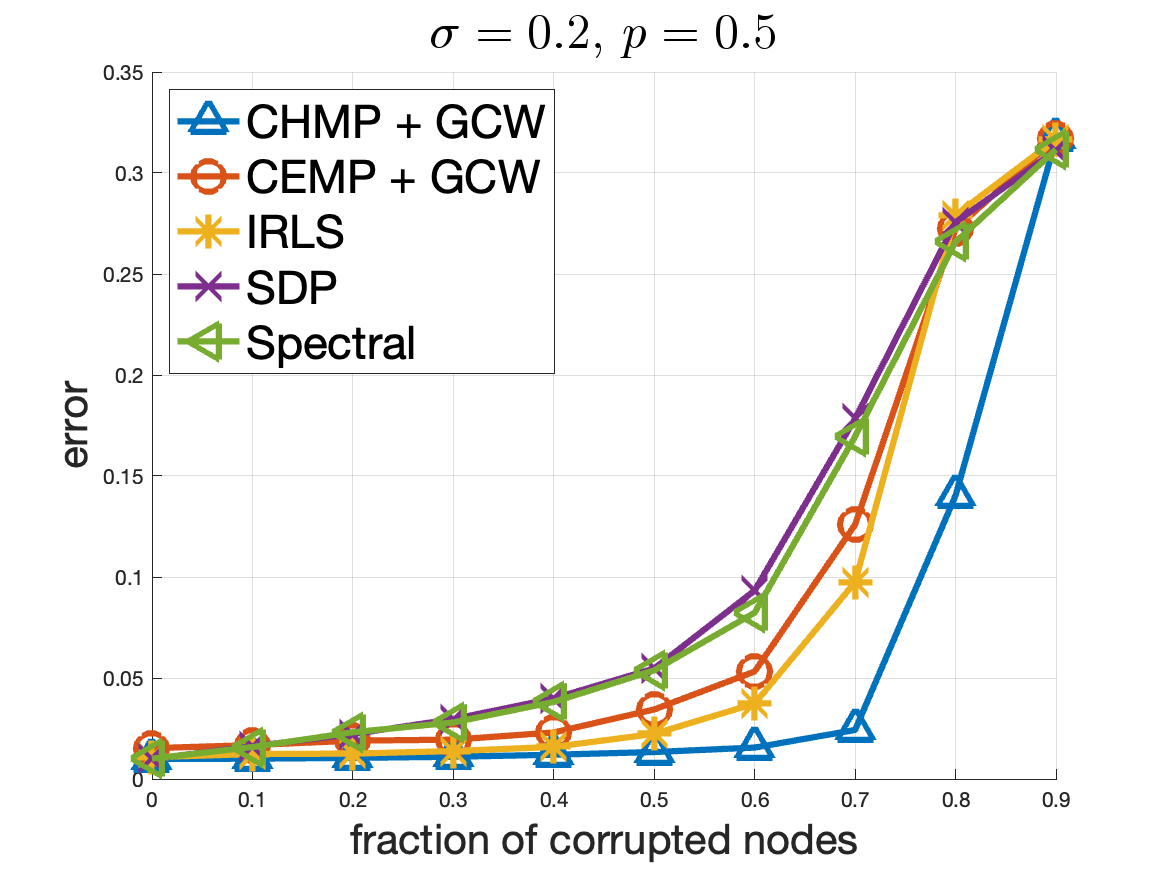}
    \includegraphics[width=0.3\textwidth]{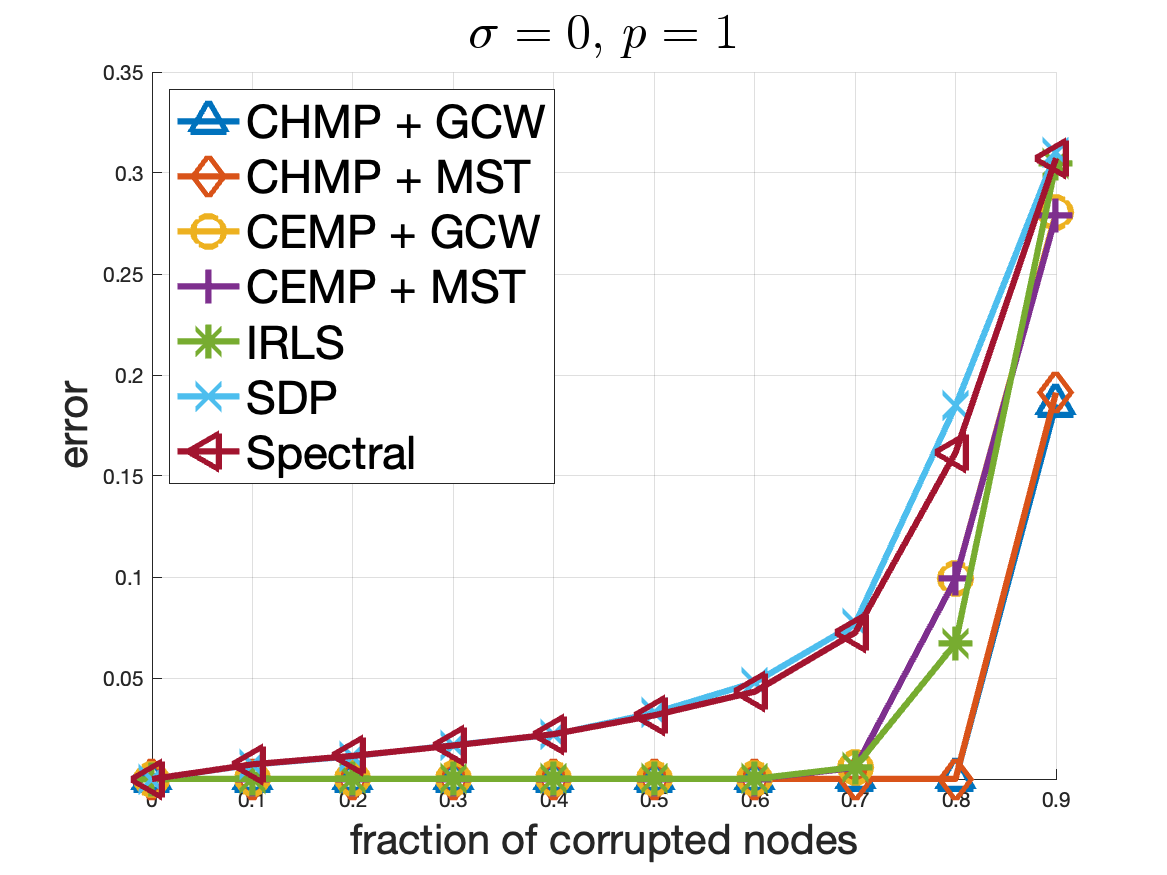}
    \includegraphics[width=0.3\textwidth]{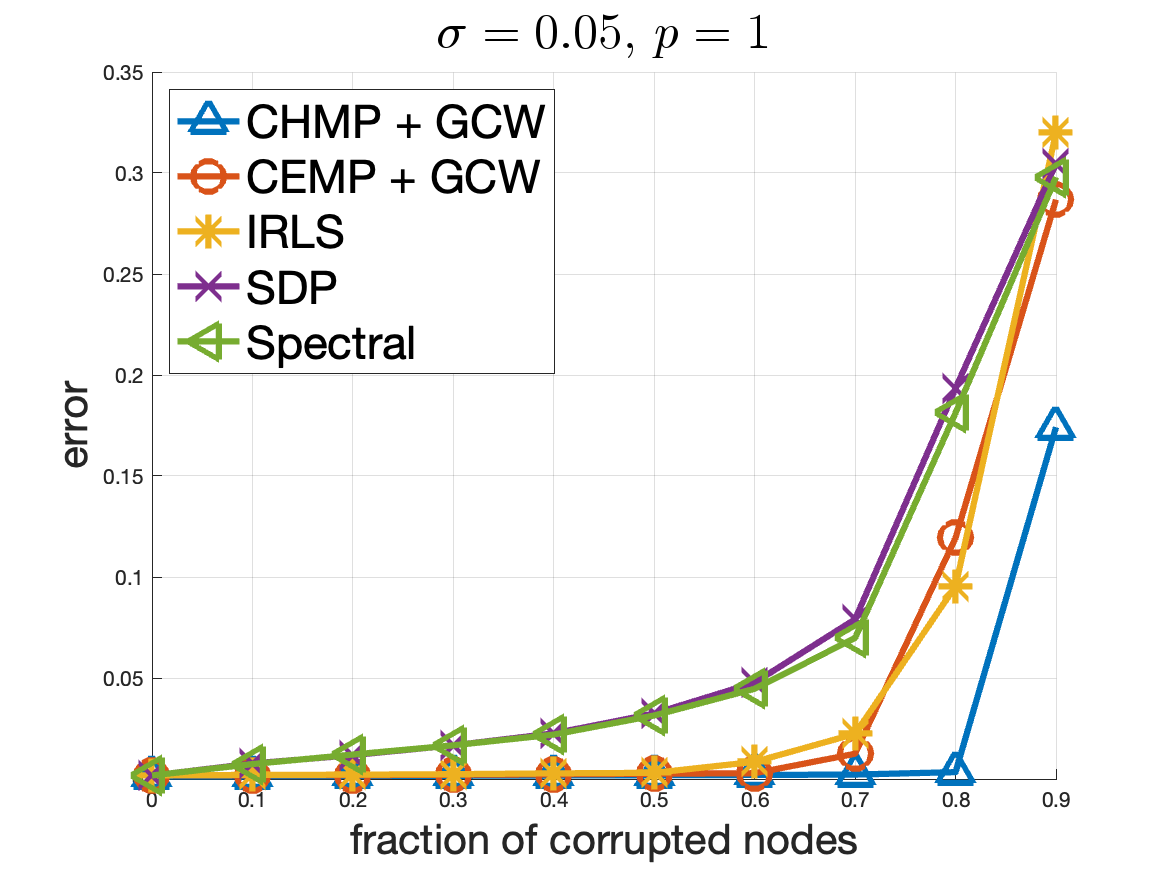} 
    \includegraphics[width=0.3\textwidth]{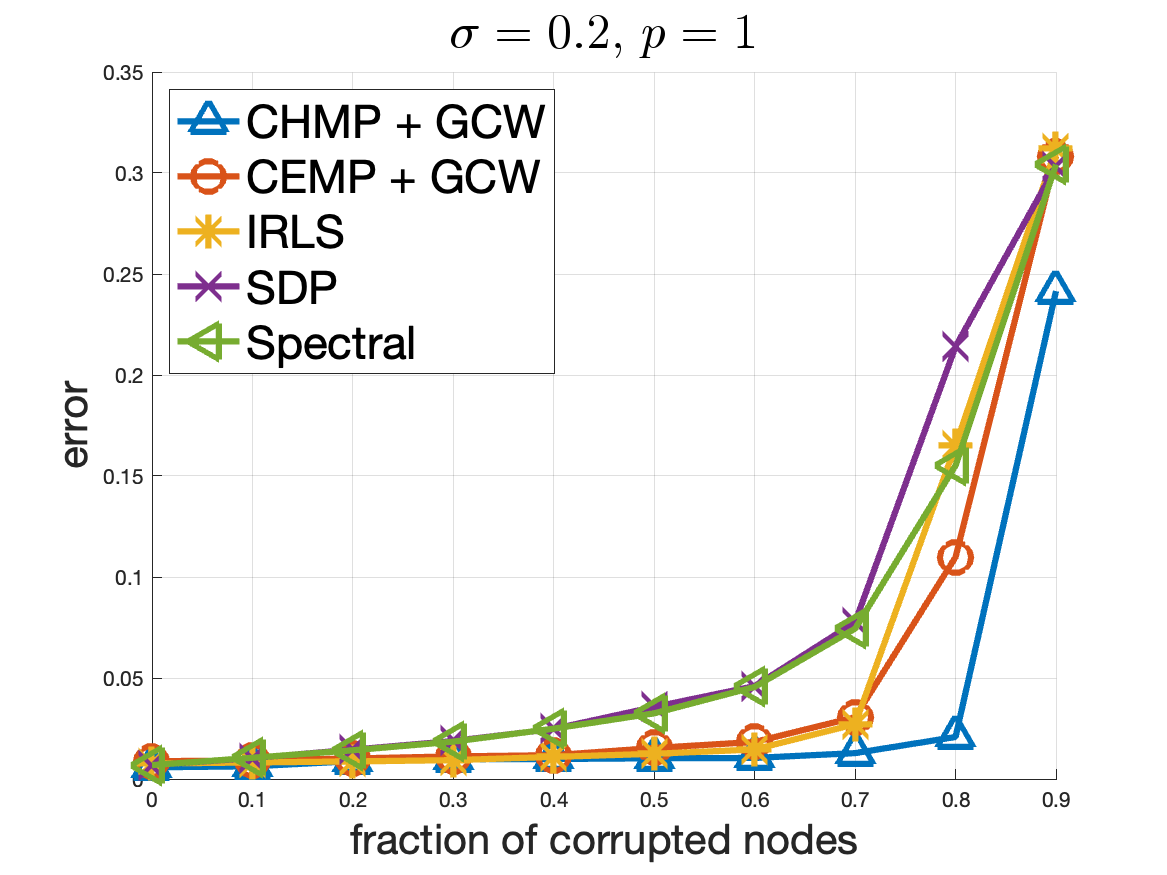}
    \\
    \caption{\centering Comparison of rotation recovery methods for \(G = SO(3)\) on data with different levels of corruption and Gaussian noise.}
    \label{fig:outlier}
\end{figure}

Figure \ref{fig:outlier} compares CHMP + MST and CHMP + GCW to CEMP+MST, CEMP+GCW, IRLS, SDP, and Spectral rotation synchronization methods. 
As stated previously, CEMP is a special case of CHMP. 
IRLS, SDP, and Spectral are standard synchronization methods with straightforward implementations. 
For more information about these methods for synchronization on \(SO(3)\) see \cite{lerman_robust_2022} and \cite{singer_structure_2011}. 
Our dataset is created by generating 50 random rotations. 
Some rotations are then corrupted according to UCMH(\(3,50,p,q\)) for the CHMP methods and UCMH(\(2,50,p,q\)) for the other methods. 
For each value of \(p\) and \(q\), Figure \ref{fig:outlier} reports the average Procrustes error over 10 runs. 
Each row of Figure \ref{fig:outlier} represents a different choice for \(p\), either \(p = 0.5\) or \(p=1\). 
Each column of Figure \ref{fig:outlier} represents different choices of noise, \(\sigma = 0.05, 0.2, 0.3\). 
The \(x\)-axis of each plot represents different values of \(q\). 

Both CHMP methods achieve exact recovery for corruption levels up to \(q = 0.7\) or \(q = 0.8\) depending on the density of the (hyper)graph. 
This is an improved corruption tolerance compared to the other methods tested. 
In every experiment (except at the highest level of corruption, where every method performs poorly) CHMP + GCW outperforms all other algorithms, and in particular CHMP demonstrates approximate recovery at higher levels of corruption compared to all other methods. 
These results show that despite the fact that the error for CHMP is typically higher than that of CEMP given the same parameters, the denoising effect of the estimates from CHMP through the use of Algorithm \ref{alg:refine} leads to better estimates for the vertex potential in the final recovery. 

Although we don't have a rigorous theory yet to support this behavior, the numerical results demonstrate the hypothesized advantages of using higher-order information directly in the synchronization process to exploit the increased redundancies and information given in the higher-order estimates. 
In particular, the CHMP min error reported in the center plot of Figure \ref{fig:transition} that aligns more closely to the phase transitions of CEMP and may help explain the favorable performance of CHMP in a full vertex recovery pipeline. 
In Appendix \ref{appendix:angular-synch} these experiments are repeated for the angular synchronization problem, where \(G = SO(2)\). Similar conclusions about the comparison between CHMP+GCW and CHMP+MST and the other methods can be drawn from the experiments on \(SO(2)\).

However, it is important to note that the underlying UCMH models in these experiments are fundamentally different for CHMP (with \(n \geq 3\)) and CEMP and while they demonstrate the potential advantages of a higher-order synchronization framework, they are not a true apples to apples comparison. 
In fact, such a comparison is difficult to conceive since the input of the algorithms is fundamental different. 

In an attempt to compare CHMP and CEMP on the same input data, we consider a higher-order data set which can be passed to CEMP+MST or CEMP+GCW by reducing the data to the \(2\)-section of the underlying hypergraph. 
Since this refinement is done before applying CHMP to the data, we need some way to determine how data from two hyperedges overlapping on a pair of vertices will be reduced to data on an edge between the pair of vertices. 
Since the noise model for our synthetic data allows for high levels of corruption we use the geodesic medoid to compute the reduction. 
For example, if \(ij\) is an edge in the \(2\)-section of the underlying hypergraph \(\mathcal{H}(V,H)\) of the hyperedge potential, let \(H_{ij} := \{h \in H : \{i,j\} \subseteq h\}\).
Further let 
\[
    \Gamma_{ij} := \left\{\widehat{\gamma}_{h} = \tau \circ \operatorname{Res}_{h \to \{i,j\}}(\gamma_h)\right\}_{h \in H_{ij}} \subseteq SO(3)
\]
be the set of hyperedge potentials for hyperedges in \(H_{ij}\) restricted to the pair \(i,j\). 
The geodesic medoid of this set is 
\[
    R^{\textrm{med}}_{ij} = \arg\min_{\widehat{\gamma} \in \Gamma_{ij}} \sum_{\widehat{\gamma}_h \in \Gamma_{ij}} d_{SO(3)}(\widehat{\gamma}, \widehat{\gamma}_h).
\]
Then \(\overline{\gamma}_{ij}\) is assigned \(R^{\textrm{med}}_{ij}\) in the reduction to the \(2\)-section.

\begin{figure}[htb]
    \centering
    \includegraphics[width=0.3\textwidth]{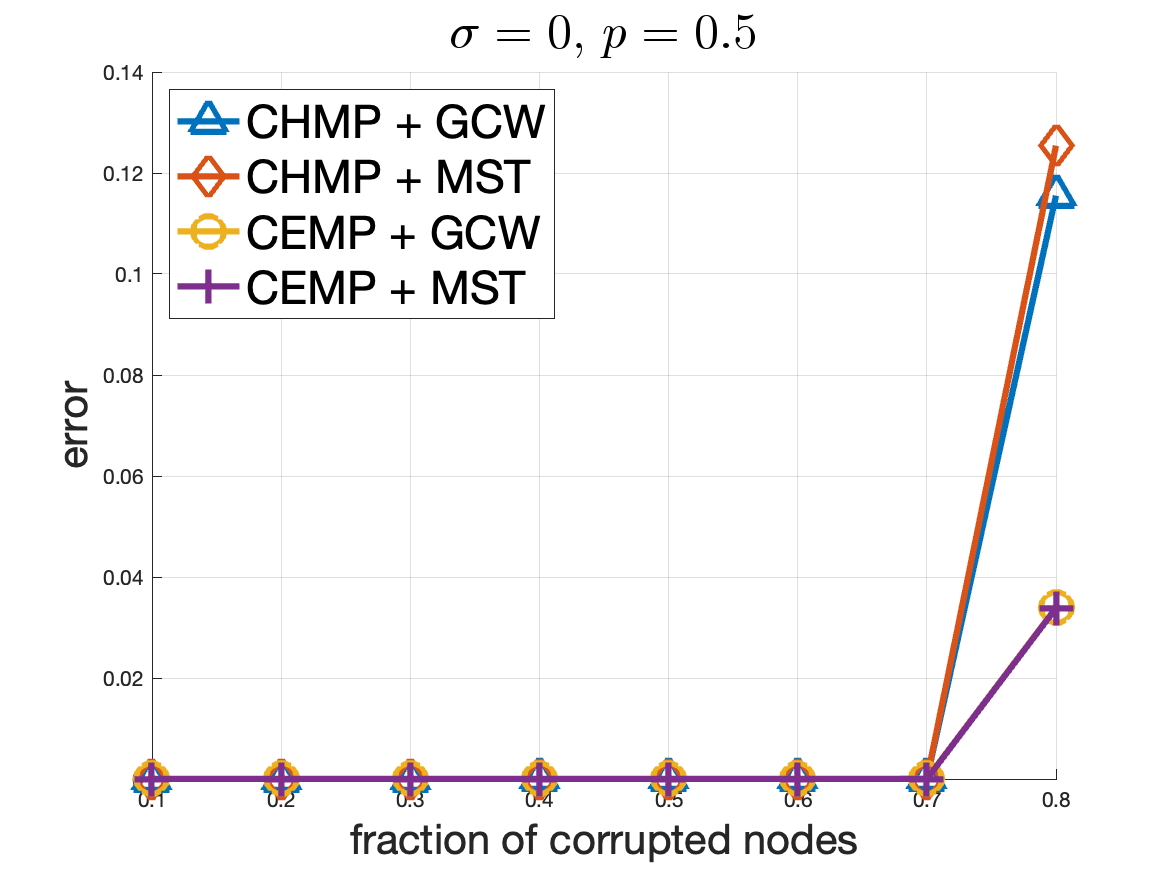}
    \includegraphics[width=0.3\textwidth]{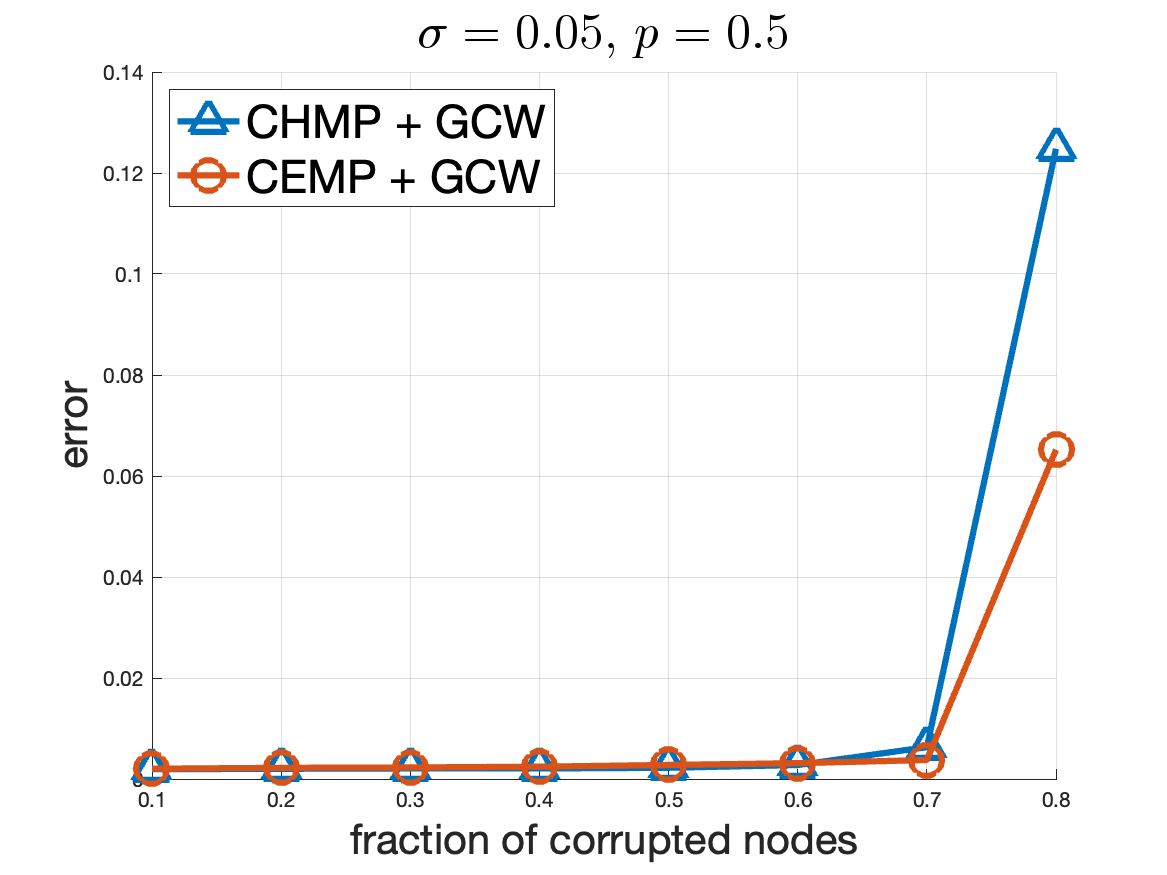} 
    \includegraphics[width=0.3\textwidth]{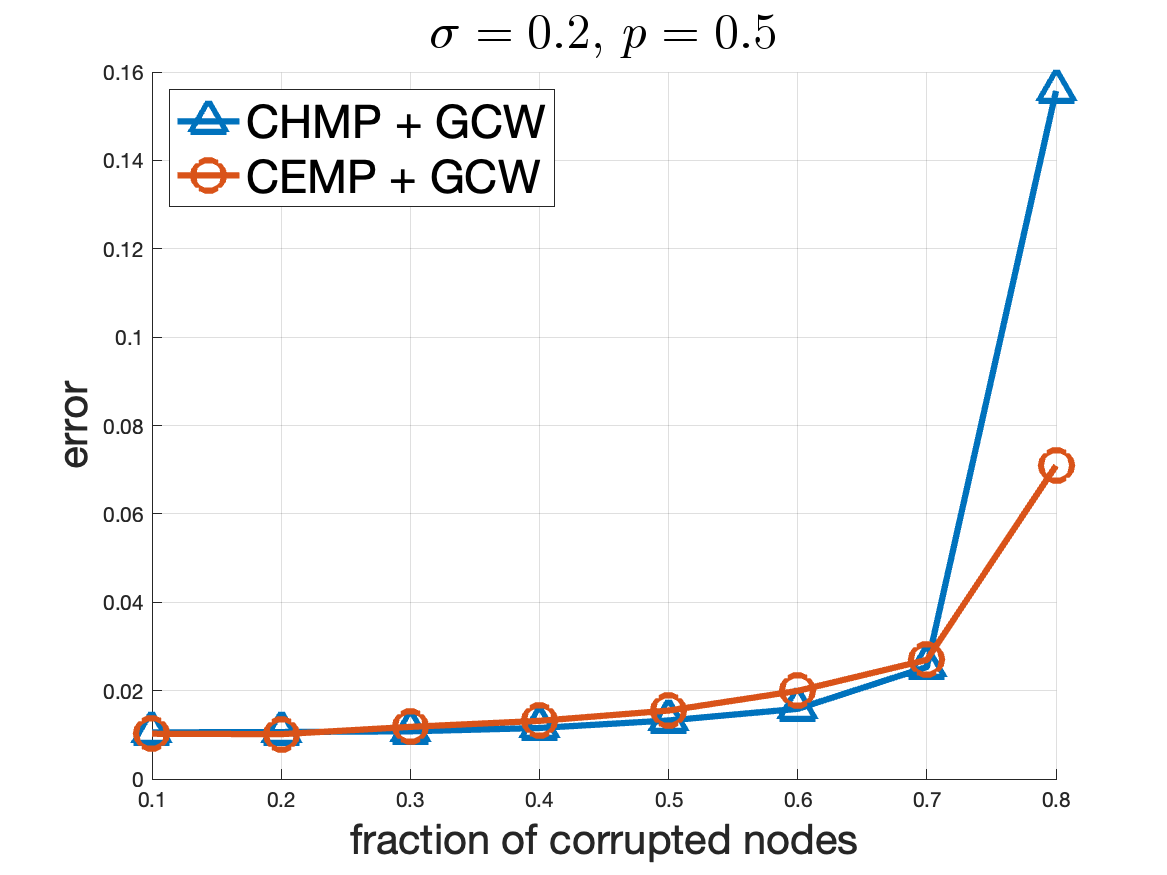}
    \\
    \includegraphics[width=0.3\textwidth]{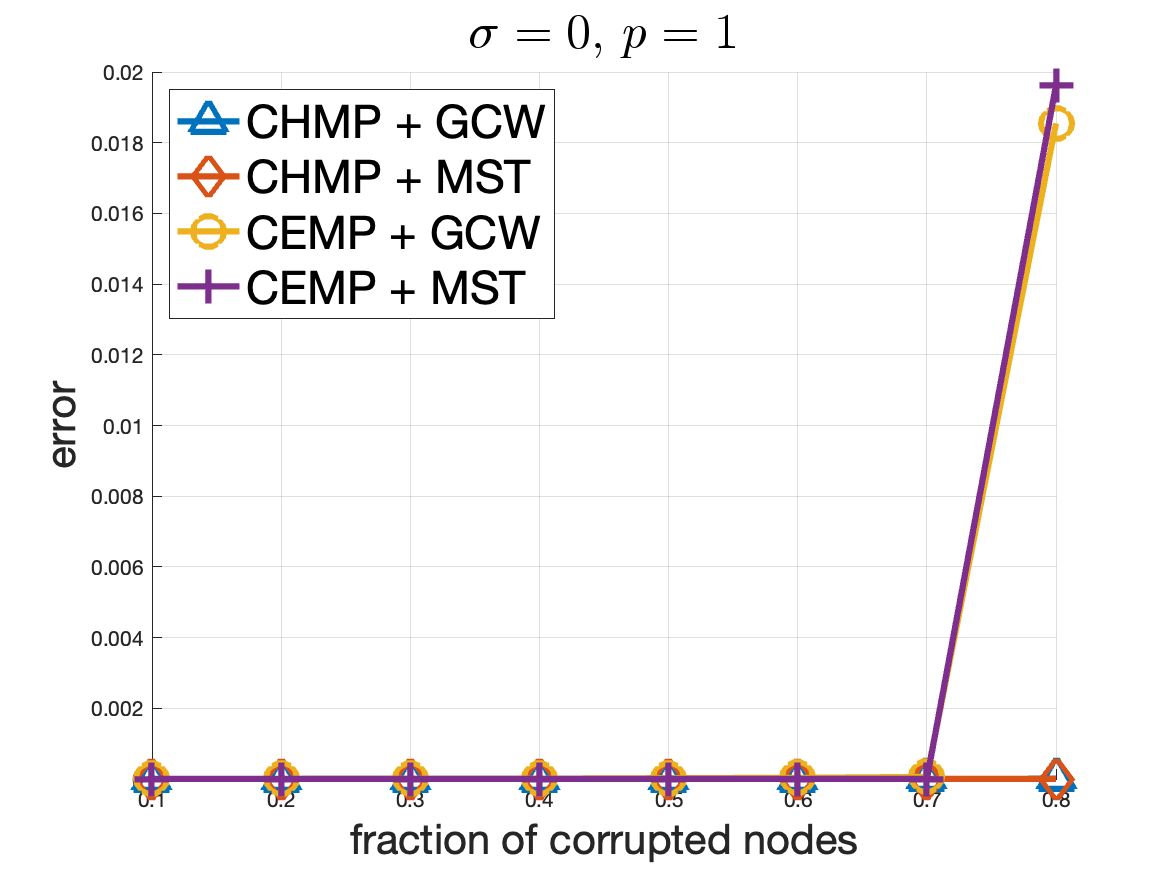}
    \includegraphics[width=0.3\textwidth]{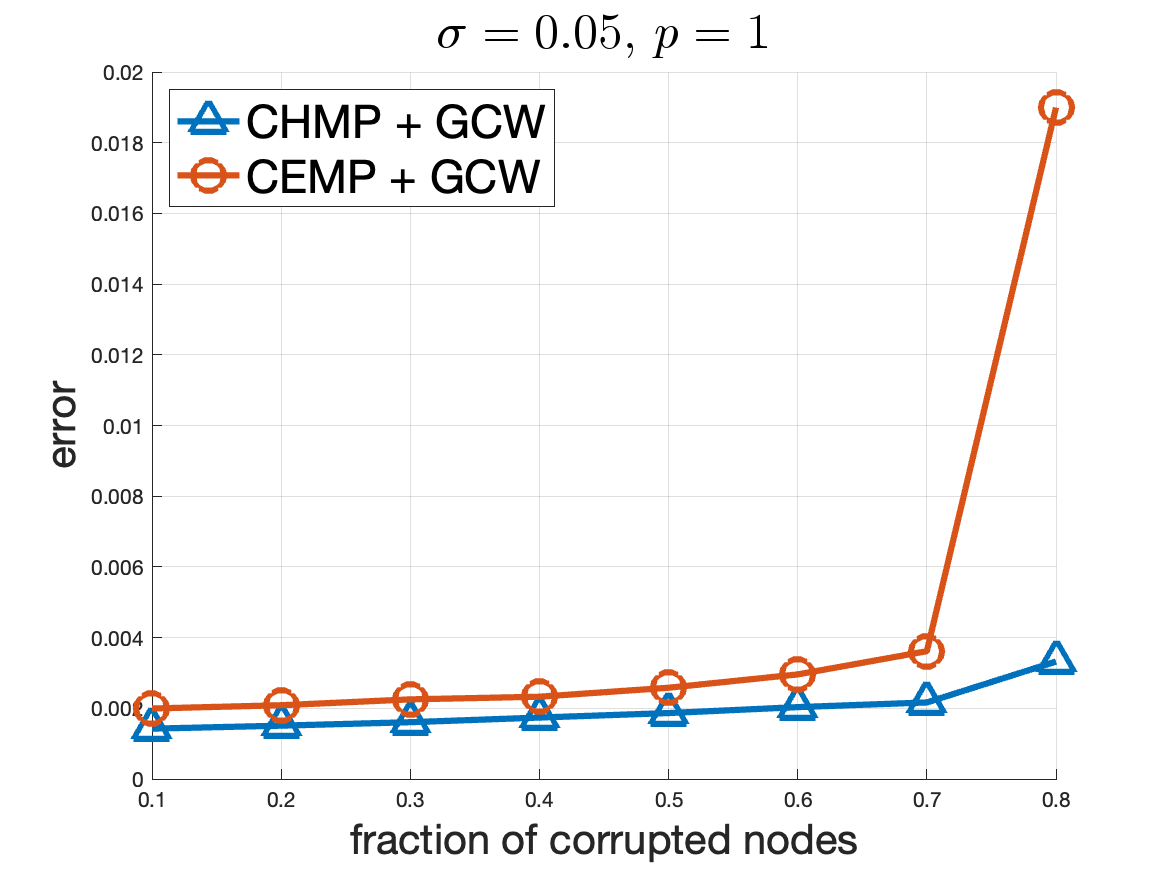} 
    \includegraphics[width=0.3\textwidth]{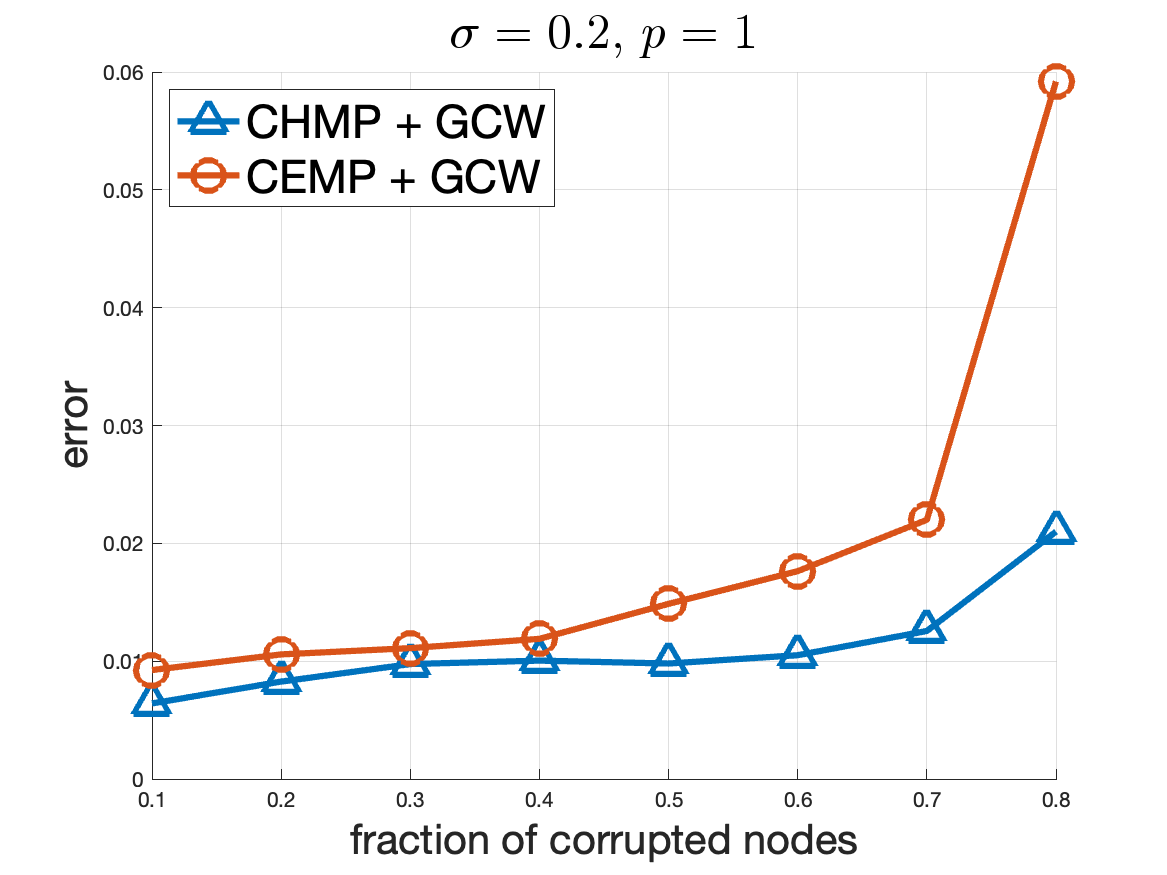}
    \\
    \caption{\centering Comparison of CHMP and CEMP where the underlying data comes from UCMH\((3,50,p,q)\) with Gaussian noise \(\sigma\).}
    \label{fig:CEMPmedoid}
\end{figure}

To construct the dataset we choose \(G = SO(3)\), and generate \(50\) random rotations according to UCMH\((3,50,p,q)\) with Gaussian noise according to \(\sigma\). 
The the rotations are recovered either through a CHMP pipeline (CHMP + MST for \(\sigma = 0\) or CHMP+GCW for \(\sigma \geq 0\)) or through a CEMP pipeline by first reducing the triple wise hyperedge potential to a pairwise edge potential using the geodesic medoid as described above, then applying CEMP+MST (\(\sigma = 0\)) or CEMP+GCW (\(\sigma \geq 0\)). 
Figure \ref{fig:CEMPmedoid} gives the results of the experiment where the rotation recovery error is reported as the Procrustes error in \eqref{eq:procrustes}. Rows of Figure \ref{fig:CEMPmedoid} correspond to \(p = 0.5\) and \(p=1\) while the columns correspond to \(\sigma = 0, 0.05,\) and \(0.2\).
In these experiments CHMP and CEMP both achieve exact recovery for low levels of corruption. For sparse datasets (\(p = 0.5)\)) with high corruption, CEMP outperforms CHMP. However for dense datasets (\(p = 1\)), CHMP outperforms CEMP when \(q\) is large and in particular can achieve exact recovery for larger values of \(q\).

\subsubsection{Runtime Comparisons}

Tables \ref{tbl:runtime1} and \ref{tbl:runtime0.5} compare the runtime of CHMP + GCW against the other methods considered above.
The tests are performed on \(m = 50\) rotations from \(G = SO(3)\) with \(\sigma = 0.05\) noise and \(q = 0.2\) corruption probability. 
Table \ref{tbl:runtime1} tests the runtime for a dense (hyper)graph, when \(p = 1\), and Table \ref{tbl:runtime0.5} tests a sparser (hyper)graph, when \(p = 0.5\). 

\begin{table}[htb] 
    \centering
    \begin{tabular}{|c|c|c|c|}
        \hline
        Algorithm  & Initialization Time & Iteration Time & Total Time \\ \hline
        CHMP + GCW & 79.0503             & 0.2994         & 79.577     \\ \hline
        CEMP + GCW & 0.0305              & 0.0156         & 0.0484     \\ \hline
        IRLS       & 0.0222              & 0.0334         & 0.0557     \\ \hline
        SDP        & 0.0221              & 2.0703         & 2.0943     \\ \hline
        Spectral   & 0.0293              & 0.0022         & 0.0324     \\ \hline
    \end{tabular}
    \caption{\centering Runtime in seconds for \(m=50\) rotations and \(p = 1\).}
    \label{tbl:runtime1}
\end{table}

\begin{table}[htb] 
    \centering
    \begin{tabular}{|c|c|c|c|}
        \hline
        Algorithm  & Initialization Time & Iteration Time & Total Time \\ \hline
        CHMP + GCW & 17.7359             & 0.077          & 17.9451    \\ \hline
        CEMP + GCW & 0.0141              & 0.0067         & 0.024      \\ \hline
        IRLS       & 0.0142              & 0.0219         & 0.0361     \\ \hline
        SDP        & 0.0137              & 2.3359         & 2.3516     \\ \hline
        Spectral   & 0.0141              & 0.0017         & 0.0166     \\ \hline
    \end{tabular}
    \caption{\centering Runtime in seconds for \(m=50\) rotations and \(p = 0.5\).}
    \label{tbl:runtime0.5}
\end{table}

The total time of the full pipeline is reported along with the runtime for the initialization and iteration phases. 
In Section \ref{subsec:CHMPalgorithm}, the complexity per iteration of CHMP for \(n = 3\) was stated to be of the order \(O(m^4)\) compared the complexity of CEMP per iteration which is \(O(m^3)\). 
In practice the iteration runtime of CHMP for \(n = 3\) is still quite tractable and that the main computational bottleneck comes from the initialization step that generates CHG. 
This is due to the computational challenge of enumerating the cycles of the hypergraph and if one knew apriori the list of cycles, the runtime could be significantly improve. 
It is also of note that the iteration runtime of CHMP is still faster than the iteration time for SDP.

\subsection{Simulated Rotation Recovery Using Common Lines in Cryo-EM} \label{subsec:simulated}

In this section,  CHMP + GCW is tested on simulated cryo-EM images. 
We choose to simulate the images since it allows us to have an established ground truth set of rotations, tough in typical cryo-EM pipelines synchronization is applied to class averages rather than directly to the projection images \cite{fan_representation_2021, frank_electron_2006}. 
Three set of \(30\) simulated projection images are generated each from a different known molecular structure in the Electron Microscopy Data Bank \cite{lawson_unified_2015}. 
The molecular structures considered are EMD-2858, EMD-2811, and EMD-4214 which correspond to the yeast 80S, 60S, and 40S ribosomes, respectively.

\begin{figure}[htb]
    \centering
    \includegraphics[width=0.2\textwidth]{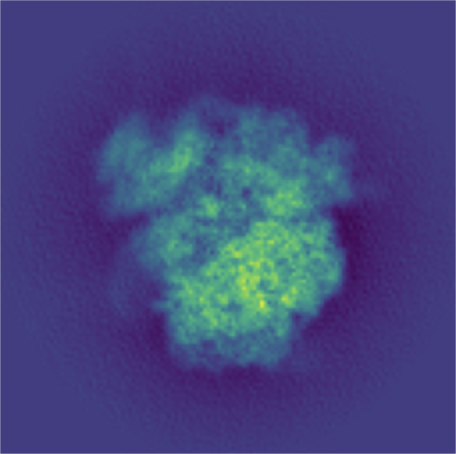} 
    \includegraphics[width=0.2\textwidth]{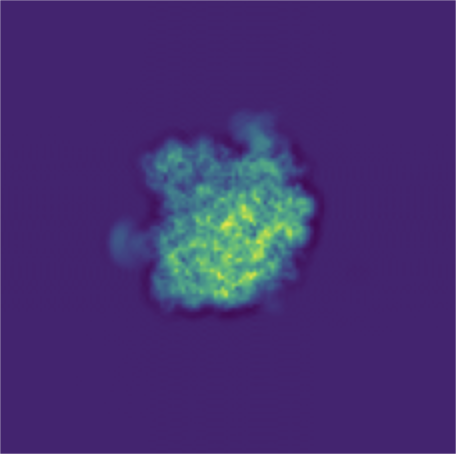}
    \includegraphics[width=0.2\textwidth]{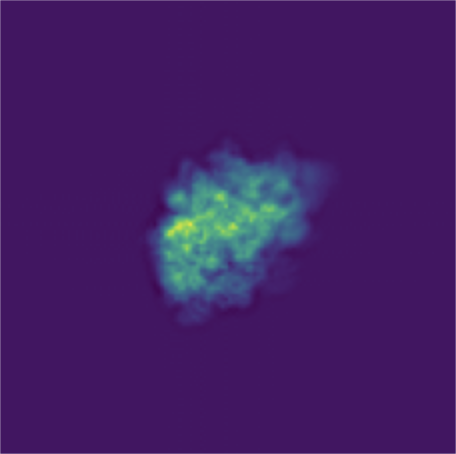} 
    \\
    \caption{\centering Example simulated cryo-EM projection images EMD-2858 (left), EMD-2811 (middle) and EMD-4214 (right).}
    \label{fig:simulatedclean}
\end{figure}

\begin{figure}[htb]
    \centering
    \includegraphics[width=0.19\textwidth]{figures/EMD2811_clean.png} 
    \includegraphics[width=0.19\textwidth]{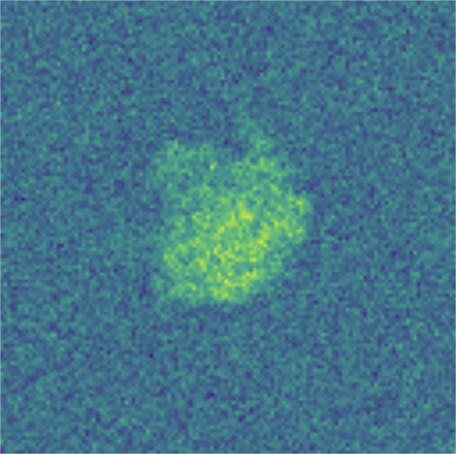} 
    \includegraphics[width=0.19\textwidth]{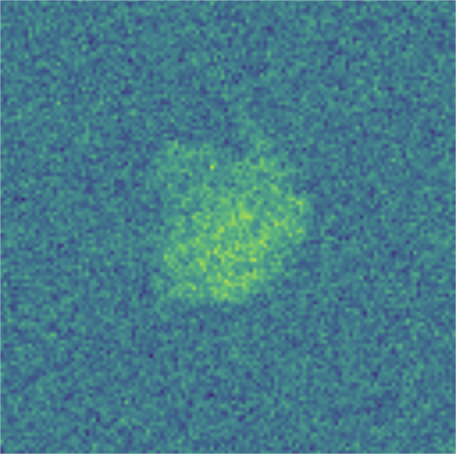}
    \includegraphics[width=0.19\textwidth]{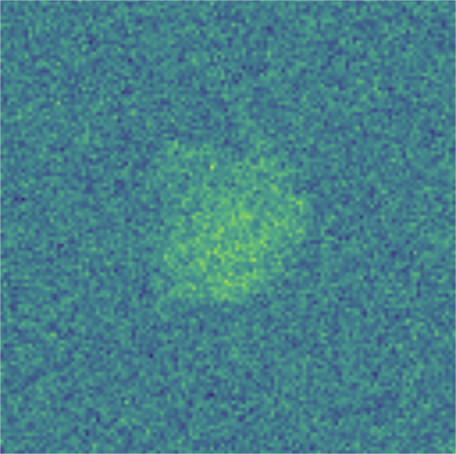} 
    \includegraphics[width=0.19\textwidth]{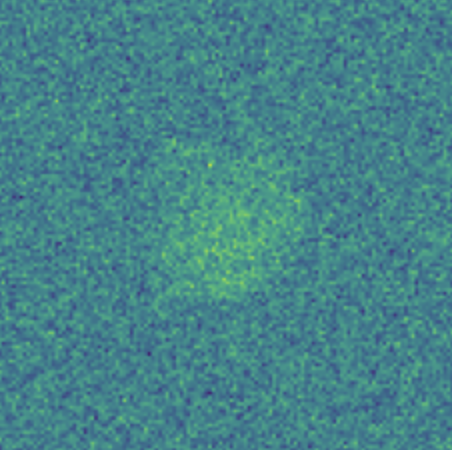} 
    \\
    \caption{\centering Example simulated projection images for EMD-2811. From left to right the simulated images have SNR \(= \infty, 1, 0.5, 0.25, 0.125\).}
    \label{fig:simulatedSNR}
\end{figure}

Using ASPIRE (\cite{wright_aspire_2023}), \(30\) projection images are simulated from randomly sampled viewing angles. Example simulated images are given in Figure \ref{fig:simulatedclean}.
Then white Gaussian noise is added to the images according to a specified signal-to-noise ratio (SNR). Figure \ref{fig:simulatedSNR} shows example simulated images for different levels of SNR.

\begin{figure}[htb]
    \centering
    \includegraphics[width=0.3\textwidth]{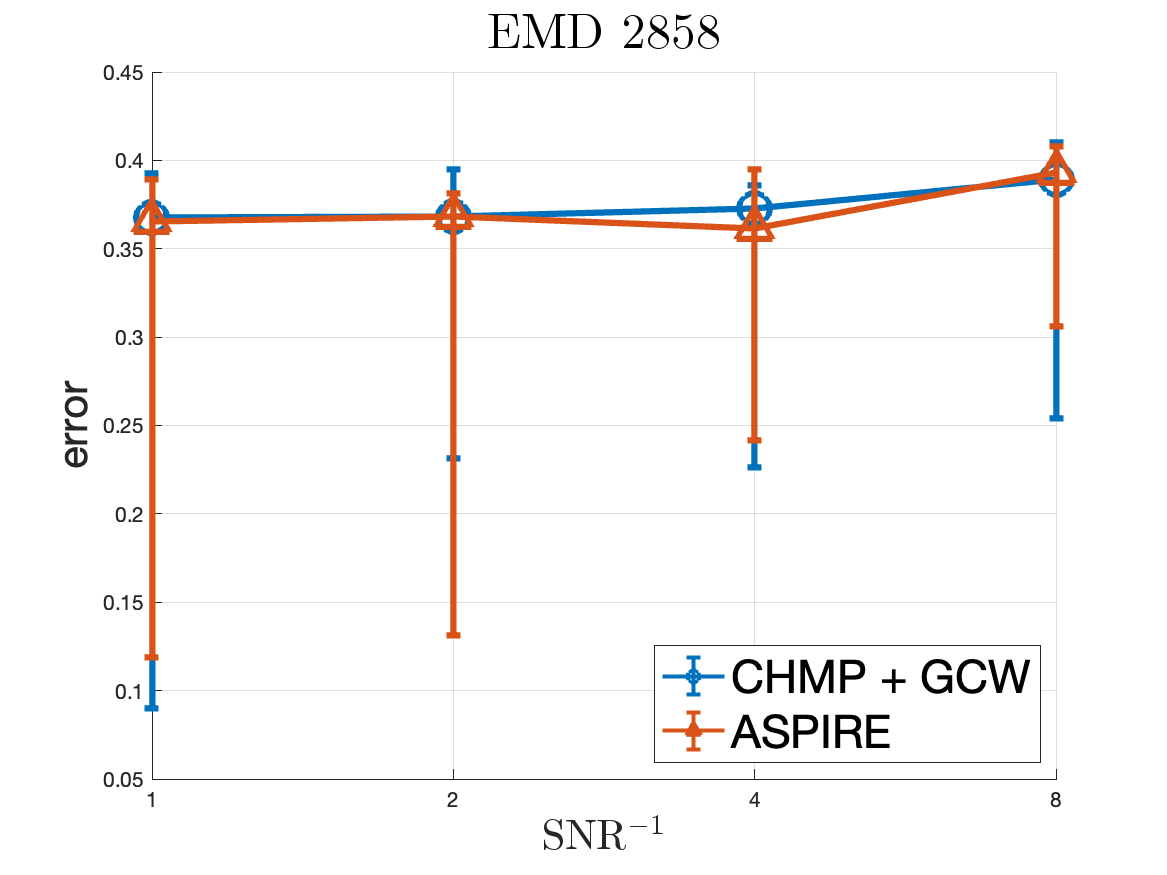} 
    \includegraphics[width=0.3\textwidth]{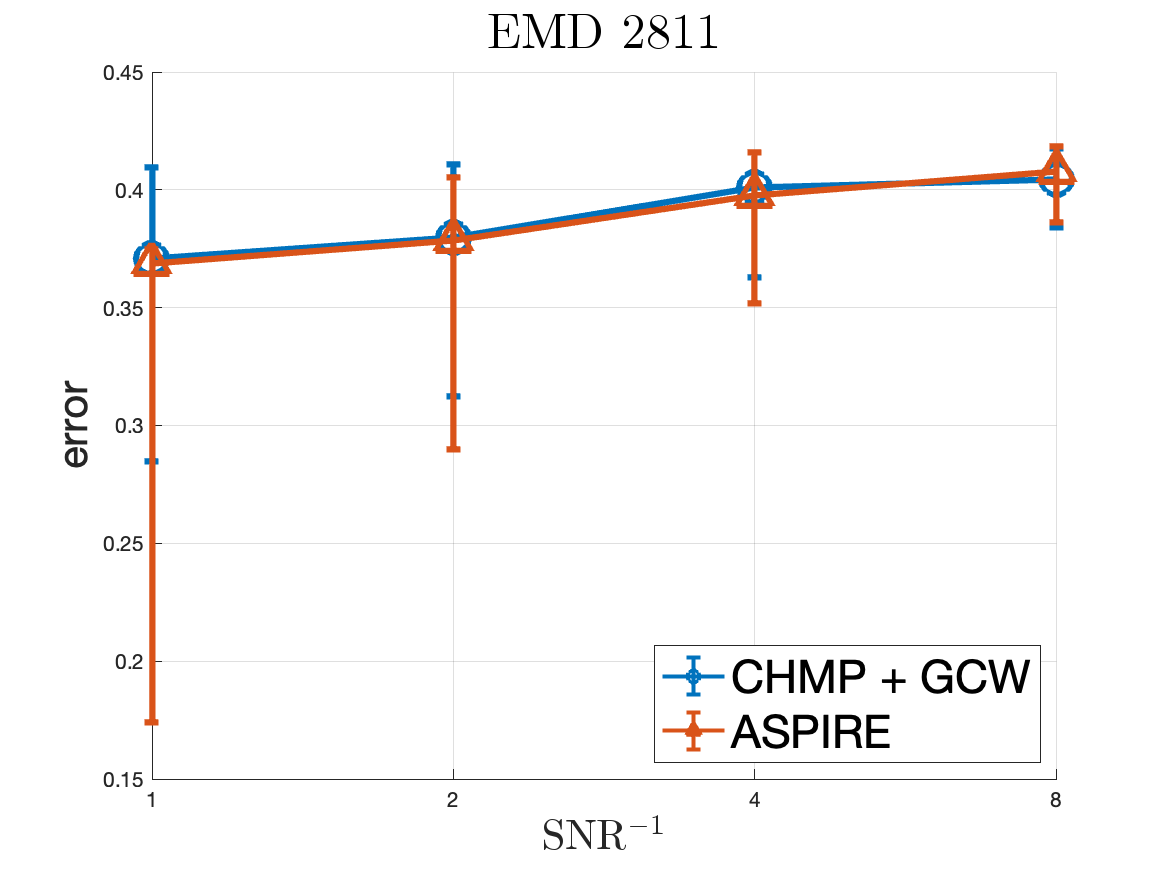}
    \includegraphics[width=0.3\textwidth]{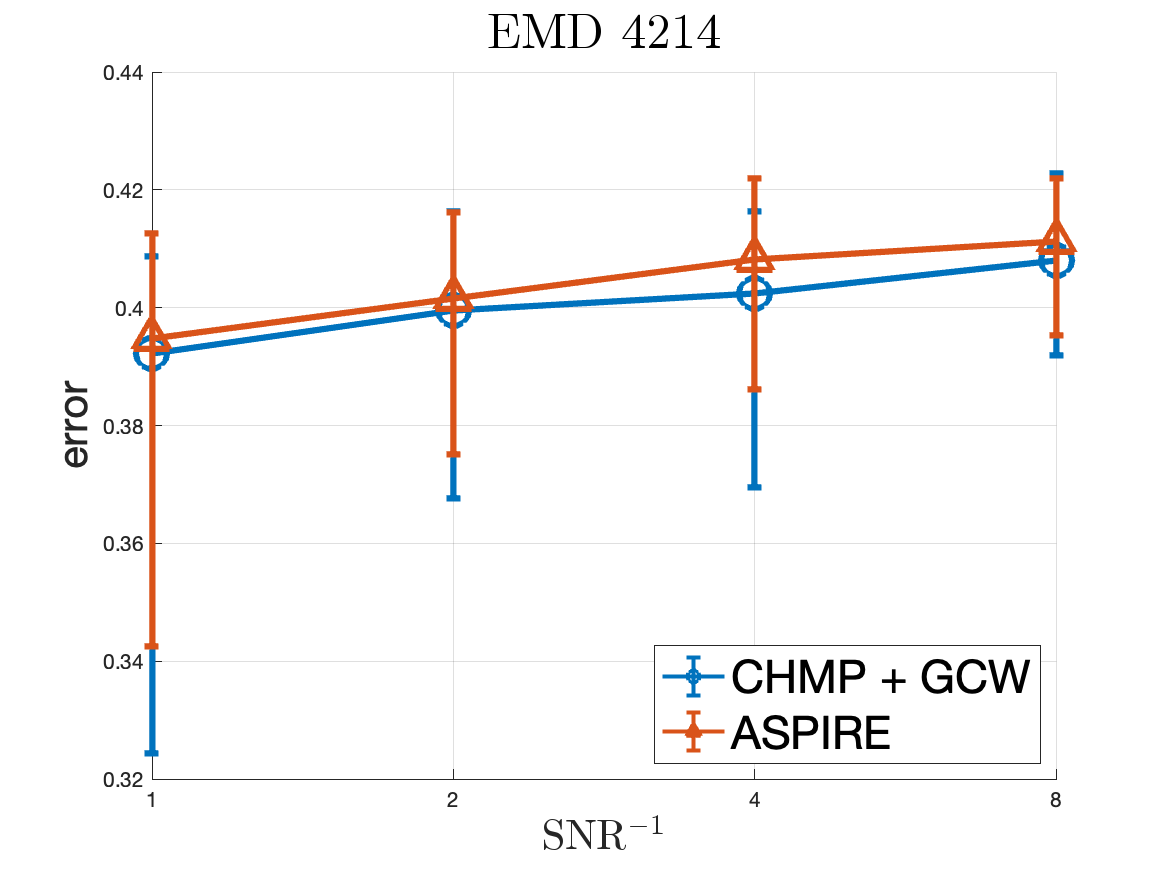} 
    \\
    \caption{\centering Comparison of rotation recovery error for simulated cryo-EM projection images for EMD-2858 (left), EMD-2811 (middle) and EMD-4214 (right).}
    \label{fig:simulated}
\end{figure}

Angular Reconstitution is used to determine the relative angles between triple of images \cite{vanheel_angular_1987}. 
One limitation of common lines data is that the relative rotations between images can only be determined up to global rotation and reflection. 
In particular it is impossible to tell directly from common lines data which chiral orientation the relative rotations are in. 
This is a well documented problem of common lines methods in cryo-EM (See \cite{shkolnisky_viewing_2012} and \cite{pragier_graph_2016}). 
To avoid this ambiguity, the chiral orientation of each estimated triple of viewing angles is artificially adjusted so that every triple belongs to the same orientation. 
Finally, CHMP + GCW is applied to the collection of triples to recover the rotations. 
This method is compared to the standard ASPIRE rotation recovery method which estimates viewing angles by voting  and pairwise synchronization of the common lines data. 
The results, recorded as the average Procrustes error \eqref{eq:procrustes} over \(50\) runs with \(10\%\) and \(90\%\) percentile error bars, are in Figure \ref{fig:simulated}. 

There is variability in the performance across the different molecules. 
For EMD-4214, CHMP outperforms ASPIRE for every level of SNR. 
For the other two molecules, CHMP has comparable recovery performance. 
One thing to note is that the ASPIRE method is optimized in several ways to denoise the common lines data for pairwise synchronization. 
The voting procedure, for example, uses all possible triples formed by a pair of images to determine the best relative angle estimation through averaging common lines data. 
Using angular reconstitution to form triples in the CHMP pipeline means that all triples of common lines data are used in the viewing angle estimation directly, many of which may be severely corrupted. 

\section{Conclusion} \label{sec:conclusion}

We established the higher-order group synchronization problem and proposed the first general framework for solving it. Our framework is based on a cycle-based synchronization criteria that forms the mathematical foundations of the higher-order synchronization problem. 
We gave recovery guarantees for CHMP under outliers and noise, and tested its performance on both angular and rotational synchronization tasks. 
We also compared CHMP to a standard cryo-EM reconstruction package, ASPIRE.

Our results indicate that higher-order synchronization has the potential to provide improved estimates in certain domains. 
One area for further explanation however is to test CHMP on real data sets coming from applications such as cryo-EM or other computer vision tasks. 
While the noise models that the synthetic experiments of Section \ref{subsec:synthetic} use to generate the datasets are general and allow for adversarial levels of corruption, they may not represent an accurate noise model for real data sets. 
For example in a computer vision data set, the corruption may be correlated if one bad image is the source of many inaccurate measurements. 
It would be interesting to compare the performance of higher-order synchronization methods such as CHMP on different noise models that may better reflect real world data.

One of the main challenges of higher-order synchronization is dealing with higher-order data.
Hypergraph data structures suffer from increased storage and computational costs compared to more traditional graphs. 
Improvements in hypergraph tasks such as cycle searching will lead to direct improvements in the speed of CHMP, and potentially other higher-order synchronization methods. 

While our framework was tested on \(SO(2)\) and \(SO(3)\), it remains to see how CHMP performs on other compact groups such as \(\mathbb{Z}/2\mathbb{Z}\) or \(S_n\). 
Further, not all groups of interest in synchronization applications are compact. 
It will be useful to construct a higher-order synchronization method that works on broader classes of groups, including groups such as \(SE(d)\).

CHMP synchronizes higher-order data by estimating the corruption levels of the higher-order measurements using cycle consistency measures. 
With these estimates in hand, we then finish the group synchronization task by considering a refinement of the data 
which transforms the data into pairwise data and applies traditional pairwise synchronization methods (Section~\ref{sec:recovery}). 
But to retain the information redundancy, one might consider a method for the latter step of recovering the vertex potential that acts directly on the higher-order structures too. For example, if the higher-order data can be encoded in a tensor, previous work suggests the potential for an extension of the weighted spectral method GCW to tensors by higher-order singular value decomposition (HOSVD) \cite{miao_tensor_2024,muller_multilinear_2025}. 

\section*{Acknowledgments}
The authors are grateful to Gilad Lerman for suggesting they consider building on ideas from CEMP for higher-order synchronization.  
The authors would also like to thank Tommi Muller for valuable discussions during the early stages of this research.

\bibliographystyle{plainurl}
\bibliography{references.bib}

\appendix

\section{Additional Proofs} \label{appendix:CHMP-analysis-proofs}

\begin{proof}[Proof of Theorem \ref{thm:noisyrecovery}]
    Let \(h \in H\) and \(C \in \mathcal{C}_{n+1}^{n-1}\) such that \(h \in N_C\). 
    As in the proof of Theorem \ref{thm:noiselessrecovery}, the induction hypothesis is \(\epsilon(t) + \delta/2  < 1/2n\beta_t\) for some \(t\). 
    First, it can be shown that
    \begin{equation}\label{eq:basestepinequality}
        \epsilon_{h}(0) \leq \frac{\sum_{C\in N_{{h}}} |d_C  - s_{h}^*|}{|N_{{h}}|} \leq \frac{|B_{{h}}| + \sum_{C\in G_{{h}}} \sum_{h' \in N_C \setminus h} s^*_{h'}}{|N_{{h}}|}
        \leq \frac{|B_{{h}}|}{|N_{{h}}|}  + \frac{|G_{{h}}|}{|N_{{h}}|}\cdot n\delta \leq \lambda  + n\delta.
    \end{equation}
    Thus \eqref{eq:basestepinequality} and the assumptions of the theorem imply that 
    \[
        \epsilon(0) +\frac{\delta}{2} \leq  \lambda + \frac{2n+1}{2}\delta < \frac{1}{2n\beta_0}.
    \]
    Now using \eqref{eq:inequalityforerror} and the fact that \(\max_{h \in H_g} s_{h}^* < \delta\) by the assumption, it follows that
    \begin{align} \label{eq:noisyfinalinequality}
        \epsilon_{h}(t+1) & \leq n\delta + \frac{\sum_{C\in B_{{h}}} \exp\left({-\beta_t\left(\sum_{h' \in N_C \setminus h} s_{h'}(t)\right)}\right) \left(\sum_{h' \in N_C \setminus h} s^*_{h'}\right)}{\sum_{C\in G_{{h}}} \exp\left({-\beta_t\left(\sum_{h' \in N_C \setminus h} s_{h'}(t)\right)}\right)} \nonumber
        \\
        & \leq n\delta + \frac{\sum_{C\in B_{{h}}} \exp\left({\beta_t\left(\sum_{h' \in N_C \setminus h} \epsilon_{h'}(t)\right)} \right) \exp\left({-\beta_t\left(\sum_{h' \in N_C \setminus h} s_{h'}^* \right)}\right)  \left(\sum_{h' \in N_C \setminus h} s^*_{h'}\right)}{\sum_{C\in G_{{h}}} \exp\left({-\beta_t\left(n \delta + \sum_{h' \in N_C \setminus h} \epsilon_{h'}(t)\right)}\right)} \nonumber
        \\
        &\leq n\delta + \frac{(e\beta_t)^{-1}\cdot \sum_{C\in B_{{h}}} \exp\left({\beta_t\left(\sum_{h' \in N_C \setminus h} \epsilon_{h'}(t)\right)} \right) }{\sum_{C\in G_{{h}}} \exp\left({-\beta_t\left(n \delta + \sum_{h' \in N_C \setminus h} \epsilon_{h'}(t)\right)}\right)}. 
    \end{align}
    Maximizing \eqref{eq:noisyfinalinequality} over \(h\) on both sides gives us
    \begin{equation}\label{eq:reducedinequality}
        \epsilon(t+1) \leq  n\delta + \beta_t^{-1} \cdot \frac{|B_{{h}}|}{|G_{{h}}|} \cdot \exp\left({\beta_t(2n\epsilon(t) + n \delta) - 1} \right).
    \end{equation}
    Then using the induction and theorem assumptions and \eqref{eq:reducedinequality} it can be shown that
    \[
        \epsilon(t+1) + \frac{\delta}{2} \leq \frac{(2n+1)\delta}{2} + \frac{1}{\beta_t} \cdot \frac{\lambda}{1-\lambda} < \frac{1}{2n\beta_t},
    \]
    which gives \eqref{eq:max_error}.
    Finally since 
    \[
        \frac{1}{\beta_{t+1}} \geq (2n^2+n)\delta + \frac{2n\lambda}{(1-\lambda)\beta_t} \text{ and } \beta_0 < \frac{(1-(2n+1)\lambda)}{(2n^2+n)(1-\lambda)\delta},
    \]
    for all \(t\geq 0\), \(\beta_t\) is bounded by
    \[
        \beta_t < \frac{(1-(2n+1)\lambda)}{(2n^2+n)(1-\lambda)\delta}.
    \]
    Since \(\beta_t\) is increasing, \(\varepsilon\) (defined in equation \eqref{eq:varepsilon}) is well defined  and satisfies \(0 < \varepsilon \leq 1\). 
    Moreover, equation \eqref{eq:max_error} implies
    \[
        \limsup_{t \to \infty} \max_{h \in H} |s_{h}(t) - s_{h}^*|  \leq \frac{1}{2n\varepsilon} \cdot \frac{(2n^2+n)(1-\lambda)\delta}{(1-(2n+1)\lambda)} - \frac{1}{2}\delta = \left(\frac{2n+1}{2\varepsilon} \cdot \frac{(1-\lambda)}{(1-(2n+1)\lambda)} - \frac{1}{2}\right)\delta. \qedhere
    \]
\end{proof}

\begin{proof}[Proof of Proposition \ref{prop:sample-complexity}]
    Under the UCMH model, \(s^*_{h} = \widehat{s}_{h}\) with probability \(1\) if and only if there are at least \(2\) cycles in \(G_h\). 
    To see this, consider any two cycles \(C_1\) and \(C_2\) such that \(C_1, C_2 \in B_h\).
    Then, by the assumption that \(\mathbb{P}(g = 1)=0\) for any \(g \in G\), \(\mathbb{P}(d_{C_1} = d_{C_2}) = 0\). 
    So with probability \(1\), \(s^*_h\) is the mode of \(D_h\). 
    
    Define \(X_C := \mathbf{1}_{C \in G_h}\) where \(C\) is a cycle. 
    These are independent and identically distributed Bernoulli random variables with mean \(\mu = p^nq_g^n\) and thus they satisfy the Chernoff bound
    \begin{equation}\label{eq:chernoff}
        \mathbb{P}\left(\left|\frac{1}{m}\sum_{l=1}^m X_l - \mu\right| > \eta\mu\right) < 2\exp\left(-(\eta^2\mu m)/3\right)
    \end{equation}
    for \(0 <\eta <1\), where the cycles \(C \in N_h\) are indexed by \(l \in V\), since the cycle set is \(\mathcal{C}_{n+1}^{n-1}\).
    Equation \eqref{eq:chernoff} implies that
    \[
        \mathbb{P}\left(|G_h| \geq 2\right) = \mathbb{P}\left(\frac{1}{m}\sum_{l = 1}^m X_l \geq \frac{2}{m}\right) > 1 - \exp\left({-\frac{1}{3}\left(1 - \frac{2}{m\mu}\right)^2 p^nq_g^nm}\right).
    \]
    Then for \(c \geq 75n/16\) and \(m >2\), if \(m /\log(m) \geq c/(p^nq_g^n)\) then \(\frac{2}{mp^nq_g^n} < 1/5\). Thus 
    \[
        \mathbb{P}(|G_h| \geq 2) > 1 - \exp\left(-\frac{16}{75} (\mu m)\right).
    \]
    Finally, take the union bound over all hyperedges and apply the assumption to get:
    \[
        \mathbb{P} \left(\min_{h \in H} |G_h| \geq 2\right) > 1 - m^n e^{-\frac{16}{75} (\mu m)} = 1-m^{n - 16c/75}. \qedhere
    \]
\end{proof}

\section{Angular Synchronization} \label{appendix:angular-synch}

In this section the experiments of Section \ref{subsubsec:rotation-synch} are repeated for the group \(SO(2)\). 
Elements of \(SO(2)\) can be thought of as \(2 \times 2\) orthogonal matrices with determinant \(1\) defined by the rotation angle \(\theta\): 
\begin{equation}\label{eq:angularrotations}
    R_{\theta} = \begin{pmatrix} \cos(\theta) & \sin(\theta) \\ -\sin(\theta) & \cos(\theta)\end{pmatrix}.
\end{equation}
To measure the angular error between the set of recovered angles \(\{\theta_i\}_{i \in V}\) and the ground truth angles \(\{\theta_i^*\}_{i \in V}\) the orthogonal Procrustes problem in \eqref{eq:procrustes} is solved for \(\{R_{\theta_i}\}_{i \in V}\) and \(\{R_{\theta_i^*}\}_{i \in V}\). 
This can be equivalently formulated as finding the circular mean of the difference in the angles. 
A derivation of this fact is in Appendix \ref{appendix:results}. 

\begin{figure}[htb]
    \centering
    \includegraphics[width=0.3\textwidth]{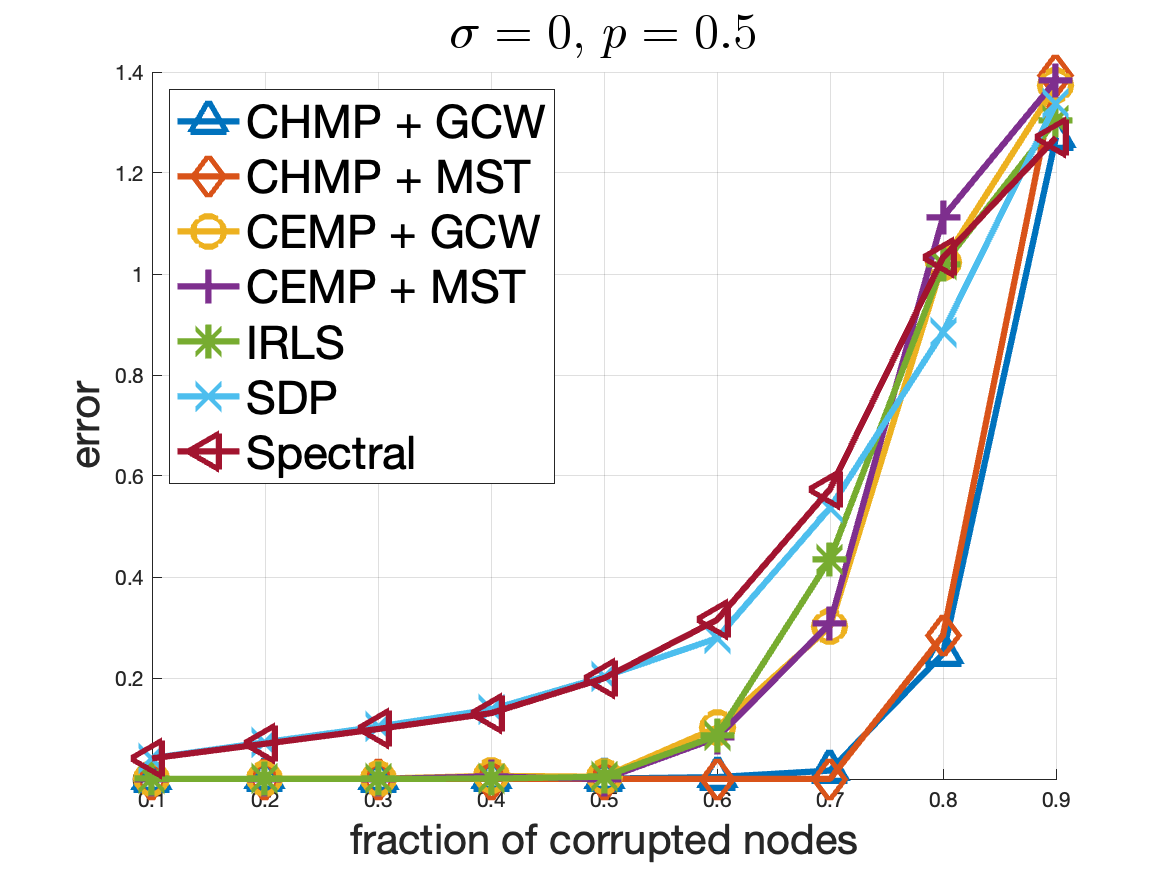}
    \includegraphics[width=0.3\textwidth]{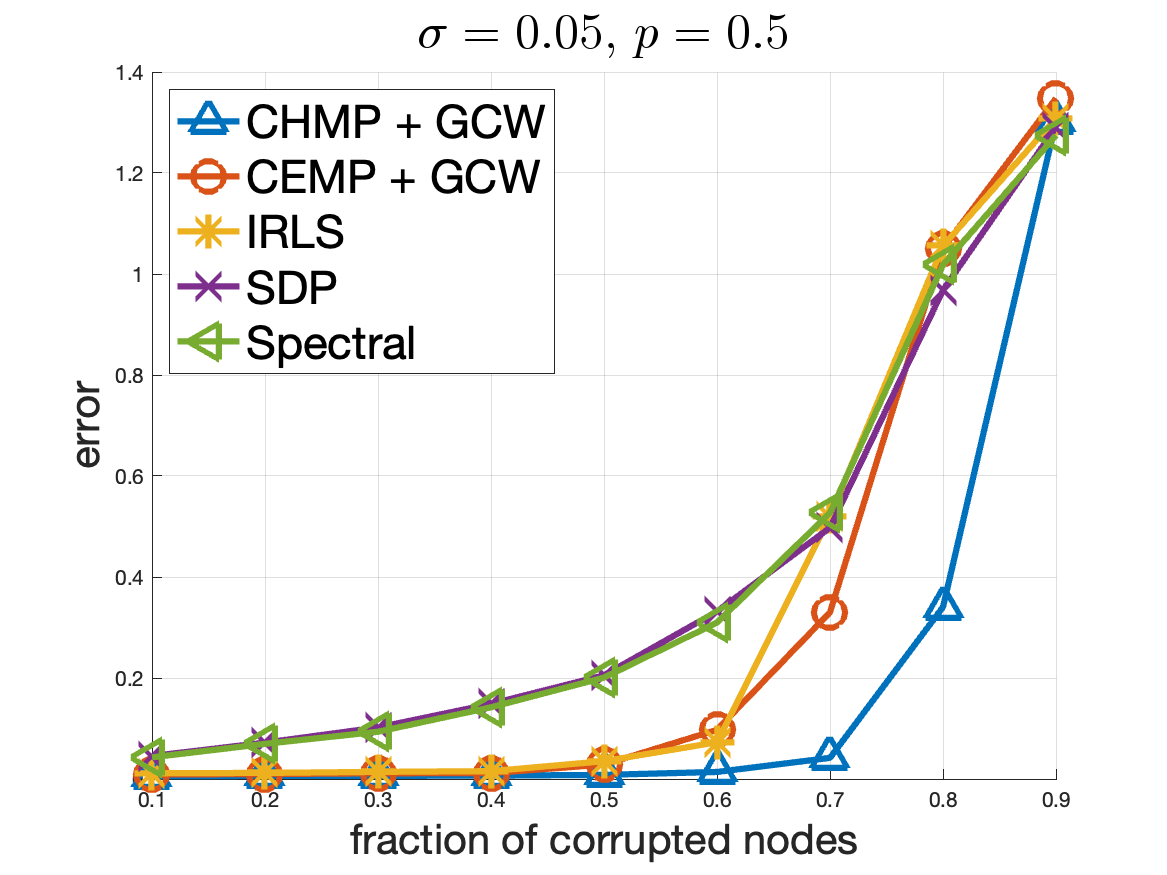} 
    \includegraphics[width=0.3\textwidth]{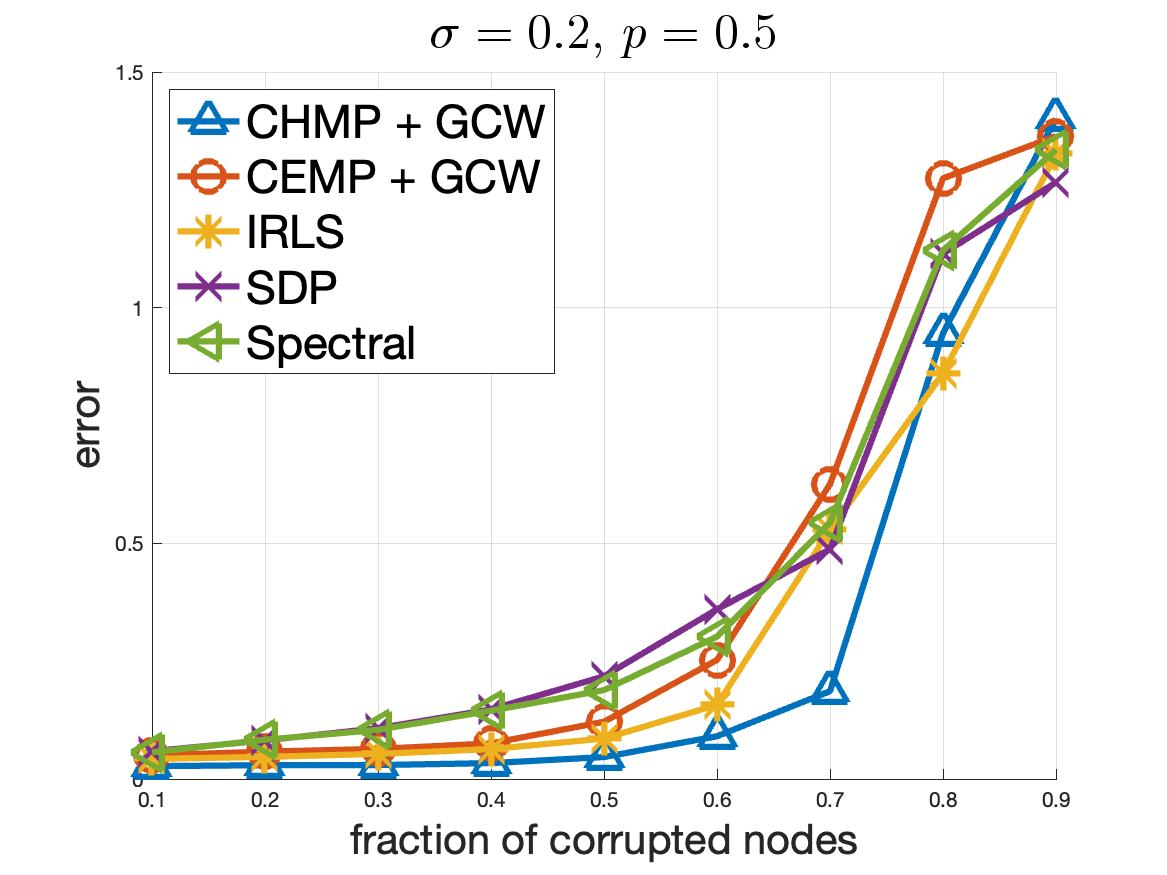}
    \\
    \includegraphics[width=0.3\textwidth]{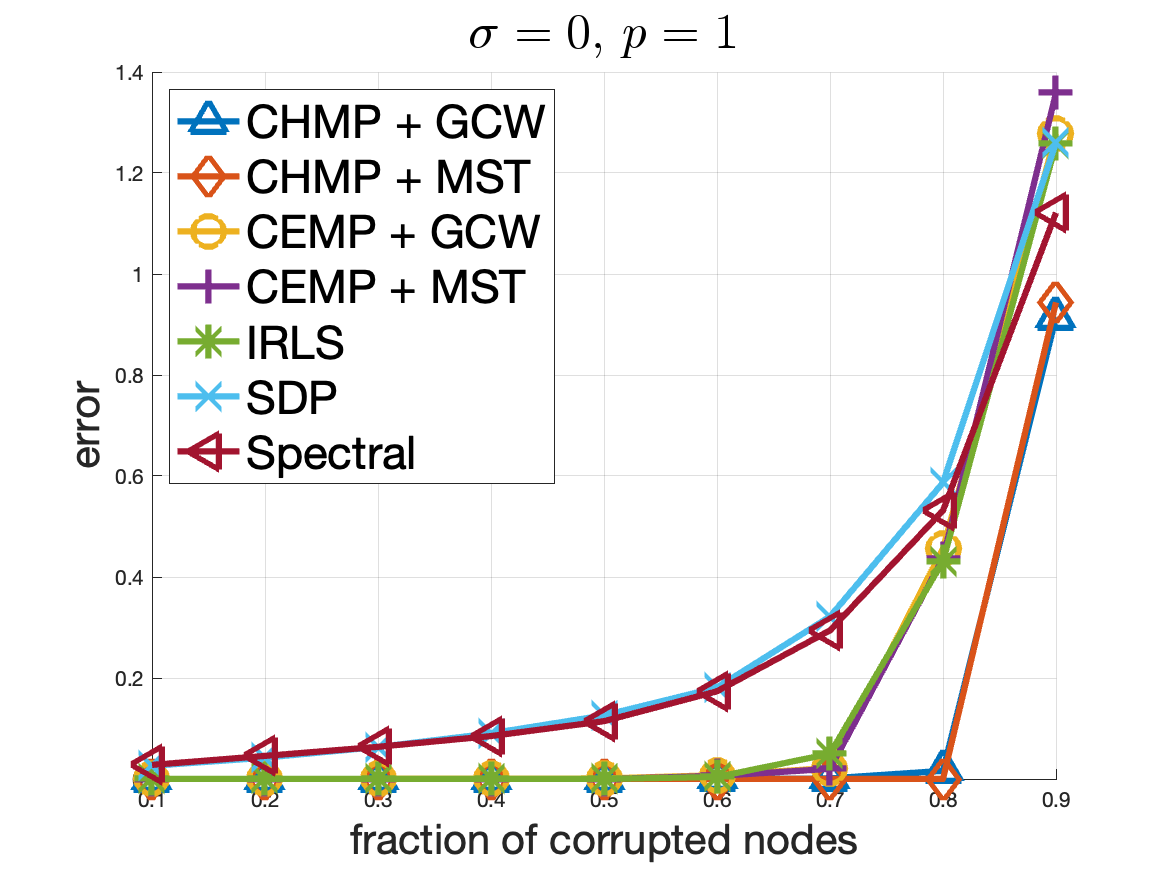}
    \includegraphics[width=0.3\textwidth]{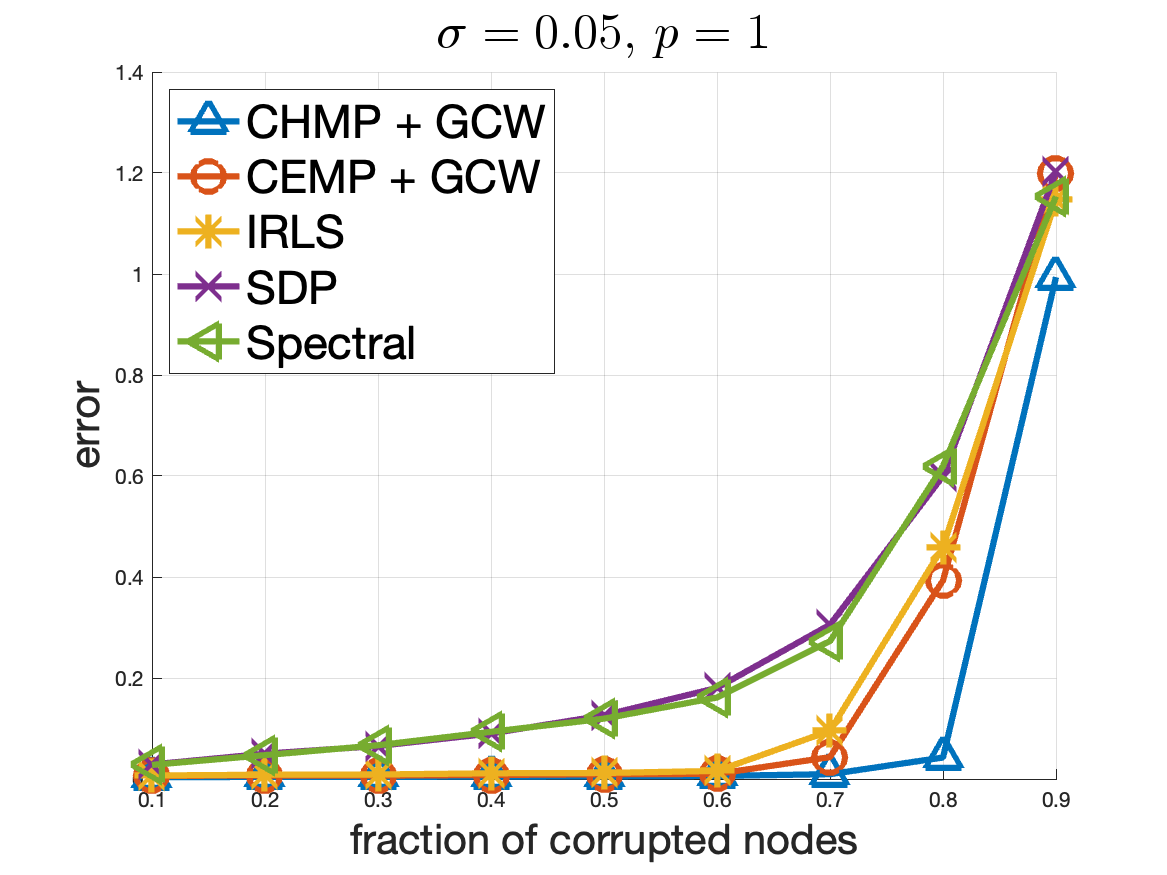} 
    \includegraphics[width=0.3\textwidth]{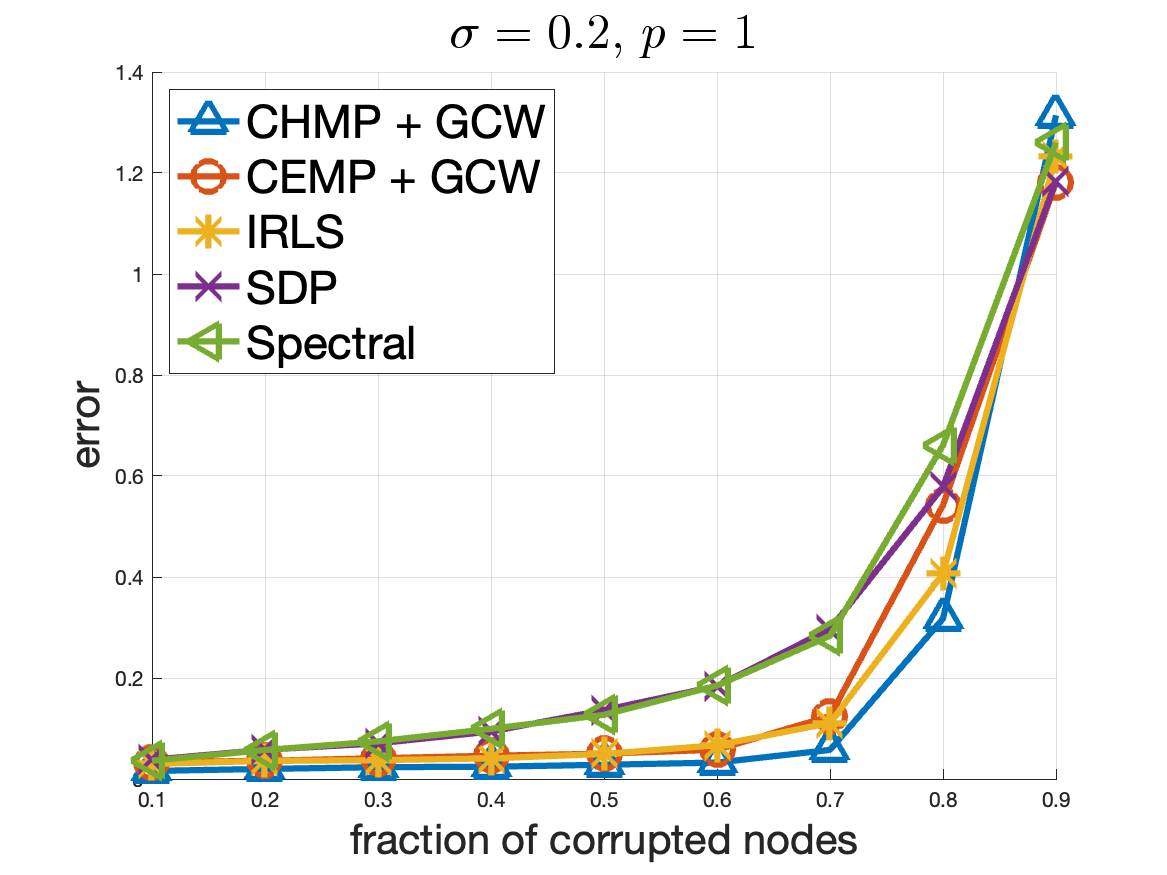}
    \\
    \caption{\centering Comparison of angular rotation recovery methods for \(G = SO(2)\) on data with different levels of corruption and Gaussian noise.}
    \label{fig:outlierSO2}
\end{figure}

The rows of Figure \ref{fig:outlierSO2} correspond to \(p = 0.5\) and \(p=1\) while the columns correspond to \(\sigma = 0, 0.05, 0.2\). 
Both CHMP methods are again outperforming all other methods in the noiseless case, with exact recovery at higher levels of corruption. 
In the noisy experiments CHMP methods outperform the other methods except for possibly in the highest corruption domains where all algorithms do not perform well. 

\section{Additional Remarks}

\subsection{Derivation of the Angular Procrustes Solution} \label{appendix:results}

The angular Procrustes problem for two sets of rotations in \(SO(2)\) is given by
\begin{equation}\label{eq:angularprocrustes}
    \min_{R_{\widetilde{\theta}} \in SO(2)} \frac{1}{m} \sum_{i = 1}^m\left\| R_{\theta_i} - R_{\theta_i^*} R_{\widetilde{\theta}} \right\|^2_F,
\end{equation}
where \(R_{\theta}\) is defined in \eqref{eq:angularrotations}. 
Since \(R_{\theta_1}R_{\theta_2} = R_{\theta_1 + \theta_2}\), the Frobenius norm of \eqref{eq:angularprocrustes} can be rewritten as
\[
    \min_{\widetilde{\theta} \in SO(2)} \frac{1}{m} \sum_{i = 1}^m\left\| R_{\theta_i} - R_{\theta_i^*+\widetilde{\theta}} \right\|^2_F = \min_{\widetilde{\theta} \in SO(2)} \frac{1}{m} \sum_{i = 1}^m \left(4-4\cos(\theta_i - \theta_i^* - \widetilde{\theta})\right).
\]
This is equivalent to the problem
\[
    \max_{\widetilde{\theta} \in SO(2)} \sum_{i = 1}^m \cos(\theta_i - \theta_i^* - \widetilde{\theta}),
\]
whose solution is given by finding the circular mean of the difference \(\theta_i - \theta_i^*\):
\begin{equation}
    \widetilde{\theta} = \arctan2\left(\sum_{i = 1}^m \sin(\theta_i - \theta_i^*),\sum_{i = 1}^m \cos(\theta_i - \theta_i^*)\right).
\end{equation}

\end{document}